\pdfoutput=1
\documentclass[11pt]{article}

\usepackage[preprint]{coling}
\usepackage{marvosym}

\usepackage{times}
\usepackage{latexsym}

\usepackage[T1]{fontenc}
\usepackage[utf8]{inputenc}

\usepackage{microtype}

\usepackage{inconsolata}

\usepackage{graphicx}
\usepackage{algorithm}
\usepackage{algorithmic}

\usepackage{tabularx}
\usepackage{array}
\usepackage{hyperref}
\usepackage{makecell}

\title{TACOMORE: Leveraging the Potential of LLMs in Corpus-based Discourse Analysis with Prompt Engineering}

\author{Bingru Li \\
  University of Birmingham\\ \texttt{bxl426@student.bham.ac.uk} \\\And
  Han Wang\textsuperscript{\Letter} \\
  \texttt{toameek@gmail.com} \\}

\begin{document}

\maketitle

\begin{abstract}
The capacity of LLMs to carry out automated qualitative analysis has been questioned by corpus linguists, and it has been argued that corpus-based discourse analysis incorporating LLMs is hindered by issues of unsatisfying performance, hallucination, and irreproducibility. Our proposed method, TACOMORE, aims to address these concerns by serving as an effective prompting framework in this domain. The framework consists of four principles, i.e., Task, Context, Model and Reproducibility, and specifies five fundamental elements of a good prompt, i.e., Role Description, Task Definition, Task Procedures, Contextual Information and Output Format. We conduct experiments on three LLMs, i.e., GPT-4o, Gemini-1.5-Pro and Gemini-1.5.Flash, and find that TACOMORE helps improve LLM performance in three representative discourse analysis tasks, i.e., the analysis of keywords, collocates and concordances, based on an open corpus of COVID-19 research articles. Our findings show the efficacy of the proposed prompting framework TACOMORE in corpus-based discourse analysis in terms of Accuracy, Ethicality, Reasoning, and Reproducibility, and provide novel insights into the application and evaluation of LLMs in automated qualitative studies.

\end{abstract}

\section{Introduction}
\label{sec:intro}
\begin{figure}
    \centering
    \includegraphics[width=\linewidth]{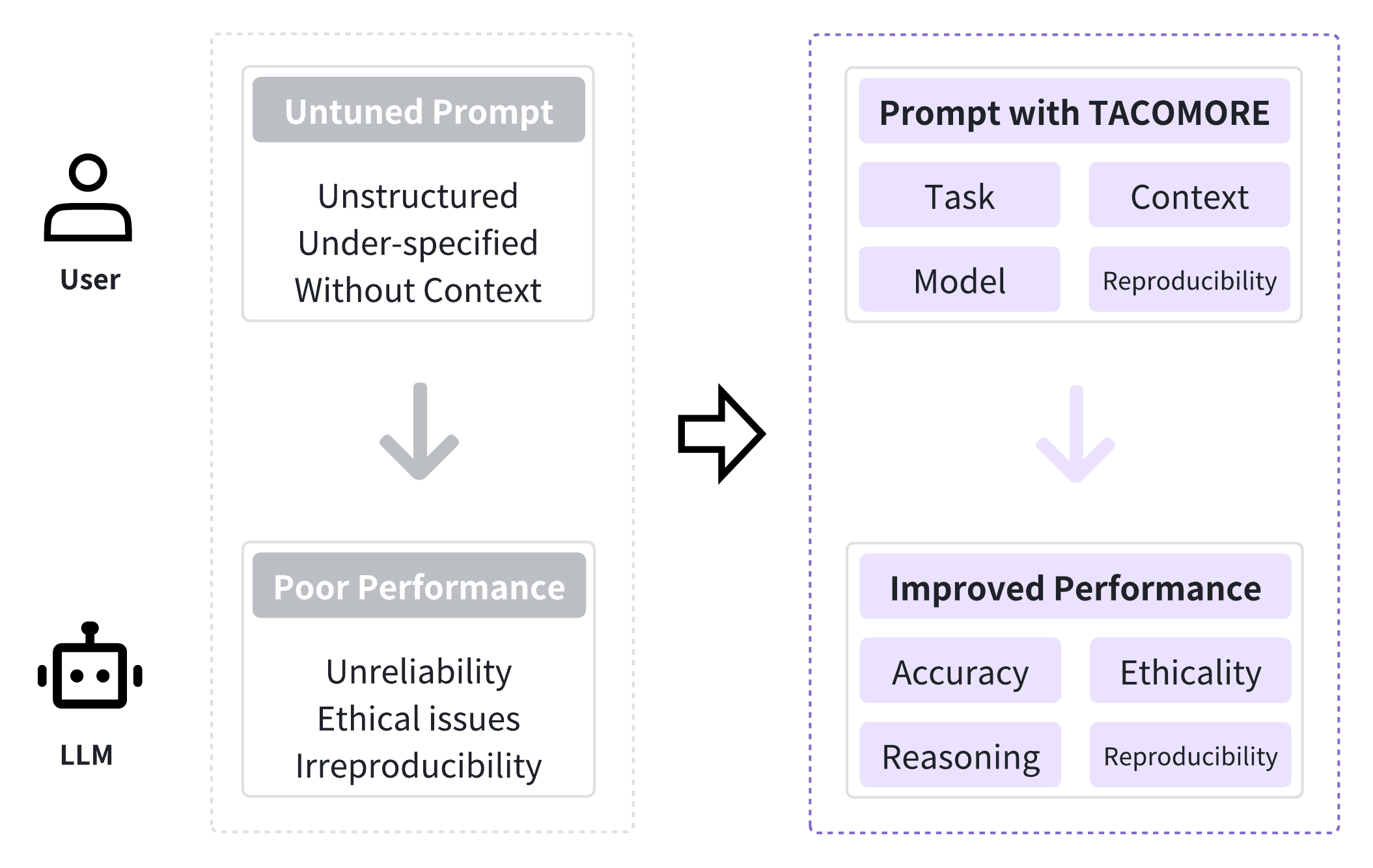}
    \caption{An illustration of comparing LLM performance of an untuned prompt with a prompt with our proposed framework TACOMORE in corpus-based discourse analysis tasks.} 
    \label{fig:intro}
\end{figure}

Before the emergence of Large Language Models (LLMs), AI in corpus linguistics fell short of human-like qualities such as in-depth language understanding and complicated logical reasoning, with only limited abilities in data processing and analysis such as POS tagging \cite{chiche2022posTagging} and sentiment analysis \cite{wankhade2022sentimenAnalysis} based on machine learning and deep learning. Therefore, at an earlier stage, AI was applied in assisting with quantitative studies, while qualitative analysis has remained the human sanctuary. With the breakthrough of LLMs, these advanced models like ChatGPT \footnote{https://openai.com/chatgpt/}, and the latest GPT-4o \footnote{https://openai.com/index/hello-gpt-4o/} and Gemini-1.5-Pro \footnote{https://deepmind.google/technologies/gemini/pro/},  are equipped with such capabilities as long-context modeling, inductive and deductive reasoning, and comprehensive world knowledge that were lacked in previous models, boosting efficiency and efficacy in general tasks and intriguing new discussion about automated qualitative analysis in corpus linguistics. 

Previous research has explored the feasibility of LLMs in qualitative tasks, and some cast doubts on issues of unreliability, ethical concerns such as information fabrication, and poor reproducibility in corpus-based discourse analysis \cite{curry2024generativeAi}, which is an important and challenging domain in corpus linguistics that requires linguistic expertise, logical reasoning and context understanding. In investigating why previous attempts fail, we find that there has not been enough emphasis on designing a structured, task-based prompt, though prompt engineering has been acknowledged as an essential strategy in improving performance \cite{schulhoff2024promptReport}. Also, the importance of contextual information has been omitted for the LLMs in conducting specific tasks, even if human analysts have access to the information when doing the same tasks. 

% Prompt engineering is an emerging area of expertise together with the rise of AI, in that it is widely acknowledged that “Knowing how to effectively structure, evaluate, and perform other tasks with prompts is essential to using these models” \cite{schulhoff2024promptReport}. According to a survey of prompt engineering \cite{sahoo2024reviewofPrompts}, a prompt usually contains instructions including directives, examples, output formatting, style, role, as well as additional information, and many prompting techniques have been developed in order for LLMs to boost performance, for example, Few-shot Prompting \cite{brown2020fewshot} for new tasks without extensive training, and Chain-of-Thought (CoT) Prompting \cite{wei2022chainofthought} for reasoning and logic.

Recognizing the research gap, we propose TACOMORE, an effective prompting framework tailored for LLM application in corpus-based discourse analysis (see Figure \ref{fig:intro}). This framework consists of four principles, i.e., Task, Context, Model, and Reproducibility, which could be further explained through five fundamental elements of a good prompt, including Role Description, Task Definition, Task Procedures, Contextual Information, and Output Format. With TACOMORE, we conduct three representative corpus tasks in discourse analysis, i.e., the analysis of keywords, collocates and concordances, based on an open corpus of COVID-19 Research Articles. We evaluate the efficacy of TACOMORE with established metrics, i.e., Accuracy, Ethicality, Reasoning, and Reproducibility, and compare mainstream LLMs including GPT-4o, Gemimi-1.5-Pro, and Gemini-1.5-Flash. We find that TACOMORE effectively boosts the performance of LLMs, achieving satisfying results the first time ever.

In summary, our contributions are concluded as follows:
\begin{itemize}
\item  We propose an effective prompting framework named TACOMORE to assist with corpus linguists in automated  discourse analysis, which proves useful in improving LLM performance.
\item  We build the evaluation metrics for LLM outputs in discourse analysis tasks, which could be applied to other types of qualitative analysis as well. 
\item We help clarify the standard operating procedure (SOP) of representative discourse analysis tasks with LLMs, and enlighten linguists about the implement of other qualitative tasks in corpus linguistics. 
\end{itemize}
\section{Related Work}
\label{sec:relatedwork}
\subsection{AI Application in Corpus Linguistics}
Before the emergence of ChatGPT, corpus linguistics has benefited from the development of AI in terms of quantitative approaches to data pre-processing, e.g., discourse parsing and segmentation, and common Natural Language Processing (NLP) approaches such as topic modelling \cite{brookes2019utility} and sentiment analysis \cite{lei2021conducting}. With the advancement of LLMs, corpus linguists have explored the capabilities of LLMs in assisting with quantitative analysis \cite{uchida2024using,crosthwaite2023generative,lin2023friendorfoe,kocon2023chatgpt}, with varying degrees of agreement pertaining to the feasibility of replacing handcrafted measures with models like ChatGPT. In comparison, there is a relatively smaller body of literature concerning automating qualitative analysis with LLMs, and scholars hold divergent standpoints. While some have recognized the potential of LLMs in analyzing qualitative data, others have argued that LLMs might not be as reliable as expected. On the one hand, LLMs have demonstrated capabilities in data coding and interpretation that could align with human analysis when given appropriate prompts \cite{siiman2023opportunities,hamalainen2023evaluating}, and showed potential of simplifying inductive learning compared with corpus-based data-driven learning (DDL) that requires complex use of corpus tools \cite{crosthwaite2023generative}. On the other hand, researchers are less optimistic about LLMs for corpus approaches to discourse analysis by comparing the performance of LLMs with human analysis, indicating poor reliability, issues of repeatability and ethical concerns \cite{curry2024generativeAi}. Nevertheless, though the significance of a delicate prompt and relevant context could not be emphasized enough in LLM application in other disciplines, it has not been recognized in linguistic academia. In summary, previous research findings have presented preliminary explorations of LLMs' potential in automated qualitative analysis, but there is still a gap pertaining to a thorough investigation into leveraging the capacity of LLMs with advanced prompting techniques, which serves as the inspiration of the present paper. 

\subsection{Prompt Engineering}
Prompt engineering aims to optimize prompts and thus boost LLM performance. In what follows, we introduce text-based prompting techniques drawing on previous systematic surveys \cite{schulhoff2024promptReport,sahoo2024reviewofPrompts}. 

\noindent \textbf{Zero-Shot}. In Zero-Shot Prompting, no example is given to LLMs. Instead, various settings are provided for them to follow, for example, assigning a specific role to the LLM \cite{wang2023rolePrompting}, specifying the style or tone of the output \cite{lu2023stylePrompting}, instructing LLMs to rephrase questions and respond \cite{deng2023rephraseandRespond}, etc..

\noindent \textbf{In-Context Learning (ICL)}. ICL refers to the capacity of LLMs in understanding a task better by providing examples and/or relevant instructions,  which gives rise to the technique of Few-Shot Prompting \cite{brown2020fewshot}. 

\noindent \textbf{Reasoning}. Designating different thinking patterns could boost the reasoning ability of LLMs. A well-known technique is Chain-of-Thought (CoT) Prompting \cite{wei2022chainofthought}, which guides LLMs to present their thinking process before jumping to a conclusion. This can be combined with Zero-Shot or Few-Shot techniques. Others include Automatic Chain-of-Thought (Auto-CoT) Prompting \cite{zhang2022AutoCoT}, Self-Consistency \cite{wang2022selfConsistency}, and Thread of Thought \cite{zhou2023threadofthought}, to name a few. To further improve LLMs' problem solving ability, researchers also propose techniques of decomposing complex problems into sub-questions, e.g., Least-to-Most Prompting \cite{zhou2022leasttoMost}. 

\noindent \textbf{Self-Reflection}. It can be useful to ask LLMs to reflect on their own outputs in order to improve performance \cite{huang2022self-improve}. Techniques using self-criticism include Take a Step Back Prompting \cite{zheng2023takeaStepBack} and Cumulative Reasoning \cite{zhang2023cumulative}.

There are also prompting templates developed based on the aforementioned techniques that could be transferred to many types of tasks. For instance, the well-known CO-STAR framework proposed by Sheila Teo \cite{teo2023COSTAR}, which consists of 6 parts, i.e., Context, Objective, Style, Tone, Audience, and Response. 
% Though this template works for many tasks, its ability of generalization falls short in corpus linguistic tasks of discourse analysis, which require specific field of expertise and tailored design of prompts, raising a need of a more linguistics-specific framework. 

% \subsection{Corpus Tasks}
% In the present study, we choose three representative methods in corpus linguistics that are used to extract information from corpora, i.e., keywords, collocates, and concordances \cite{baker2023usingCorpora}. Keyword analysis is used to identify the words that are significantly characteristic in a corpus when compared with a reference corpus \cite{scott1997keywords}. Corpus linguists in discourse analysis tend to group the top keywords into different themes which could describe the aboutness and linguistic patterns of a corpus \cite{gabrielatos2018keyness}. After exploring the keywords and having an overall picture of the target corpus, linguists also tap into the text by conducting collocate (co-occurring words) and concordance (co-text) analysis if we intend to investigate a specific theme or the relevant keywords. By exploring the pattern how a keyword is collocated or the context where a keyword is concordanced, we get to obtain a deeper insight into the topic of interest \cite{2010routledgeHandbook}. 

\section{Methodology}

In this section, we first provide a brief overview of the SOP of corpus-based discourse analysis, particularly the three representative tasks under study. We then introduce the methods used in constructing corpus, building our prompting framework TACOMORE, and designing the experiments. In the end, we also clarify how we assess the effectiveness of the proposed framework.

\subsection{Review of SOP in Corpus-based Discourse Analysis Tasks}

Previous literature \cite{baker2023usingCorpora} has defined the SOP of how to use corpora in discourse analysis with corpus tools such as WordSmith \footnote{https://lexically.net/wordsmith} and AntConc \footnote{https://www.laurenceanthony.net/software/antconc}, including frequency and dispersion,  concordances, collocates, and keyness. In our study, we focus on the more complicated tasks, i.e., the analysis of keywords (calculated based on keyness), collocates, and concordances using AntConc 4.2.4 \cite{Antconc}. In keyword analysis \cite{baker2004queryingKeywods}, a frequency list taken from the target corpus is compared against the frequency list of another corpus (either a larger ``reference" corpus or a corpus with a similar length), and a keyword list is computed using statistical probability such as log-likelihood and the chi-squared test, so that we can see which words are key according to how significant they are in terms of frequency when compared with the reference corpus. The next step would be to carry out a close analysis of the collocations and concordances of each keyword and then group them together based on how they serve to construct particular discourse. In the analysis of collocates \cite{gries2004collocates}, linguists look at the common patterns of co-occurrence of the words we find worth studying in a given span of concordance around the node word, e.g., +/-3 and +/-5, which in the corpus tools will be set as L3-R3 and L5-R5 respectively. As for concordance analysis, given a user-defined window size of context to the left and right of the search word or phrase, varying from five words \cite{mahlberg2016dickensConcordance} to 150 characters \cite{baker2020concordance} to the left and right of the node word, concordances help describe and explain observable patterns in context \cite{wulff2021analyzingConcordances}.

\subsection{Corpus Construction} 
Gathering corpora used in previous studies is always hard, expensive, time-consuming, or close-source. To ensure handy reproducibility of our experiments, we choose a corpus that is free of charge and open to public, which comes from an open-source database of COVID-19 research articles archived by the U.S. Centers for Disease Control and Prevention (CDC) \footnote{https://www.cdc.gov/library/researchguides}, collected from December, 2019 to October 9, 2020. In this paper, we focus only on the abstracts of the research articles in order to maintain both the scale and generality of our data. We download the data provided on the official website and retrieve the abstracts. Due to untidiness in the original data, we detect each abstract and filter out non-English text using langdetect \cite{cavnar1994langdetect}.

\subsection{The Prompting Framework}
\begin{figure}[ht]
    \centering
    \includegraphics[width=\linewidth]{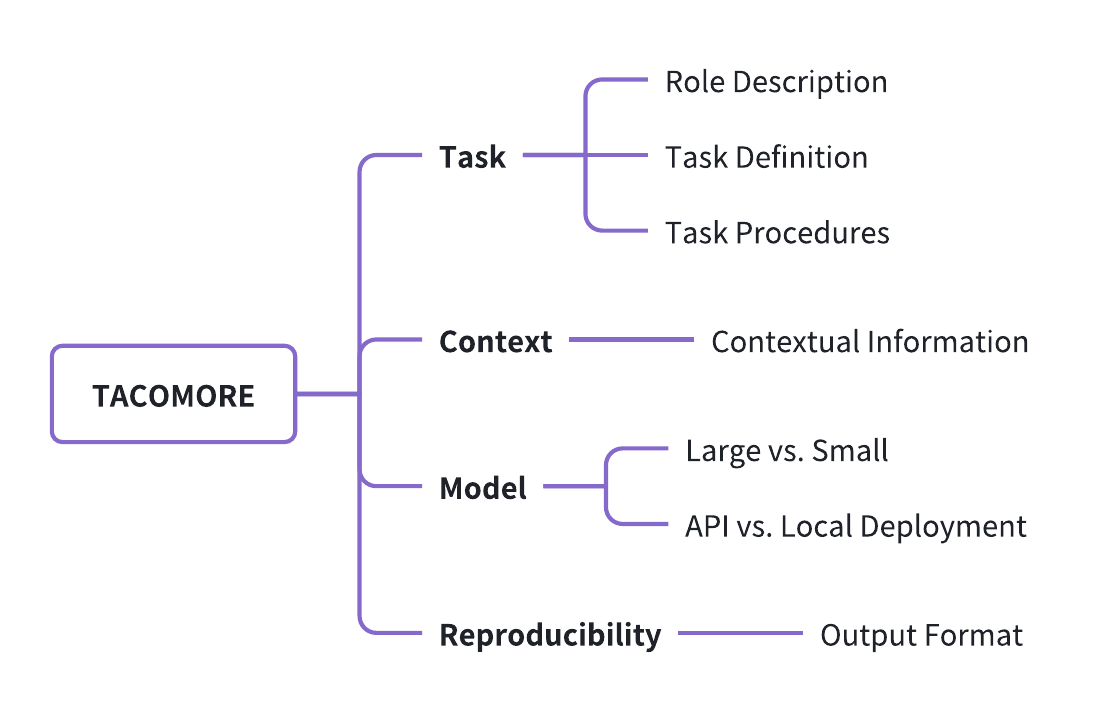}
    \caption{TACOMORE, prompting framework for corpus-based discourse analysis}
    \label{fig:promptingframework}
\end{figure}

\noindent \textbf{TACOMORE.}
To unleash the power of LLMs in discourse analysis, we present the following prompting framework, TACOMORE, as shown in Figure \ref{fig:promptingframework}. 

% First, we introduce the four principles of LLM application specific to this area. Tasks in corpus-based discourse analysis are quite different from the ones with which LLMs are familiarized, in that these tasks require related expertise and contextual information, which make them distinctive from the common benchmarks trained on LLMs \cite{hendrycks2020MMLU,talmor2018commonsenseqa}. We also declare the varying features of different models, and suggest linguists to opt for the one that best suits their need. What's more, to address the concerns about the repeatability LLMs, we also discuss some of the strategies that could help maintain the experiments reproducible. 

\textbf{Task}. Compared with an under-specified goal for the LLMs to stick to in general prompting engineering, the Task element in TACOMORE requires description of a task as detailed as possible and breaking it down to steps to allow our assistants to think and react. For qualitative corpus linguistics tasks, the first thing to determine is the definition and standardized process of a given task. Users are supposed to tell the LLMs the specific field of expertise to activate and the expected style of response by giving it a role, i.e., a corpus linguist, and explain what the target task is about. Also, give them time to think by disassembling  the task into a few steps, so that we can also observe LLMs' chain of thought and quickly pinpoint which step goes wrong.

\textbf{Context}. Linguistic tasks require task-aware context, instead of one that is task-agnostic.  Different from the Context element in other prompting framework that provides the backstory of  the user’s specific scenario, in linguistic tasks, particularly pertaining to discourse analysis, contextual information mostly refers to the co-text around a target word or phrase, a concordance line or the original text, for example. It is necessary to provide as much necessary contextual information as one can to the LLMs, which is available to human analysts when carrying out discourse analysis, so as to align the LLMs' understanding of the task and access to relevant information with those of human linguists. 

\textbf{Model}. Following the second principle, if the target task requires processing of large input data, or context, one needs to choose the right models that could accommodate such demands, e.g., GPT-4o and Gemini-1.5-Pro. As for other tasks with less or little context, smaller models such as ChatGPT, GPT-4o-Mini and Claude-3.5-Sonnet should be able to satisfy one's need. Using Application Programming Interfaces (APIs) or choosing local deployment also makes a difference in implementing  experiments; the former is more convenient whereas the latter is more flexible with default settings. In this study, we opt for APIs out of convenience. 

\textbf{Reproducibility}. Given the inherent characteristics of LLMs, which could not guarantee identical outputs across different trials from the same prompt due to underlying variance in deployment environments, hardware or operators, we settle for the second best and define the reproducibility of task results as meaningful, reasonable, and similar output every time. Another rationale lies in the difficulty for different linguistic professionals in reaching exactly the same conclusions each time, considering factors such as the ambiguity of certain word usage, preference for generalization or specification, divergence in angles and opinions across individuals. 

\noindent \textbf{Fundamental Elements.}
Following the four principles pertaining to how to apply LLMs to corpus-based discourse analysis, we also specify the structure of a so-called ``good" prompt. 
 
\textbf{Role Description}. Describe what kind of role we would like LLMs to take in the specific task, i.e., a corpus linguist. The definition of roleplay helps to narrow down their scope of capabilities according to our expectation. 

\textbf{Task Definition}. It is necessary to designate what task is on target, providing relevant background information, e.g., the data used and the objectives of this task, so that LLMs could retrieve related knowledge and skills. 

\textbf{Task Procedures}. Drawing on the CoT technique, we could split the task into several steps as human would do when conducting the same task. Give clear and orderly instructions to activate the thinking process of LLMs. 

\textbf{Contextual Information}. This could be the original text that the analytical task is based on. For example, the corresponding concordance lines of a list of keywords or collocates. Contextual information could be attached to the end of a prompt as input data.

\textbf{Output Format}. The last compulsory element to require LLMs to present the results in a format that we desire by using delimiters, i.e., marks or symbols that show the beginning and/or end of separate elements, and give examples of an expected output. The more detailed the instruction is, the more organized the outputs would become.

\subsection{Evaluation Metrics}
\begin{figure}[ht]
    \centering
    \includegraphics[width=\linewidth]{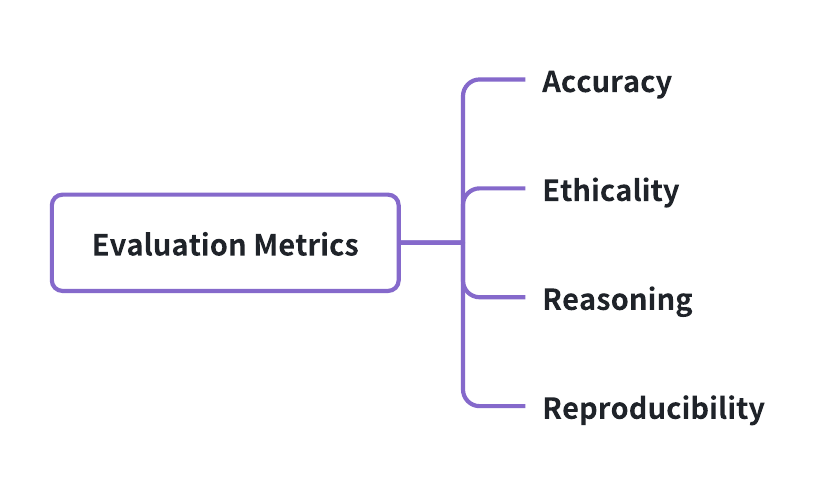}
    \caption{Evaluation Metrics}
    \label{fig:eva_metrics}
\end{figure}
In order to evaluate the effectiveness of TACOMORE, we also propose the metrics of evaluation in Figure \ref{fig:eva_metrics} which are commonly used in LLM assessment \cite{hu2024LLMevaluation} and could address linguists' central concerns. Appraised by human experts, a 5-point Likert scale is applied, where scores 1-5 refer to ``strongly disagree", ``disagree, ``neutral", ``agree", and ``strongly agree", respectively, in the following 4 aspects. 

\noindent \textbf{Accuracy (Acc)} \cite{bengio2000naccuracy}. The LLM performs well in producing desired results, following instructions and yielding high-quality output.

\noindent \textbf{Ethicality (Eth)} \cite{jobin2019ethicality}. The LLM correctly retrieves information from the prompt and data, without any fabrication or omission, and the decisions are air and ethical.

\noindent \textbf{Reasoning (Rea)} \cite{valmeekam2022reasoning}. The LLM logically infers or deduces information, and makes reasonable and coherent decisions. 

\noindent \textbf{Reproducibility (Rep)} \cite{goel2021robustness}. The output remains stable when the same prompt is given many times.

\section{Experiments}
\label{sec:experiments}
\subsection{Experiment Setup}
When designing our target tasks, we are inspired by \cite{curry2024generativeAi} and have followed the established methods mentioned above. The LLMs under examination are GPT-4o, Gemini-1.5-Pro, and Gemini-1.5-Flash. In order to keep human evaluation down to a manageable scale, we do not try to exhaust all possible context but narrow it down when necessary, as is discussed below.

\noindent \textbf{Keyword Analysis.}
For the first task, the thematic and semantic categorization of keywords, we first import our abstract corpus into AntConc 4.2.4 as target corpus, and use the default AmE06 and BEo6 as reference corpora to generate top 100 keyword lists. We then compare the two lists and keep only the concurrent, meaningful keywords, which makes 83 at last. In our prompt, we provide the keyword list and corresponding concordance lines from L10 to R10 (50 random concordances of each keyword), and ask the LLMs to label all keywords from number 1 and then give explanations for the meanings of each keyword in the concordance lines containing it. After that, we require the LLMs should conclude several themes based on the meanings of the keywords and then assign each keyword to one of the themes with reasons given. 

\noindent \textbf{Collocate Analysis.}
For the second task,  we look into the collocates of the keyword \textit{china}, in order to investigate how China is linked to COVID-19 in our corpus. AntConc 4.2.4 is used again to generate the top 100 most frequent collocates of \textit{china} from L5 to R5, and to retrieve 50 random concordance lines of each collocate.  Then we ask LLMs to do two things: 1) to list the content collocates, i.e., nouns, adjectives, verbs, and adverbs; 2) to summarize in what ways China is linked to COVID-19 based on the content collocates from 1) and their corresponding concordance lines. 

\noindent \textbf{Concordance Analysis.}
From the collocate analysis, we find collocations of ``china virus" and ``chinese virus", which are biased and racially charged \cite{darling2020covid19Bias,budhwani2020covidStigma}, and it is advised by the WHO not to attach attach locations or ethnicity to the disease \cite{WHO2020socialStigma}. We attempt to tap into the pragmatic use of these collocations, and so in the last corpus task,  we require LLMs to discuss if the 20 concordance lines containing the two expressions across our corpus actually carry biased indication. 

\noindent \textbf{Target LLMs.}
The three target models carry different characteristics. The Gemini family, i.e., Gemini-1.5-Pro and Gemini-1.5-Flash are capable of dealing with very long input, up to 2 million tokens, while GPT-4o could tolerate a 128k input. Apart from input size, GPT-4o and Gemini-1.5-Pro are trained on more extensive parameters and subsequently perform generally better compared with Gemini-1.5-Flash. 
For the first two tasks, considering the demand of  long inputs containing contextual information, we employ the APIs of these models, and set the greedy decoding strategy in order to limit the diversity of outputs. As for concordance analysis that does not require very long context, we use a third-party platform \footnote{https://poe.com} for convenience. Raw experiment results are shown in Section \ref{subsec:a.1}.

\noindent \textbf{Inte-rater Reliability.}
Two human experts worked independently to evaluate LLM performance, and the results of evaluation in the following sections are presented in their mean scores. An inter-rater reliability test across all 3 tasks and ablation study found near perfect agreement between the two human raters (Krippendorff's Alpha (Ordinal Scale): 0.927).

\subsection{Performance Evaluation}
\label{sec:evalution}

\noindent \textbf{Keyword Analysis}
\begin{table}[ht]
\centering
\setlength{\tabcolsep}{4pt}
\begin{tabular}{lccccc}
\Xhline{3\arrayrulewidth} % 加粗最下面的横线
\textbf{LLMs} & \textbf{Acc} & \textbf{Eth} & \textbf{Rea} & \textbf{Rep} & \textbf{Total}\\
\hline
GPT-4o & 4 & 4 & \textbf{5} & 4 & \textbf{17} \\
Gemini-1.5-Pro & 4 & 4 & 4 & 4 & 16\\
Gemini-1.5-Flash & 3 & 2.5 & 3 & 3.5 & 12\\
\Xhline{3\arrayrulewidth} % 加粗最下面的横线
\end{tabular}

\caption{Evaluation of keyword analysis.}
\label{tab:keywords_score}

\end{table}
\begin{figure}[ht]
    \centering
    \includegraphics[width=\linewidth]{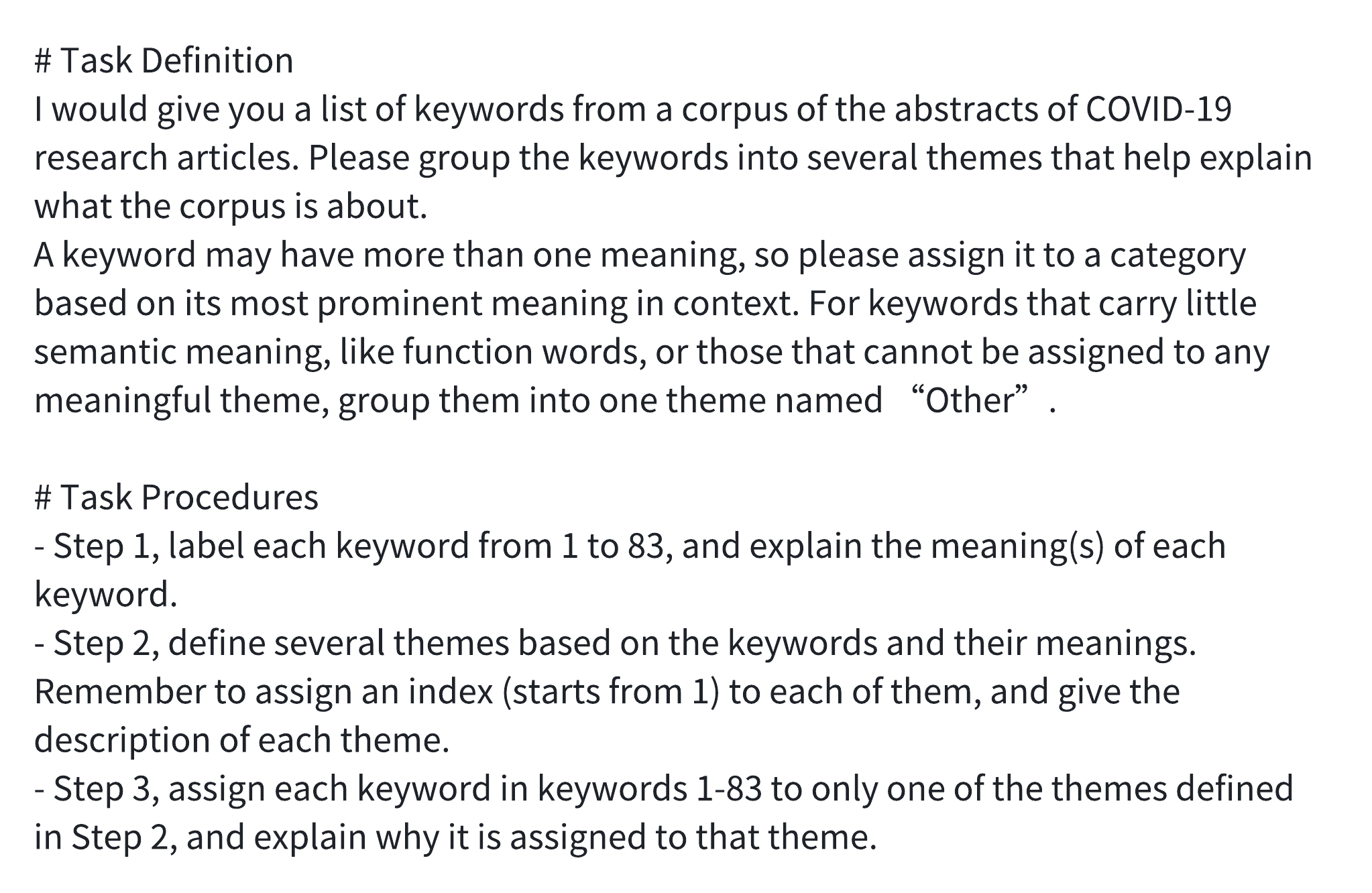}
    \caption{Excerpt of the prompt for keyword analysis.}
    \label{fig:keyword_prompt}
\end{figure}
\begin{figure}
    \centering
    \includegraphics[width=\linewidth]{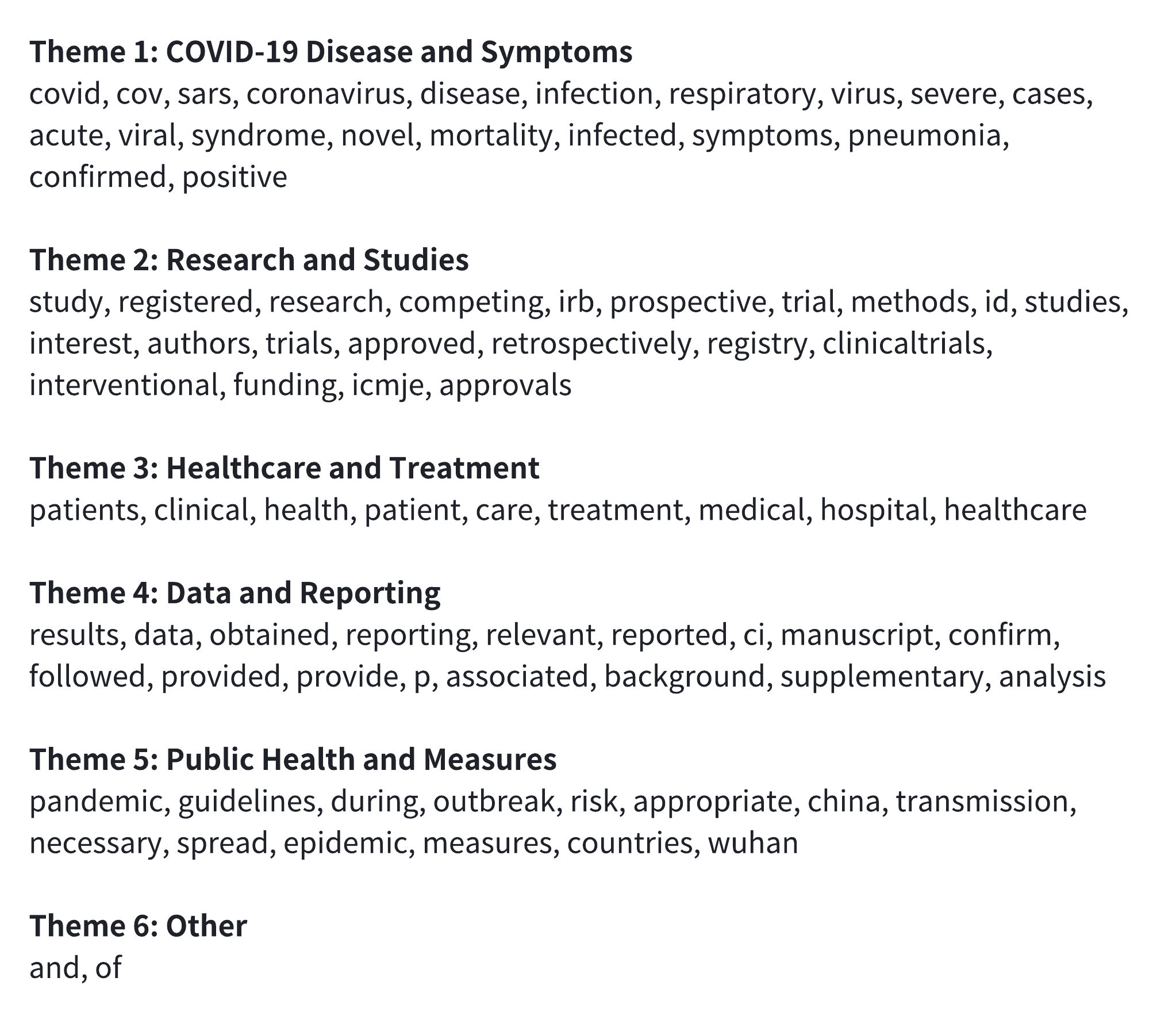}
    \caption{An example of theme categorization by GPT-4o.}
    \label{fig:keyword_example}
\end{figure}

As shown in Table \ref{tab:keywords_score}, both GPT-4o and Gemini-1.5-Pro achieve high scores with TACOMORE, which demonstrates the high-quality of their performance in completing this task, while Gemini-1.5-Flash shows less competence. Figure \ref{fig:keyword_prompt} is part of the prompt for our keyword analysis task. Figure \ref{fig:keyword_example}  is a replicate of GPT-4o's output of categorizing keywords into themes, and by referring back to the original contextual data provided to the LLM, we can see the categorization is reasonable and correct in terms of the number of themes, theme description, and the grouping of keywords. 
Besides, it is worth noting that GPT-4o and Gemini-1.5-Flash give quite stable results pertaining to both the number and description of themes, while Gemini-1.5-Pro tends to shift its analytical perspective across time, giving 5-8 themes with varying degrees of specificity in our experiments. 

\noindent \textbf{Collocate Analysis}
\begin{table}[ht]
\centering
\setlength{\tabcolsep}{4pt}
\begin{tabular}{lcccccc}
\Xhline{3\arrayrulewidth} % 加粗最下面的横线
\textbf{LLMs} & \textbf{Acc} & \textbf{Eth} & \textbf{Rea} & \textbf{Rep} & \textbf{Total}\\
\hline
Gemini-1.5-Pro & 4 & 4 & \textbf{4.5} & 4 & \textbf{16.5}\\
Gemini-1.5-Flash & 4 & 3.5 & \textbf{4.5} & 3.5 & 15.5\\
\Xhline{3\arrayrulewidth} % 加粗最下面的横线
\end{tabular}

\caption{Evaluation of collocate analysis.}
\label{tab:collocates_score}

\end{table}

\begin{figure}[ht]
    \centering
    \includegraphics[width=\linewidth]{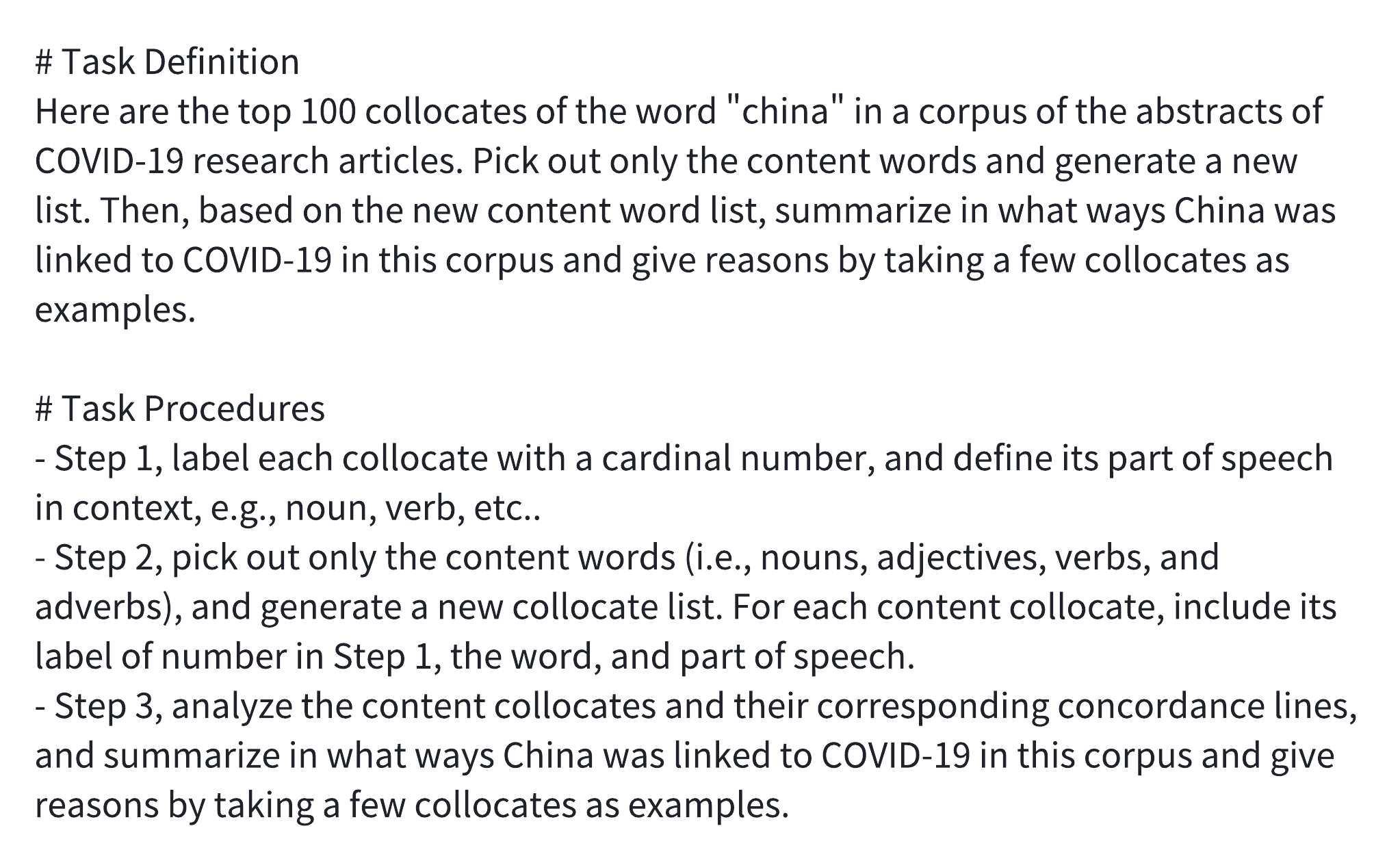}
    \caption{Excerpt of the prompt for collocate analysis.}
    \label{fig:collocates_prompt}
\end{figure}

Due to the length of input for this task, which contains both the collocates and corresponding concordance lines, only Gemini-1.5-Pro and Gemini-1.5-Flash are able to accommodate our need. Both LLMs complete the task as instructed and yield reasonable summaries, as shown in Table \ref{tab:collocates_score}. 

With our prompt (see Figure \ref{fig:collocates_prompt}), in Step 1, both models assign the word classe to each collocate, but the decisions are not always stable because in the original text the same word may be used in different ways. In the second step, both LLMs are able to rule out all function words defined in Step 1 and generate a complete list of content collocates. Finally, based on the new list, both models yield satisfying outputs. Gemini-1.5-Pro precisely pinpoints how China is associated to COVID-19 in the corpus at a more generalized scale, typically in the aspect of ``Origin and Spread", ``Research and Response", and ``Global Impact". As for Gemini-1.5-Flash, it is prone to be more detailed, defining more themes (5-6) with more examples. However, we detect slight errors and edition in both models' citation of the original corpus, and due to Flash's preference for showing more instances, it also makes more mistakes in citing context.

\noindent \textbf{Concordance Analysis}
\begin{table}[ht]
\centering
\setlength{\tabcolsep}{4pt}
\begin{tabular}{lcccccc}
\Xhline{3\arrayrulewidth} % 加粗最下面的横线
\textbf{LLMs} & \textbf{Acc} & \textbf{Eth} & \textbf{Rea} & \textbf{Rep} & \textbf{Total}\\
\hline
GPT-4o & 4 & \textbf{5} & 4 & 4 & \textbf{17}\\
Gemini-1.5-Pro & 4 & \textbf{5} & 4 & 4 & \textbf{17}\\
Gemini-1.5-Flash & 2.5 & 4 & 2.5 & 2 & 11\\
\Xhline{3\arrayrulewidth} % 加粗最下面的横线
\end{tabular}

\caption{Evaluation of concordance analysis.}
\label{tab:concordance_score}

\end{table}

\begin{figure}[ht]
    \centering
    \includegraphics[width=\linewidth]{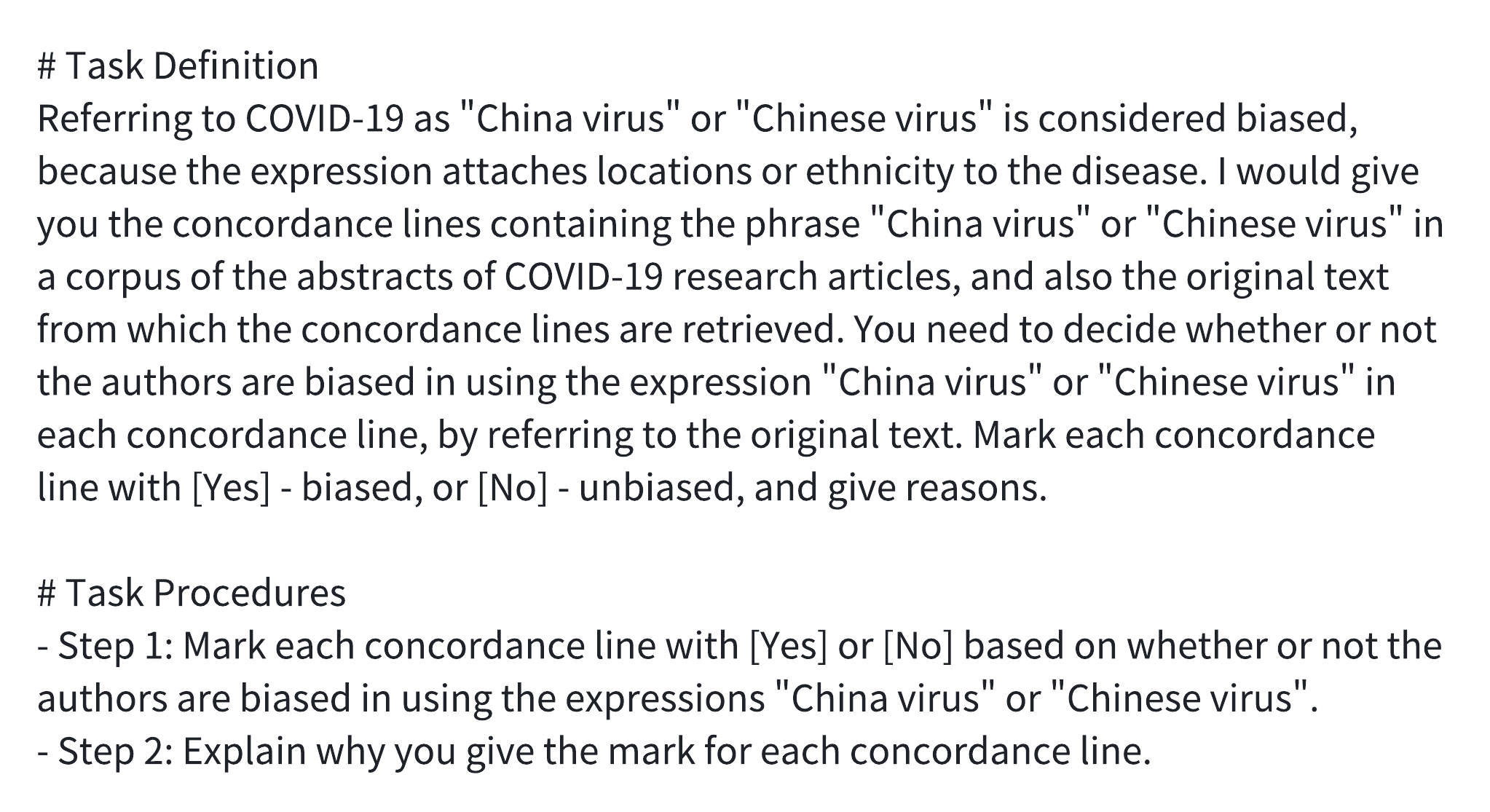}
    \caption{Excerpt of the prompt for concordance analysis.}
    \label{fig:concordance_prompt}
\end{figure}
\begin{figure}[ht]
    \centering
    \includegraphics[width=\linewidth]{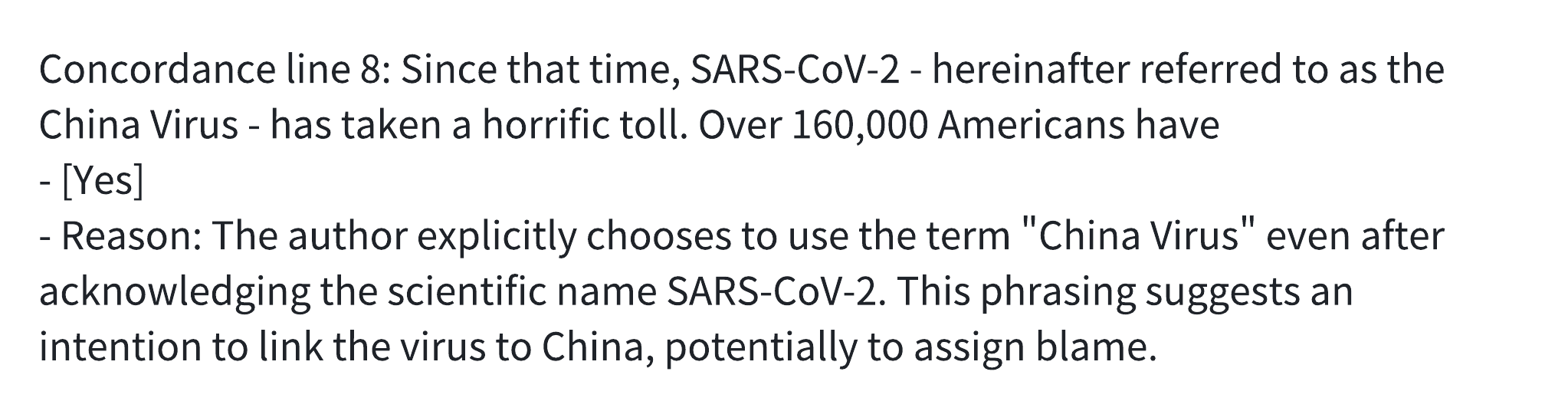}
    \caption{An example of concordance analysis by Gemini-1.5-Pro.}
    \label{fig:concordance_example}
\end{figure}

In concordance analysis, bias detection requires strong ability in understanding subtle meaning. With the given prompt with TACOMORE in Figure \ref{fig:concordance_prompt}, the two large-scale LLMs, GPT-4o and Gemini-1.5-Pro, excel in performing the task, demonstrating human-like understanding and reasoning abilities (see Table \ref{tab:concordance_score}). Of all the concordance lines containing the the collocation ``china virus" and/or ``chinese virus", both models could successfully detect whether a concordance is biased  (see Figure \ref{fig:concordance_example} as an example). Only concordance line 7 is confusing,  with the expression ``the 2019 China virus diagnosis clinical guide" which is hard to decide if it is referring to an external resource without intended bias or reinforcing bias by spontaneously using this expression, leading to unstable answers in both LLMs. Therefore, this case is not taken into consideration in evaluation. Apart from the satisfying performance of GPT-4o and Gemini-1.5-Pro, Gemini-1.5-Flash is less competitive in this task. 

\subsection{Ablation Study}
\label{ablation}
\begin{table*}[ht]
\centering
\begin{tabularx}{\textwidth}{>{\arraybackslash}p{6cm}>{\centering\arraybackslash}X>{\centering\arraybackslash}X>{\centering\arraybackslash}X>{\centering\arraybackslash}X>{\centering\arraybackslash}X}
\Xhline{3\arrayrulewidth} % 加粗最下面的横线
\hline
\textbf{Prompts} & \textbf{Acc} & \textbf{Eth} & \textbf{Rea} & \textbf{Rep} & \textbf{Total}\\
\hline
Baseline & 1 & 1 & 2 & 1 & 5\\
B. + R. D. & 2 & 2 & 2 & 1 & 7\\
B. + R. D. + T. D. & 2.5 & 2 & 2 & 2 & 8.5\\
B. + R. D. + T. D. + T. P. & 3 & 2 & 3 & 3 & 11\\
B. + R. D. + T. D. + T. P. + C. I. & 4 & 2 & 4 & 3 & 13 \\
B. + R. D. + T. D. + T. P. + C. I. + O. F. & \textbf{4} & \textbf{4} & \textbf{4} & \textbf{4} & \textbf{16} \\
\hline
\Xhline{3\arrayrulewidth} % 加粗最下面的横线
\end{tabularx}
\caption{Ablation study of keyword analysis. For evaluation metrics, Acc is short for Accuracy, Eth for Ethicality, Rea for Reasoning, Rep for Reproducibility. For prompts, B. is short for Baseline, R. D. for Role Description, T. D. for Task Definition, T. P. for Task Procedures, C. I. for Contextual Information, O. F. for Output Format.}
\label{tab:ablation}
\end{table*}
In this section, we conduct ablation study for all three tasks with Gemini-1.5-Pro as our exemplar (see Section \ref{subsec:a.2} for all results). We extend evaluation and discussion on keyword analysis, which serves as a better showcase compared with the other two tasks, in that it is usually the starting point for a linguist to study a target corpus and requires more complicated skillset. 

The Baseline prompt, referring to previous research \cite{curry2024generativeAi},  helps generate a very general and superficial summary of themes from the keyword list given. Also the instruction of “thematic and lexical categories” misleads the LLM to produce the lists of themes and lexicons separately. Then we add Role Description (R. D.) , which leads the LLM to provide a more detailed explanation about its categorization, including the needs to consider the specific context of some keywords and also a more thorough corpus investigation, showing the characteristics of a corpus linguist. But still the keywords are grouped into thematic and lexical categories separately. By giving Task Definition (T. D.), the LLM finally understands the task, successfully assigning keywords to several themes. Then, we also define Task Procedures (T. P.) based on CoT in order to know how the model thinks throughout the process. This time, the LLM yields a clearer output, presenting its analysis at each step. But limited by inadequate understanding of keyword meanings due to lack of context, the grouping is done based on only the superficial meanings. Hence we provide Contextual Information (C. I.) so that the LLM manages to understand the contextual meaning of the keywords and thus categorize them in a more practical way. However, all previous tryouts fail to exhaust all keywords when assigning themes, and to counter this issue, we add the element Output Format (O. F.) to compel the LLM to count each and every keyword provided. As shown in Table \ref{tab:ablation}, the score climbs up every time a new element is added to the prompt, illustrating the effectiveness of TACOMORE.

\section{Discussion}
\noindent \textbf{Task.} One of the main contributions of our proposed method is that we underpin the SOP of typical discourse analysis tasks for LLMs by drawing on CoT. To break down a task into implementable pieces of work helps keep LLMs on the right track and opens up a window for human supervision whenever they go astray. We also find the LLMs need intricate instructions particularly when counting numbers is involved. For example, utilize the Output Format element to make sure all necessary data are included. Another finding lies in the importance of the clarity of and coherence between task procedures when a task consists of many steps. 
% For instance, when analyzing collocates (see Appendix 2.2), we ask the LLMs to first define the parts of speech of the collocates, and then keep only the content collocates (i.e., noun, verb, adjective, adverb). Initial attempts were unsuccessful, in that although the LLMs could manage to define the part of speech of each collocate, they failed to exclude those falling out of the scope of content words, for example, prepositions like "across" and "throughout". After many tryouts, we have learned to instruct the LLMs to make coherent decisions across the prompt, e.g., "Step 2, pick out only the content words (i.e., nouns, adjectives, verbs, and adverbs), and generate a new collocate list. For each content collocate, include its label of number in Step 1, the word, and part of speech". 

\noindent \textbf{Context.} Our experiments have confirmed the necessity of contextual information for LLMs to conduct tasks that require understanding of context. Taking the analysis of keywords as an instance, previous studies, e.g., \cite{curry2024generativeAi} have not paid enough attention to providing context, but simply given a list of keywords and asked LLMs to group them into themes, whereas human analysts had access to concordance lines and other contextual information. This asymmetry of information exposure would presumably result in inaccurate conclusions by the LLMs, as demonstrated in Section \ref{ablation}.
% In Section \nameref{ablation}, we try the Baseline prompt which contains only description of the keyword analysis task and the keyword list, without any concordance lines given. In this case, the given information is not enough for the LLMs to explain the meanings of keywords in context, and thus they fail to do thematic grouping correctly. For example, the keywords \textit{clinicaltrials}, \textit{confirm}, and \textit{approvals} are related to academic ethics, which would not be learned by the LLMs without understanding the corpus context. 
However, it is worth noting that hallucination has not been completely conquered. We notice there are still issues concerning information fabrication in LLMs when referring back to the context, as pointed out in collocate analysis in Section \ref{sec:evalution}.

\noindent \textbf{Model.} Linguists should take into consideration two essential issues pertaining to choice of models, i.e., size of input and generality. We recommend large LLMs such as GPT-4o and Gemini-1.5-Pro which are capable of dealing with long context and have general and transferable abilities in different tasks. For those linguistic tasks requiring less input, e.g., limited contextual information, one could use third-party platforms that allow working with different chatbots with a user-friendly interface.

\noindent \textbf{Reproducibility.} As demonstrated in Section \ref{ablation}, we significantly improve the repeatability of LLMs' outputs with TACOMORE compared with only using a Baseline prompt. From an overall perspective, two large models give quite stable performance withstanding time, generating reasonable outputs that approximate human analysis in most tryouts. TACOMORE plays an indispensable role in this respect by determining clear definition of a corpus task, clarification of each step, and also strict instruction on the formatting of output. 
% Besides, the fact that the corpus data we use are publicly available and easy to access paves way for convenient reproduction. 

\section{Conclusion}
In this paper, we propose an effective prompting framework, TACOMORE, for corpus-based discourse analysis, and also provide standardized procedures for representative tasks. What's more, we clarify the metrics of evaluation which are not only widely used in NLP tasks but also address extant concerns of linguists, which could be reapplied in future research of assessing LLM application in automated qualitative tasks. 

Contributions aside, there are also limitations in our study. Discourse analysis is a part of the methodology in corpus linguistics, which might be further explored in following studies. Also, although the present paper focuses on prompt engineering, other techniques could be explored to boost the performance of LLMs as well, e.g., retrieval-augmented generation (RAG) \cite{lewis2020RAG} and Fine-Tuning (FT). 
% Moreover, an ultimate aspiration might be to turn LLMs into a versatile agent in highly-automated linguistic research, from data collection and processing to quantitative and qualitative analysis, where human experts act as supervisors in monitoring the performance of LLMs, which is expected to be investigated in the future. 

\bibliography{custom}

\appendix

\section{Appendix}
\label{sec:appendix}
% Do not position here
% This is an appendix.

\subsection{Discourse Analysis Tasks with TACOMORE}
\label{subsec:a.1}

\noindent \textbf{Keyword Analysis}.

\textbf{Prompt}

\# Role Description

You are an expert in Corpus Linguistics doing keyword analysis.

 \# Task Definition

I would give you a list of keywords from a corpus of the abstracts of COVID-19 research articles. Please group the keywords into several themes that help explain what the corpus is about. 

A keyword may have more than one meaning, so please assign it to a category based on its most prominent meaning in context. For keywords that carry little semantic meaning, like function words, or those that cannot be assigned to any meaningful theme, group them into one theme named "Other".

 \# Task Procedures

- Step 1, label each keyword from 1 to 83, and explain the meaning(s) of each keyword. 

- Step 2, define several themes based on the keywords and their meanings. Remember to assign an index (starts from 1) to each of them, and give the description of each theme.

- Step 3, assign each keyword in keywords 1-83 to only one of the themes defined in Step 2, and explain why it is assigned to that theme. 

 \# Output Format

Bellow are the outputs of the whole task, content within [] should be replaced with real values. "..." means you should list all cases.

Follow the format strictly, and do not add any extra information:

 - Step 1: 

- Keyword 1: [keyword here]

- Meaning: [here is the meaning(s) of keyword 1]

- Keyword 2: [keyword here]

- Meaning: [here is the meaning(s) of keyword 2]

...

 - Step 2:

- Theme 1: [description of theme 1 here]

- Theme 2: [description of theme 2 here]

...

 - Step 3:

- Keyword 1: [keyword here]

- Theme [index of theme in step 2, e.g., 2]

- Reason: [the reason why keyword 1 is assigned to this theme]

- Keyword 2: [keyword here]

- Theme [index of theme in step 2, e.g., 1]

- Reason: [the reason why keyword 2 is assigned to this theme]

...

  \# Contextual Information

(keywords and concordance lines here)

\textbf{Outputs}

\textbf{GPT-4o}

- Step 1: 

- Keyword 1: covid

- Meaning: COVID-19 refers to the disease caused by the novel coronavirus SARS-CoV-2, which led to a global pandemic starting in late 2019.

- Keyword 2: patients

- Meaning: Individuals receiving medical care, particularly those diagnosed with COVID-19.

- Keyword 3: cov

- Meaning: Abbreviation for coronaviruses, including SARS-CoV-2, the virus responsible for COVID-19.

- Keyword 4: sars

- Meaning: Severe Acute Respiratory Syndrome, a disease caused by the coronavirus SARS-CoV, related to SARS-CoV-2.

- Keyword 5: pandemic

- Meaning: A global outbreak of a disease, specifically referring to the COVID-19 pandemic.

- Keyword 6: coronavirus

- Meaning: A family of viruses that includes SARS-CoV-2, the cause of COVID-19.

- Keyword 7: disease

- Meaning: Refers to COVID-19 and its impact on health.

- Keyword 8: study

- Meaning: Research conducted to understand various aspects of COVID-19.

- Keyword 9: clinical

- Meaning: Pertaining to the treatment and management of COVID-19 in healthcare settings.

- Keyword 10: infection

- Meaning: The invasion and multiplication of SARS-CoV-2 in the body, leading to COVID-19.

- Keyword 11: health

- Meaning: The overall well-being and medical status of individuals, particularly in the context of the COVID-19 pandemic.

- Keyword 12: registered

- Meaning: Studies or trials that have been officially recorded, often in a clinical trials registry.

- Keyword 13: respiratory

- Meaning: Relating to the respiratory system, which is primarily affected by COVID-19.

- Keyword 14: virus

- Meaning: Refers to SARS-CoV-2, the virus causing COVID-19.

- Keyword 15: severe

- Meaning: Describes the intensity of COVID-19 symptoms and cases.

- Keyword 16: cases

- Meaning: Instances of COVID-19 infection.

- Keyword 17: research

- Meaning: Scientific studies conducted to understand COVID-19.

- Keyword 18: and

- Meaning: A conjunction used to connect words or phrases.

- Keyword 19: results

- Meaning: Findings from studies or tests related to COVID-19.

- Keyword 20: data

- Meaning: Information collected for analysis in COVID-19 research.

- Keyword 21: patient

- Meaning: An individual receiving medical care, particularly for COVID-19.

- Keyword 22: guidelines

- Meaning: Recommendations for managing COVID-19.

- Keyword 23: acute

- Meaning: Describes the sudden onset and severe nature of COVID-19 symptoms.

- Keyword 24: viral

- Meaning: Pertaining to the virus SARS-CoV-2.

- Keyword 25: competing

- Meaning: Refers to competing interests or conflicts of interest in research.

- Keyword 26: care

- Meaning: Medical attention and treatment provided to COVID-19 patients.

- Keyword 27: irb

- Meaning: Institutional Review Board, which approves research studies.

- Keyword 28: prospective

- Meaning: Studies designed to follow participants forward in time.

- Keyword 29: trial

- Meaning: Clinical trials conducted to test treatments for COVID-19.

- Keyword 30: obtained

- Meaning: Data or consent that has been collected.

- Keyword 31: treatment

- Meaning: Medical interventions for COVID-19.

- Keyword 32: during

- Meaning: Refers to the time period of the COVID-19 pandemic.

- Keyword 33: outbreak

- Meaning: The sudden increase in COVID-19 cases.

- Keyword 34: reporting

- Meaning: The act of documenting and sharing research findings.

- Keyword 35: relevant

- Meaning: Pertinent to the topic of COVID-19.

- Keyword 36: methods

- Meaning: Techniques used in COVID-19 research.

- Keyword 37: risk

- Meaning: The likelihood of adverse outcomes from COVID-19.

- Keyword 38: appropriate

- Meaning: Suitable or proper in the context of COVID-19 management.

- Keyword 39: reported

- Meaning: Documented or shared information about COVID-19.

- Keyword 40: id

- Meaning: Identifier for clinical trials or studies.

- Keyword 41: syndrome

- Meaning: A group of symptoms that occur together, such as in COVID-19.

- Keyword 42: transmission

- Meaning: The spread of SARS-CoV-2.

- Keyword 43: necessary

- Meaning: Required actions or measures in the context of COVID-19.

- Keyword 44: novel

- Meaning: New or unique, referring to the novel coronavirus.

- Keyword 45: spread

- Meaning: The dissemination of COVID-19.

- Keyword 46: mortality

- Meaning: Death rate associated with COVID-19.

- Keyword 47: epidemic

- Meaning: A widespread occurrence of COVID-19 in a community.

- Keyword 48: ci

- Meaning: Confidence Interval, a statistical measure used in research.

- Keyword 49: manuscript

- Meaning: A document detailing research findings.

- Keyword 50: confirm

- Meaning: To verify or validate information.

- Keyword 51: of

- Meaning: A preposition used to indicate belonging or association.

- Keyword 52: medical

- Meaning: Pertaining to medicine and healthcare.

- Keyword 53: studies

- Meaning: Research investigations.

- Keyword 54: interest

- Meaning: Refers to competing interests or conflicts of interest.

- Keyword 55: followed

- Meaning: Adhered to guidelines or protocols.

- Keyword 56: china

- Meaning: The country where COVID-19 was first identified.

- Keyword 57: hospital

- Meaning: Healthcare facility where COVID-19 patients are treated.

- Keyword 58: provided

- Meaning: Supplied or made available.

- Keyword 59: infected

- Meaning: Individuals who have contracted SARS-CoV-2.

- Keyword 60: symptoms

- Meaning: Signs of COVID-19 infection.

- Keyword 61: authors

- Meaning: Researchers who conducted the studies.

- Keyword 62: trials

- Meaning: Clinical trials testing COVID-19 treatments.

- Keyword 63: healthcare

- Meaning: Medical services and systems.

- Keyword 64: pneumonia

- Meaning: A severe lung infection caused by COVID-19.

- Keyword 65: provide

- Meaning: To supply or make available.

- Keyword 66: approved

- Meaning: Officially accepted or authorized.

- Keyword 67: p

- Meaning: P-value, a statistical measure in research.

- Keyword 68: associated

- Meaning: Linked or connected with.

- Keyword 69: confirmed

- Meaning: Verified cases of COVID-19.

- Keyword 70: measures

- Meaning: Actions taken to control COVID-19.

- Keyword 71: retrospectively

- Meaning: Looking back at past data.

- Keyword 72: positive

- Meaning: Confirmed presence of SARS-CoV-2.

- Keyword 73: background

- Meaning: Context or information that sets the scene.

- Keyword 74: registry

- Meaning: A database of clinical trials or studies.

- Keyword 75: clinicaltrials

- Meaning: A registry for clinical trials.

- Keyword 76: countries

- Meaning: Nations affected by COVID-19.

- Keyword 77: interventional

- Meaning: Studies involving active treatment.

- Keyword 78: funding

- Meaning: Financial support for research.

- Keyword 79: icmje

- Meaning: International Committee of Medical Journal Editors.

- Keyword 80: wuhan

- Meaning: The city in China where COVID-19 was first identified.

- Keyword 81: approvals

- Meaning: Official permissions for research.

- Keyword 82: supplementary

- Meaning: Additional material provided with research.

- Keyword 83: analysis

- Meaning: Detailed examination of data.

 - Step 2:

- Theme 1: COVID-19 Disease and Symptoms

- Description: Keywords related to the disease itself, its symptoms, and clinical manifestations.

- Theme 2: Research and Studies

- Description: Keywords related to scientific research, studies, and clinical trials.

- Theme 3: Healthcare and Treatment

- Description: Keywords related to healthcare services, treatment, and patient care.

- Theme 4: Data and Reporting

- Description: Keywords related to data collection, analysis, and reporting of research findings.

- Theme 5: Public Health and Measures

- Description: Keywords related to public health measures, guidelines, and actions taken to control the spread of COVID-19.

- Theme 6: Other

- Description: Keywords that do not fit into the above themes or are function words with little semantic meaning.

 - Step 3:

- Keyword 1: covid

- Theme 1

- Reason: Refers to the disease COVID-19.

- Keyword 2: patients

- Theme 3

- Reason: Refers to individuals receiving medical care.

- Keyword 3: cov

- Theme 1

- Reason: Refers to coronaviruses, including SARS-CoV-2.

- Keyword 4: sars

- Theme 1

- Reason: Refers to Severe Acute Respiratory Syndrome, related to COVID-19.

- Keyword 5: pandemic

- Theme 5

- Reason: Refers to the global outbreak of COVID-19.

- Keyword 6: coronavirus

- Theme 1

- Reason: Refers to the family of viruses including SARS-CoV-2.

- Keyword 7: disease

- Theme 1

- Reason: Refers to COVID-19 and its impact on health.

- Keyword 8: study

- Theme 2

- Reason: Refers to research conducted on COVID-19.

- Keyword 9: clinical

- Theme 3

- Reason: Pertains to the treatment and management of COVID-19.

- Keyword 10: infection

- Theme 1

- Reason: Refers to the invasion of SARS-CoV-2 in the body.

- Keyword 11: health

- Theme 3

- Reason: Refers to the overall well-being and medical status.

- Keyword 12: registered

- Theme 2

- Reason: Refers to studies or trials officially recorded.

- Keyword 13: respiratory

- Theme 1

- Reason: Relates to the respiratory system affected by COVID-19.

- Keyword 14: virus

- Theme 1

- Reason: Refers to SARS-CoV-2.

- Keyword 15: severe

- Theme 1

- Reason: Describes the intensity of COVID-19 symptoms.

- Keyword 16: cases

- Theme 1

- Reason: Refers to instances of COVID-19 infection.

- Keyword 17: research

- Theme 2

- Reason: Refers to scientific studies on COVID-19.

- Keyword 18: and

- Theme 6

- Reason: A conjunction with little semantic meaning.

- Keyword 19: results

- Theme 4

- Reason: Refers to findings from studies or tests.

- Keyword 20: data

- Theme 4

- Reason: Refers to information collected for analysis.

- Keyword 21: patient

- Theme 3

- Reason: Refers to an individual receiving medical care.

- Keyword 22: guidelines

- Theme 5

- Reason: Refers to recommendations for managing COVID-19.

- Keyword 23: acute

- Theme 1

- Reason: Describes the sudden onset and severe nature of symptoms.

- Keyword 24: viral

- Theme 1

- Reason: Pertains to the virus SARS-CoV-2.

- Keyword 25: competing

- Theme 2

- Reason: Refers to competing interests in research.

- Keyword 26: care

- Theme 3

- Reason: Refers to medical attention and treatment.

- Keyword 27: irb

- Theme 2

- Reason: Refers to Institutional Review Board approvals for research.

- Keyword 28: prospective

- Theme 2

- Reason: Refers to studies designed to follow participants forward in time.

- Keyword 29: trial

- Theme 2

- Reason: Refers to clinical trials testing treatments.

- Keyword 30: obtained

- Theme 4

- Reason: Refers to data or consent collected.

- Keyword 31: treatment

- Theme 3

- Reason: Refers to medical interventions.

- Keyword 32: during

- Theme 5

- Reason: Refers to the time period of the pandemic.

- Keyword 33: outbreak

- Theme 5

- Reason: Refers to the sudden increase in cases.

- Keyword 34: reporting

- Theme 4

- Reason: Refers to documenting and sharing research findings.

- Keyword 35: relevant

- Theme 4

- Reason: Pertinent to the topic of COVID-19.

- Keyword 36: methods

- Theme 2

- Reason: Refers to techniques used in research.

- Keyword 37: risk

- Theme 5

- Reason: Refers to the likelihood of adverse outcomes.

- Keyword 38: appropriate

- Theme 5

- Reason: Suitable or proper in the context of management.

- Keyword 39: reported

- Theme 4

- Reason: Documented or shared information.

- Keyword 40: id

- Theme 2

- Reason: Identifier for clinical trials or studies.

- Keyword 41: syndrome

- Theme 1

- Reason: A group of symptoms that occur together.

- Keyword 42: transmission

- Theme 5

- Reason: Refers to the spread of SARS-CoV-2.

- Keyword 43: necessary

- Theme 5

- Reason: Required actions or measures.

- Keyword 44: novel

- Theme 1

- Reason: Refers to the new coronavirus.

- Keyword 45: spread

- Theme 5

- Reason: Refers to the dissemination of COVID-19.

- Keyword 46: mortality

- Theme 1

- Reason: Refers to the death rate associated with COVID-19.

- Keyword 47: epidemic

- Theme 5

- Reason: Refers to a widespread occurrence of COVID-19.

- Keyword 48: ci

- Theme 4

- Reason: Confidence Interval, a statistical measure.

- Keyword 49: manuscript

- Theme 4

- Reason: A document detailing research findings.

- Keyword 50: confirm

- Theme 4

- Reason: To verify or validate information.

- Keyword 51: of

- Theme 6

- Reason: A preposition with little semantic meaning.

- Keyword 52: medical

- Theme 3

- Reason: Pertains to medicine and healthcare.

- Keyword 53: studies

- Theme 2

- Reason: Refers to research investigations.

- Keyword 54: interest

- Theme 2

- Reason: Refers to competing interests in research.

- Keyword 55: followed

- Theme 4

- Reason: Adhered to guidelines or protocols.

- Keyword 56: china

- Theme 5

- Reason: The country where COVID-19 was first identified.

- Keyword 57: hospital

- Theme 3

- Reason: Healthcare facility where patients are treated.

- Keyword 58: provided

- Theme 4

- Reason: Supplied or made available.

- Keyword 59: infected

- Theme 1

- Reason: Individuals who have contracted SARS-CoV-2.

- Keyword 60: symptoms

- Theme 1

- Reason: Signs of COVID-19 infection.

- Keyword 61: authors

- Theme 2

- Reason: Researchers who conducted the studies.

- Keyword 62: trials

- Theme 2

- Reason: Clinical trials testing treatments.

- Keyword 63: healthcare

- Theme 3

- Reason: Medical services and systems.

- Keyword 64: pneumonia

- Theme 1

- Reason: A severe lung infection caused by COVID-19.

- Keyword 65: provide

- Theme 4

- Reason: To supply or make available.

- Keyword 66: approved

- Theme 2

- Reason: Officially accepted or authorized.

- Keyword 67: p

- Theme 4

- Reason: P-value, a statistical measure.

- Keyword 68: associated

- Theme 4

- Reason: Linked or connected with.

- Keyword 69: confirmed

- Theme 1

- Reason: Verified cases of COVID-19.

- Keyword 70: measures

- Theme 5

- Reason: Actions taken to control COVID-19.

- Keyword 71: retrospectively

- Theme 2

- Reason: Looking back at past data.

- Keyword 72: positive

- Theme 1

- Reason: Confirmed presence of SARS-CoV-2.

- Keyword 73: background

- Theme 4

- Reason: Context or information that sets the scene.

- Keyword 74: registry

- Theme 2

- Reason: A database of clinical trials or studies.

- Keyword 75: clinicaltrials

- Theme 2

- Reason: A registry for clinical trials.

- Keyword 76: countries

- Theme 5

- Reason: Nations affected by COVID-19.

- Keyword 77: interventional

- Theme 2

- Reason: Studies involving active treatment.

- Keyword 78: funding

- Theme 2

- Reason: Financial support for research.

- Keyword 79: icmje

- Theme 2

- Reason: International Committee of Medical Journal Editors.

- Keyword 80: wuhan

- Theme 5

- Reason: The city in China where COVID-19 was first identified.

- Keyword 81: approvals

- Theme 2

- Reason: Official permissions for research.

- Keyword 82: supplementary

- Theme 4

- Reason: Additional material provided with research.

- Keyword 83: analysis

- Theme 4

- Reason: Detailed examination of data.

\textbf{Gemini-1.5-Pro}

Step 1: 

- Keyword 1: covid

- Meaning: Refers to the disease caused by the SARS-CoV-2 virus, commonly known as COVID-19.

- Keyword 2: patients

- Meaning: Individuals diagnosed with or receiving treatment for a medical condition, specifically COVID-19 in this context.

- Keyword 3: cov

- Meaning: Abbreviation for Coronavirus, particularly used in the context of SARS-CoV-2.

- Keyword 4: sars

- Meaning:  Severe Acute Respiratory Syndrome, often used in reference to SARS-CoV, the virus that caused the 2003 SARS outbreak. It is also used in the context of SARS-CoV-2.

- Keyword 5: pandemic

- Meaning: A global outbreak of an infectious disease.

- Keyword 6: coronavirus

- Meaning: A family of viruses that cause respiratory illnesses, including SARS-CoV-2.

- Keyword 7: disease

- Meaning: A medical condition or disorder, often used in reference to COVID-19.

- Keyword 8: study

- Meaning:  A research investigation, commonly used in academic and scientific contexts.

- Keyword 9: clinical

- Meaning: Pertaining to the observation and treatment of patients, emphasizing real-world medical practice.

- Keyword 10: infection

- Meaning: The invasion and multiplication of a pathogen, like a virus, in a host organism.

- Keyword 11: health

- Meaning: The overall state of well-being, often used in the context of public health and healthcare systems.

- Keyword 12: registered

- Meaning:  Officially documented, typically referring to research studies being registered in a database.

- Keyword 13: respiratory

- Meaning: Relating to the respiratory system, including the lungs and airways.

- Keyword 14: virus

- Meaning: A microscopic infectious agent that replicates inside living cells, particularly SARS-CoV-2 in this context.

- Keyword 15: severe

- Meaning:  Serious or critical, often used to describe the severity of a medical condition.

- Keyword 16: cases

- Meaning: Instances or occurrences of a disease, specifically COVID-19 cases.

- Keyword 17: research

- Meaning:  Systematic investigation to gain new knowledge, often used in scientific and academic contexts.

- Keyword 18: and

- Meaning:  A conjunction used to connect words or phrases, carrying little semantic meaning in this context.

- Keyword 19: results

- Meaning: Findings or outcomes of a study or investigation.

- Keyword 20: data

- Meaning: Information collected and used for analysis, often in the context of research studies.

- Keyword 21: patient

- Meaning: An individual receiving medical care, specifically a person with COVID-19.

- Keyword 22: guidelines

- Meaning:  Recommendations or instructions for practice, often used in the context of medical treatment or research.

- Keyword 23: acute

- Meaning:  Sudden or severe, typically referring to a medical condition with a rapid onset.

- Keyword 24: viral

- Meaning: Pertaining to a virus, specifically SARS-CoV-2.

- Keyword 25: competing

- Meaning:   Conflicts of interest, particularly those that could influence research findings.

- Keyword 26: care

- Meaning: Medical treatment or attention provided to patients, often in the context of healthcare systems.

- Keyword 27: irb

- Meaning: Institutional Review Board, responsible for ethical oversight of research involving human subjects.

- Keyword 28: prospective

- Meaning: Looking forward in time, often used to describe research studies that follow participants over time.

- Keyword 29: trial

- Meaning: A research study that tests the efficacy or safety of a treatment or intervention.

- Keyword 30: obtained

- Meaning:  Acquired or gathered, typically referring to data or information.

- Keyword 31: treatment

- Meaning: Medical interventions to manage or cure a disease, specifically for COVID-19.

- Keyword 32: during

- Meaning:  A preposition indicating a period of time, carrying little semantic meaning in this context.

- Keyword 33: outbreak

- Meaning: A sudden increase in the occurrence of a disease, particularly the COVID-19 pandemic.

- Keyword 34: reporting

- Meaning:  Documenting and disseminating information, often used in the context of research findings.

- Keyword 35: relevant

- Meaning: Applicable or related to the topic at hand, carrying little semantic meaning in this context.

- Keyword 36: methods

- Meaning: Procedures or techniques used in research investigations.

- Keyword 37: risk

- Meaning:  The possibility of harm or danger, particularly the risk of infection or complications.

- Keyword 38: appropriate

- Meaning: Suitable or correct, carrying little semantic meaning in this context.

- Keyword 39: reported

- Meaning: Documented or communicated, often used in the context of research findings or disease cases.

- Keyword 40: id

- Meaning: Abbreviation for identification, carrying little semantic meaning in this context. 

- Keyword 41: syndrome

- Meaning: A collection of symptoms that characterize a medical condition, often used in the context of severe COVID-19.

- Keyword 42: transmission

- Meaning: The spread of an infectious disease from one person to another.

- Keyword 43: necessary

- Meaning:  Required or essential, carrying little semantic meaning in this context.

- Keyword 44: novel

- Meaning:  New or previously unknown, often used to describe the SARS-CoV-2 virus.

- Keyword 45: spread

- Meaning:  The dissemination or expansion of a disease, particularly the COVID-19 pandemic.

- Keyword 46: mortality

- Meaning: Death, often used in the context of death rates from COVID-19.

- Keyword 47: epidemic

- Meaning: A widespread occurrence of an infectious disease in a community or region.

- Keyword 48: ci

- Meaning: Confidence Interval, a statistical measure of the reliability of an estimate.

- Keyword 49: manuscript

- Meaning: A written document, typically a research paper submitted for publication.

- Keyword 50: confirm

- Meaning:  To verify or establish the truth, often used in the context of research findings or disease diagnosis.

- Keyword 51: of

- Meaning:  A preposition indicating a relationship, carrying little semantic meaning in this context.

- Keyword 52: medical

- Meaning:  Relating to medicine and healthcare, often used in the context of treatments, interventions, and professionals.

- Keyword 53: studies

- Meaning: Research investigations, often used to refer to multiple research projects.

- Keyword 54: interest

- Meaning:  Matters of concern or relevance, often used in the context of competing interests in research.

- Keyword 55: followed

- Meaning:   Adhered to or complied with, carrying little semantic meaning in this context.

- Keyword 56: china

- Meaning:  The country where the COVID-19 outbreak originated.

- Keyword 57: hospital

- Meaning:  A medical facility providing healthcare services.

- Keyword 58: provided

- Meaning:   Supplied or given, carrying little semantic meaning in this context.

- Keyword 59: infected

- Meaning: Affected by a pathogen, specifically individuals with COVID-19.

- Keyword 60: symptoms

- Meaning: Physical or mental signs of a medical condition.

- Keyword 61: authors

- Meaning: Researchers or writers of scientific papers.

- Keyword 62: trials

- Meaning: Research studies testing the efficacy or safety of a treatment, often used to refer to multiple clinical trials.

- Keyword 63: healthcare

- Meaning: The system providing medical care and services, including professionals, institutions, and policies.

- Keyword 64: pneumonia

- Meaning:  Inflammation of the lungs, a common symptom of severe COVID-19.

- Keyword 65: provide

- Meaning:   To supply or offer, carrying little semantic meaning in this context.

- Keyword 66: approved

- Meaning:  Authorized or endorsed, often used in the context of medical treatments or research ethics.

- Keyword 67: p

- Meaning:  P-value, a statistical measure of the significance of research findings.

- Keyword 68: associated

- Meaning:  Connected or related to, often used to describe factors linked to COVID-19 severity or outcomes.

- Keyword 69: confirmed

- Meaning:   Verified or established as true, typically referring to disease diagnosis.

- Keyword 70: measures

- Meaning: Actions or steps taken, often used in the context of public health interventions or treatment strategies.

- Keyword 71: retrospectively

- Meaning:  Looking back in time, often used to describe research studies that analyze past data.

- Keyword 72: positive

- Meaning:  Indicating the presence of a condition, typically a positive COVID-19 test result.

- Keyword 73: background

- Meaning:  Contextual information or history, often used to describe patient characteristics or research settings.

- Keyword 74: registry

- Meaning: A database for collecting and organizing information, often used for tracking clinical trials or patient data.

- Keyword 75: clinicaltrials

- Meaning: A website for registering and reporting clinical trials, specifically ClinicalTrials.gov.

- Keyword 76: countries

- Meaning:   Nations or geographic regions, often used in the context of the global spread of COVID-19.

- Keyword 77: interventional

- Meaning: Involving an intervention or treatment, typically used to describe research studies that test a specific action.

- Keyword 78: funding

- Meaning:   Financial support for research projects.

- Keyword 79: icmje

- Meaning: International Committee of Medical Journal Editors, an organization that sets standards for medical journal publication.

- Keyword 80: wuhan

- Meaning:  The city in China where the COVID-19 outbreak originated.

- Keyword 81: approvals

- Meaning: Authorizations or endorsements, often used in the context of research ethics or medical treatment decisions.

- Keyword 82: supplementary

- Meaning:  Additional or supporting, often used in the context of providing extra information in research papers.

- Keyword 83: analysis

- Meaning:   The process of examining and interpreting data, a key component of research studies.

 - Step 2:

- Theme 1: COVID-19 Pandemic

- Description: This theme encompasses keywords related to the global outbreak of COVID-19, its spread, impact, and associated public health concerns.

- Theme 2: Virus and Infection

- Description: This theme includes keywords related to the SARS-CoV-2 virus, its characteristics, transmission, and the infection it causes.

- Theme 3: Patients and Symptoms

- Description: This theme encompasses keywords related to individuals diagnosed with COVID-19, their symptoms, and clinical manifestations of the disease.

- Theme 4: Treatment and Care

- Description: This theme includes keywords related to medical interventions, healthcare systems, and the management of COVID-19 patients.

- Theme 5: Research and Studies

- Description: This theme encompasses keywords related to scientific investigations, data analysis, clinical trials, and the publication of findings.

- Theme 6: Ethics and Reporting

- Description: This theme includes keywords related to ethical oversight of research, reporting guidelines, and the dissemination of research findings.

- Theme 7: Other

- Description:  This theme includes keywords that do not fit into the other defined themes, such as conjunctions or prepositions that carry little semantic meaning.

 - Step 3:

- Keyword 1: covid

- Theme 1

- Reason: Refers to the disease itself, which is the central focus of the COVID-19 pandemic.

- Keyword 2: patients

- Theme 3

- Reason: Directly related to individuals diagnosed with COVID-19.

- Keyword 3: cov

- Theme 2

- Reason: An abbreviation for Coronavirus, specifically used in the context of SARS-CoV-2.

- Keyword 4: sars

- Theme 2

- Reason:  Related to both SARS-CoV and SARS-CoV-2, emphasizing the viral aspect.

- Keyword 5: pandemic

- Theme 1

- Reason: Describes the global nature of the COVID-19 outbreak.

- Keyword 6: coronavirus

- Theme 2

- Reason: Refers to the family of viruses, including SARS-CoV-2.

- Keyword 7: disease

- Theme 1

- Reason: A general term for a medical condition, but often used in reference to COVID-19.

- Keyword 8: study

- Theme 5

- Reason:  A common term in research contexts.

- Keyword 9: clinical

- Theme 3

- Reason:  Pertains to patient observation and treatment.

- Keyword 10: infection

- Theme 2

- Reason: Directly related to the process of viral invasion.

- Keyword 11: health

- Theme 4

- Reason: Encompasses healthcare systems and public health responses.

- Keyword 12: registered

- Theme 6

- Reason:   Refers to the official documentation of research studies.

- Keyword 13: respiratory

- Theme 3

- Reason:  Describes a key system affected by COVID-19.

- Keyword 14: virus

- Theme 2

- Reason: A general term for a microscopic infectious agent, often used in reference to SARS-CoV-2.

- Keyword 15: severe

- Theme 3

- Reason:   Used to indicate the seriousness of COVID-19 cases.

- Keyword 16: cases

- Theme 1

- Reason: Refers to instances of COVID-19, a key indicator of the pandemic's spread.

- Keyword 17: research

- Theme 5

- Reason:  A general term for scientific investigation.

- Keyword 18: and

- Theme 7

- Reason: A conjunction with no specific semantic meaning in this context.

- Keyword 19: results

- Theme 5

- Reason:  Relates to the findings of research studies.

- Keyword 20: data

- Theme 5

- Reason: A core component of research and analysis.

- Keyword 21: patient

- Theme 3

- Reason: Refers to an individual with COVID-19.

- Keyword 22: guidelines

- Theme 6

- Reason:  Provides recommendations for medical practice or research.

- Keyword 23: acute

- Theme 3

- Reason: Describes the rapid onset of severe COVID-19 symptoms.

- Keyword 24: viral

- Theme 2

- Reason: Pertaining to the virus, specifically SARS-CoV-2.

- Keyword 25: competing

- Theme 6

- Reason:  Relates to conflicts of interest in research.

- Keyword 26: care

- Theme 4

- Reason: Encompasses medical treatment and support for patients.

- Keyword 27: irb

- Theme 6

- Reason:  Focuses on ethical oversight in research.

- Keyword 28: prospective

- Theme 5

- Reason: Describes a type of research study design.

- Keyword 29: trial

- Theme 5

- Reason: Refers to a research study testing a treatment or intervention.

- Keyword 30: obtained

- Theme 7

- Reason:   A general term with no specific semantic meaning in this context.

- Keyword 31: treatment

- Theme 4

- Reason: Refers to medical interventions for COVID-19.

- Keyword 32: during

- Theme 7

- Reason:   A preposition with no specific semantic meaning.

- Keyword 33: outbreak

- Theme 1

- Reason:  Describes the sudden increase in COVID-19 cases.

- Keyword 34: reporting

- Theme 6

- Reason:  Focuses on the dissemination of research findings.

- Keyword 35: relevant

- Theme 7

- Reason:  A general term with no specific semantic meaning.

- Keyword 36: methods

- Theme 5

- Reason: Refers to research procedures.

- Keyword 37: risk

- Theme 1

- Reason:   A key concern during the pandemic.

- Keyword 38: appropriate

- Theme 7

- Reason:   A general term with no specific semantic meaning.

- Keyword 39: reported

- Theme 6

- Reason:  Relates to the communication of research findings or disease cases.

- Keyword 40: id

- Theme 7

- Reason: An abbreviation with no specific semantic meaning.

- Keyword 41: syndrome

- Theme 3

- Reason:  Describes a collection of symptoms associated with severe COVID-19.

- Keyword 42: transmission

- Theme 2

- Reason: Focuses on the spread of the virus.

- Keyword 43: necessary

- Theme 7

- Reason: A general term with no specific semantic meaning.

- Keyword 44: novel

- Theme 2

- Reason:  Highlights the newness of the SARS-CoV-2 virus.

- Keyword 45: spread

- Theme 1

- Reason:  Describes the expansion of the COVID-19 pandemic.

- Keyword 46: mortality

- Theme 1

- Reason: A key indicator of the pandemic's severity.

- Keyword 47: epidemic

- Theme 1

- Reason:  Refers to the widespread occurrence of COVID-19.

- Keyword 48: ci

- Theme 5

- Reason:  A statistical measure used in research.

- Keyword 49: manuscript

- Theme 5

- Reason:  Relates to the publication of research findings.

- Keyword 50: confirm

- Theme 5

- Reason:  Used in the context of verifying research findings.

- Keyword 51: of

- Theme 7

- Reason: A preposition with no specific semantic meaning.

- Keyword 52: medical

- Theme 4

- Reason: Encompasses healthcare practices and professionals.

- Keyword 53: studies

- Theme 5

- Reason: A general term for research investigations.

- Keyword 54: interest

- Theme 6

- Reason: Relates to potential conflicts of interest in research.

- Keyword 55: followed

- Theme 7

- Reason:  A general term with no specific semantic meaning.

- Keyword 56: china

- Theme 1

- Reason:   The origin of the COVID-19 outbreak.

- Keyword 57: hospital

- Theme 4

- Reason: A key component of the healthcare system.

- Keyword 58: provided

- Theme 7

- Reason:  A general term with no specific semantic meaning.

- Keyword 59: infected

- Theme 3

- Reason:   Refers to individuals with COVID-19.

- Keyword 60: symptoms

- Theme 3

- Reason:  Key indicators of COVID-19.

- Keyword 61: authors

- Theme 5

- Reason:  Refers to researchers who conduct studies.

- Keyword 62: trials

- Theme 5

- Reason: A general term for research studies testing treatments.

- Keyword 63: healthcare

- Theme 4

- Reason: Encompasses the system for providing medical care.

- Keyword 64: pneumonia

- Theme 3

- Reason:   A common symptom of severe COVID-19.

- Keyword 65: provide

- Theme 7

- Reason:  A general term with no specific semantic meaning.

- Keyword 66: approved

- Theme 6

- Reason:  Relates to ethical oversight and authorization of research.

- Keyword 67: p

- Theme 5

- Reason:  A statistical measure used in research.

- Keyword 68: associated

- Theme 3

- Reason:   Used to describe factors linked to COVID-19.

- Keyword 69: confirmed

- Theme 3

- Reason:    Refers to verified cases of COVID-19.

- Keyword 70: measures

- Theme 4

- Reason:  Relates to interventions and strategies to manage COVID-19.

- Keyword 71: retrospectively

- Theme 5

- Reason: Describes a research study design.

- Keyword 72: positive

- Theme 3

- Reason:   Indicates the presence of COVID-19.

- Keyword 73: background

- Theme 5

- Reason:   Provides context for research studies.

- Keyword 74: registry

- Theme 5

- Reason: A database used for tracking clinical trials or patient data.

- Keyword 75: clinicaltrials

- Theme 5

- Reason:  A specific website for registering clinical trials.

- Keyword 76: countries

- Theme 1

- Reason:   Refers to the global impact of the pandemic.

- Keyword 77: interventional

- Theme 5

- Reason:   Describes research studies involving treatments or interventions.

- Keyword 78: funding

- Theme 5

- Reason:   Relates to financial support for research.

- Keyword 79: icmje

- Theme 6

- Reason:  An organization setting standards for medical journal publication.

- Keyword 80: wuhan

- Theme 1

- Reason: The origin of the COVID-19 outbreak.

- Keyword 81: approvals

- Theme 6

- Reason:  Relates to ethical oversight and authorization of research.

- Keyword 82: supplementary

- Theme 5

- Reason:  Refers to providing extra information in research papers.

- Keyword 83: analysis

- Theme 5

- Reason:  A key component of research studies.

\textbf{Gemini-1.5-Flash}

- Step 1: 

- Keyword 1: covid

- Meaning: A shortened form of coronavirus disease 2019, a respiratory illness caused by the virus SARS-CoV-2.

- Keyword 2: patients

- Meaning: People receiving medical treatment or care.

- Keyword 3: cov

- Meaning: Short for coronavirus, a family of viruses that cause respiratory illnesses in animals and humans. 

- Keyword 4: sars

- Meaning: Severe Acute Respiratory Syndrome, a viral respiratory illness caused by a coronavirus.

- Keyword 5: pandemic

- Meaning: A widespread epidemic of a disease that affects a large population, usually across multiple countries or continents. 

- Keyword 6: coronavirus

- Meaning: A family of viruses that cause respiratory illnesses in animals and humans.

- Keyword 7: disease

- Meaning: An abnormal condition affecting the body of an organism that impairs its normal functioning.

- Keyword 8: study

- Meaning: A systematic investigation to gather and analyze information to increase understanding of a phenomenon or to solve a problem.

- Keyword 9: clinical

- Meaning: Relating to the practice of medicine and the treatment of patients.

- Keyword 10: infection

- Meaning: The process of a pathogen invading a host organism and multiplying, causing harm or disease. 

- Keyword 11: health

- Meaning: The state of being free from illness or injury.

- Keyword 12: registered

- Meaning: Officially recorded or listed in a registry. 

- Keyword 13: respiratory

- Meaning: Relating to the process of breathing and the respiratory system, which includes the lungs and airways. 

- Keyword 14: virus

- Meaning: A small infectious agent that replicates only inside the living cells of other organisms. 

- Keyword 15: severe

- Meaning: Very serious or dangerous, causing great harm or damage.

- Keyword 16: cases

- Meaning: Instances or examples of a particular occurrence or condition, often referring to instances of a disease. 

- Keyword 17: research

- Meaning: A systematic investigation to discover and interpret new knowledge, often through a scientific process. 

- Keyword 18: and

- Meaning: A conjunction used to connect words, phrases, clauses, or sentences. 

- Keyword 19: results

- Meaning: The outcome or findings of a study, experiment, or investigation.

- Keyword 20: data

- Meaning: Raw facts or figures that are collected and analyzed to gain information or draw conclusions.

- Keyword 21: patient

- Meaning: A person receiving medical treatment or care. 

- Keyword 22: guidelines

- Meaning: A set of recommendations or rules to provide guidance on a particular subject.

- Keyword 23: acute

- Meaning: Characterized by a sudden onset and a short, severe course.

- Keyword 24: viral

- Meaning: Relating to or caused by a virus. 

- Keyword 25: competing

- Meaning:  Representing a conflict of interests or a situation where two or more things or people are vying for the same thing.

- Keyword 26: care

- Meaning: The provision of medical treatment, support, or attention to someone in need.

- Keyword 27: irb

- Meaning: Institutional Review Board, a committee that reviews research proposals involving human subjects to ensure ethical and safe practices. 

- Keyword 28: prospective

- Meaning: Looking forward or considering future events. 

- Keyword 29: trial

- Meaning: A test or experiment conducted to evaluate the effectiveness or safety of a new drug, treatment, or procedure. 

- Keyword 30: obtained

- Meaning: Acquired or gained something.

- Keyword 31: treatment

- Meaning:  The care or medical procedures provided to a patient to relieve or cure a disease or injury.

- Keyword 32: during

- Meaning:  At some point within a specified time period. 

- Keyword 33: outbreak

- Meaning: A sudden occurrence of a disease in a particular location, often with a rapid increase in cases. 

- Keyword 34: reporting

- Meaning:  The act of giving information about an event, situation, or study. 

- Keyword 35: relevant

- Meaning:  Important or applicable to a particular situation or subject. 

- Keyword 36: methods

- Meaning:  The procedures or techniques used to conduct a study, experiment, or investigation.

- Keyword 37: risk

- Meaning:  The possibility of something bad happening, often expressed as a percentage or probability.

- Keyword 38: appropriate

- Meaning:  Suitable or proper for a particular situation or purpose.

- Keyword 39: reported

- Meaning:  Given information about something, often as a result of a study or observation.

- Keyword 40: id

- Meaning:  Short for identification, often referring to a unique code or number assigned to a person, object, or study.

- Keyword 41: syndrome

- Meaning:  A group of symptoms that occur together and characterize a particular disease or condition. 

- Keyword 42: transmission

- Meaning:  The process of transferring something from one person, place, or thing to another.

- Keyword 43: necessary

- Meaning:  Essential or required for a particular purpose or outcome.

- Keyword 44: novel

- Meaning:  New, original, or unique. 

- Keyword 45: spread

- Meaning:  To become widespread or to extend over a larger area.

- Keyword 46: mortality

- Meaning:  The number of deaths in a particular population or within a given time period. 

- Keyword 47: epidemic

- Meaning:  A widespread occurrence of an infectious disease in a community at a particular time.

- Keyword 48: ci

- Meaning:  Confidence Interval, a range of values that is likely to include the true value of a population parameter. 

- Keyword 49: manuscript

- Meaning:  The written text of a book, article, or other literary work, especially one that is intended for publication. 

- Keyword 50: confirm

- Meaning:  To establish the truth or accuracy of something.

- Keyword 51: of

- Meaning:  A preposition used to indicate possession, belonging, or association. 

- Keyword 52: medical

- Meaning:  Relating to medicine, healthcare, or the treatment of disease.

- Keyword 53: studies

- Meaning:  Systematic investigations to gather and analyze information to increase understanding of a phenomenon or to solve a problem. 

- Keyword 54: interest

- Meaning:  A sense of curiosity or a desire to learn more about something.

- Keyword 55: followed

- Meaning:  Came after something else in a sequence. 

- Keyword 56: china

- Meaning:  A country in East Asia, often mentioned in relation to the origin of the COVID-19 pandemic. 

- Keyword 57: hospital

- Meaning:  A medical facility where patients are admitted for treatment or care.

- Keyword 58: provided

- Meaning:  Given or supplied something.

- Keyword 59: infected

- Meaning:  Affected by a pathogen, such as a virus or bacterium. 

- Keyword 60: symptoms

- Meaning:  Signs or indications of a disease or illness experienced by a patient. 

- Keyword 61: authors

- Meaning:  The people who wrote a book, article, or other literary work.

- Keyword 62: trials

- Meaning:  Tests or experiments to evaluate the effectiveness or safety of a new drug, treatment, or procedure.

- Keyword 63: healthcare

- Meaning:  The provision of medical services to individuals and communities. 

- Keyword 64: pneumonia

- Meaning:  An inflammation of the lungs, often caused by a bacterial or viral infection.

- Keyword 65: provide

- Meaning:  To give or supply something. 

- Keyword 66: approved

- Meaning:  Officially sanctioned or permitted.

- Keyword 67: p

- Meaning:  In statistics, p-value represents the probability of obtaining the observed results or more extreme results if the null hypothesis were true.

- Keyword 68: associated

- Meaning:  Connected or related to something else. 

- Keyword 69: confirmed

- Meaning:  Verified or established as true or accurate. 

- Keyword 70: measures

- Meaning:  Actions taken to achieve a specific outcome or to address a particular problem.

- Keyword 71: interventional

- Meaning:  Involving the direct intervention or treatment of a patient.

- Keyword 72: retrospectively

- Meaning:  Looking back at past events or data.

- Keyword 73: positive

- Meaning:  Indicating the presence of a particular condition or result, often in relation to a test or diagnosis. 

- Keyword 74: background

- Meaning:  The context or information surrounding a particular topic or event. 

- Keyword 75: registry

- Meaning:  A database or record of information, often used to track specific cases or events. 

- Keyword 76: clinicaltrials

- Meaning:  A website that provides information about clinical trials for new drugs, treatments, and procedures.

- Keyword 77: countries

- Meaning:  Independent political entities with defined territories and governments. 

- Keyword 78: interventional

- Meaning:  Involving the direct intervention or treatment of a patient.

- Keyword 79: funding

- Meaning:  Financial support provided for a project or research.

- Keyword 80: icmje

- Meaning:  International Committee of Medical Journal Editors, an organization that sets ethical standards for medical publications. 

- Keyword 81: wuhan

- Meaning:  A city in China, often mentioned as the origin of the COVID-19 pandemic.

- Keyword 82: approvals

- Meaning:  Official permissions or authorizations.

- Keyword 83: supplementary

- Meaning:  Additional or supporting information, often provided as separate files or documents. 

- Keyword 84: analysis

- Meaning:  The process of examining data or information to identify patterns, trends, or relationships.

 - Step 2:

- Theme 1: Disease Characteristics

- Description: Keywords related to the characteristics of the disease COVID-19, including its symptoms, severity, transmission, and impact. 

- Theme 2: Research and Studies

- Description: Keywords related to research on COVID-19, including research methods, findings, and the scientific community's efforts to understand and combat the disease.

- Theme 3: Healthcare System and Practice

- Description: Keywords related to the impact of COVID-19 on healthcare systems, including hospital practices, patient care, resource allocation, and the role of healthcare professionals.

- Theme 4: Public Health and Pandemic

- Description: Keywords related to the public health response to the COVID-19 pandemic, including prevention measures, control efforts, and the global spread of the disease. 

- Theme 5: Other

- Description: Keywords that are not easily categorized into the other themes, such as function words and keywords with multiple meanings. 

 - Step 3:

- Keyword 1: covid

- Theme 1

- Reason: This keyword is the central term for the disease, and it represents the core focus of the corpus.

- Keyword 2: patients

- Theme 3

- Reason: This keyword is related to the provision of healthcare and the impact of COVID-19 on patient care.

- Keyword 3: cov

- Theme 1

- Reason: This keyword represents the viral family that causes COVID-19 and is important in understanding the disease's origins and characteristics.

- Keyword 4: sars

- Theme 1

- Reason: This keyword is related to a similar respiratory illness caused by a coronavirus, and it provides context for understanding COVID-19.

- Keyword 5: pandemic

- Theme 4

- Reason: This keyword describes the global nature of the disease outbreak and its impact on the world.

- Keyword 6: coronavirus

- Theme 1

- Reason: This keyword is the primary term for the family of viruses that cause COVID-19, and it is essential for understanding the disease's nature.

- Keyword 7: disease

- Theme 1

- Reason: This keyword is used to describe the illness caused by the SARS-CoV-2 virus, and it is a fundamental term in understanding the disease's impact.

- Keyword 8: study

- Theme 2

- Reason: This keyword is related to the research being done to understand and combat COVID-19.

- Keyword 9: clinical

- Theme 3

- Reason: This keyword is related to the practice of medicine and patient care, which is a crucial aspect of the corpus.

- Keyword 10: infection

- Theme 1

- Reason: This keyword is essential for describing how the virus spreads and causes illness.

- Keyword 11: health

- Theme 4

- Reason: This keyword is related to the public health impact of COVID-19 and the measures taken to protect public health. 

- Keyword 12: registered

- Theme 2

- Reason: This keyword is used in the context of research studies and clinical trials, which are crucial aspects of the corpus.

- Keyword 13: respiratory

- Theme 1

- Reason: This keyword describes the primary system affected by COVID-19 and is essential for understanding the disease's manifestations.

- Keyword 14: virus

- Theme 1

- Reason: This keyword is the causative agent of COVID-19 and is central to understanding the disease's origins and spread. 

- Keyword 15: severe

- Theme 1

- Reason: This keyword describes the potential seriousness of COVID-19 and its potential complications. 

- Keyword 16: cases

- Theme 4

- Reason: This keyword is related to the number of people infected with COVID-19 and the pandemic's spread. 

- Keyword 17: research

- Theme 2

- Reason: This keyword is used in the context of studies and investigations into COVID-19. 

- Keyword 18: and

- Theme 5

- Reason: This keyword is a conjunction used to connect words, phrases, and clauses, and it carries little semantic meaning.

- Keyword 19: results

- Theme 2

- Reason: This keyword is related to the findings or outcomes of research on COVID-19. 

- Keyword 20: data

- Theme 2

- Reason: This keyword is related to the information collected and analyzed in research studies. 

- Keyword 21: patient

- Theme 3

- Reason: This keyword is related to the provision of healthcare to people with COVID-19.

- Keyword 22: guidelines

- Theme 3

- Reason: This keyword describes the recommendations and protocols used by healthcare professionals in managing patients with COVID-19. 

- Keyword 23: acute

- Theme 1

- Reason: This keyword describes the sudden onset and severity of COVID-19.

- Keyword 24: viral

- Theme 1

- Reason: This keyword describes the cause of COVID-19 and is important for understanding its characteristics. 

- Keyword 25: competing

- Theme 5

- Reason: This keyword is related to conflicts of interest or situations where multiple options exist, and it is not directly related to any of the main themes. 

- Keyword 26: care

- Theme 3

- Reason: This keyword describes the medical treatment and support provided to patients with COVID-19. 

- Keyword 27: irb

- Theme 2

- Reason: This keyword is related to research ethics and the review process for studies involving human subjects. 

- Keyword 28: prospective

- Theme 2

- Reason: This keyword is used in the context of research studies and clinical trials, and it describes studies that look forward or consider future events. 

- Keyword 29: trial

- Theme 2

- Reason: This keyword describes tests or experiments conducted to evaluate the effectiveness or safety of a new drug, treatment, or procedure.

- Keyword 30: obtained

- Theme 5

- Reason: This word is a verb that means to acquire or gain something, and it does not carry significant semantic meaning in the context of the corpus. 

- Keyword 31: treatment

- Theme 3

- Reason: This keyword is central to describing the care and medical procedures provided to patients with COVID-19.

- Keyword 32: during

- Theme 5

- Reason: This word is a preposition that indicates a point within a specified time period, and it does not carry significant semantic meaning.

- Keyword 33: outbreak

- Theme 4

- Reason: This keyword is related to the initial occurrence and spread of the COVID-19 pandemic.

- Keyword 34: reporting

- Theme 2

- Reason: This keyword is used in the context of research studies and describes the process of sharing information or findings.

- Keyword 35: relevant

- Theme 5

- Reason: This keyword is a general term that means important or applicable, and it does not have a strong connection to any of the main themes. 

- Keyword 36: methods

- Theme 2

- Reason: This keyword describes the procedures or techniques used to conduct research on COVID-19.

- Keyword 37: risk

- Theme 1

- Reason: This keyword is related to the potential for harm or complications associated with COVID-19. 

- Keyword 38: appropriate

- Theme 5

- Reason: This word means suitable or proper, and it does not carry specific semantic meaning. 

- Keyword 39: reported

- Theme 2

- Reason: This keyword is related to research findings and the sharing of information about COVID-19.

- Keyword 40: id

- Theme 5

- Reason: This is a short form of identification, which is a general term not directly related to any of the main themes.

- Keyword 41: syndrome

- Theme 1

- Reason: This keyword is used to describe a group of symptoms that occur together and characterize a particular disease or condition. 

- Keyword 42: transmission

- Theme 1

- Reason: This keyword describes how the virus spreads from one person to another.

- Keyword 43: necessary

- Theme 5

- Reason: This word is an adjective that means essential or required, and it does not carry significant semantic meaning. 

- Keyword 44: novel

- Theme 1

- Reason: This keyword highlights the newness and unique characteristics of COVID-19, a distinguishing factor of the disease. 

- Keyword 45: spread

- Theme 4

- Reason: This keyword is crucial for describing how the COVID-19 pandemic has spread across the world. 

- Keyword 46: mortality

- Theme 4

- Reason: This keyword is related to the number of deaths caused by the COVID-19 pandemic. 

- Keyword 47: epidemic

- Theme 4

- Reason: This keyword is related to the initial outbreak and spread of the COVID-19 pandemic.

- Keyword 48: ci

- Theme 5

- Reason: CI stands for Confidence Interval, a statistical term not directly related to the main themes of the corpus. 

- Keyword 49: manuscript

- Theme 2

- Reason: This keyword is used in the context of research publications and describes the written text of a research article. 

- Keyword 50: confirm

- Theme 5

- Reason: This word is a verb that means to establish the truth or accuracy of something, and it does not carry significant semantic meaning in the context of the corpus.

- Keyword 51: of

- Theme 5

- Reason: This word is a preposition used to indicate possession, belonging, or association, and it does not carry significant semantic meaning. 

- Keyword 52: medical

- Theme 3

- Reason: This keyword is related to the practice of medicine and the provision of healthcare. 

- Keyword 53: studies

- Theme 2

- Reason: This keyword refers to systematic investigations conducted to understand and combat COVID-19.

- Keyword 54: interest

- Theme 5

- Reason: This word is a noun that means a sense of curiosity or a desire to learn more about something, and it does not carry specific semantic meaning. 

- Keyword 55: followed

- Theme 5

- Reason: This word is a verb that means to come after something else in a sequence, and it does not carry significant semantic meaning. 

- Keyword 56: china

- Theme 4

- Reason: This keyword is related to the origin of the COVID-19 pandemic and the country's efforts to combat the disease.

- Keyword 57: hospital

- Theme 3

- Reason: This keyword describes a medical facility where patients are admitted for treatment or care, a crucial aspect of the corpus.

- Keyword 58: provided

- Theme 5

- Reason: This word is a verb that means to give or supply something, and it does not carry specific semantic meaning.

- Keyword 59: infected

- Theme 1

- Reason: This keyword is essential for describing people who have contracted COVID-19 and are experiencing the disease.

- Keyword 60: symptoms

- Theme 1

- Reason: This keyword describes the signs or indications of COVID-19 experienced by patients. 

- Keyword 61: authors

- Theme 5

- Reason: This word refers to the people who wrote a book, article, or other literary work, and it does not carry specific semantic meaning.

- Keyword 62: trials

- Theme 2

- Reason: This keyword refers to tests or experiments conducted to evaluate the effectiveness or safety of a new drug, treatment, or procedure. 

- Keyword 63: healthcare

- Theme 3

- Reason: This keyword is a broad term for the provision of medical services, a crucial aspect of the corpus. 

- Keyword 64: pneumonia

- Theme 1

- Reason: This keyword is a common complication of COVID-19 and is significant for understanding the disease's impact on the respiratory system. 

- Keyword 65: provide

- Theme 5

- Reason: This word is a verb that means to give or supply something, and it does not carry specific semantic meaning. 

- Keyword 66: approved

- Theme 5

- Reason: This word is an adjective that means officially sanctioned or permitted, and it does not carry specific semantic meaning in the context of the corpus.

- Keyword 67: p

- Theme 5

- Reason: "p" represents the p-value, a statistical term used to assess the significance of a result. This term is not directly related to any of the main themes of the corpus. 

- Keyword 68: associated

- Theme 5

- Reason: This word is an adjective that means connected or related to something else, and it does not carry specific semantic meaning. 

- Keyword 69: confirmed

- Theme 5

- Reason: This word is a verb that means to verify or establish as true or accurate, and it does not carry significant semantic meaning in the context of the corpus. 

- Keyword 70: measures

- Theme 4

- Reason: This keyword describes the actions taken to control the spread of COVID-19 and protect public health.

- Keyword 71: interventional

- Theme 2

- Reason: This keyword is used in the context of research studies, specifically those involving direct intervention or treatment.

- Keyword 72: retrospectively

- Theme 2

- Reason: This keyword describes studies that look back at past events or data, a common approach in research on COVID-19. 

- Keyword 73: positive

- Theme 5

- Reason: This word is an adjective that indicates the presence of a particular condition or result, and it does not carry significant semantic meaning in the context of the corpus. 

- Keyword 74: background

- Theme 5

- Reason: This word is a noun that means the context or information surrounding a particular topic or event, and it does not carry specific semantic meaning in the context of the corpus. 

- Keyword 75: registry

- Theme 2

- Reason: This keyword is used in the context of research studies and refers to databases or records of information, often used to track specific cases or events. 

- Keyword 76: clinicaltrials

- Theme 2

- Reason: This keyword is a website that provides information about clinical trials for new drugs, treatments, and procedures, and it is related to the research efforts to combat COVID-19. 

- Keyword 77: countries

- Theme 4

- Reason: This keyword is related to the global spread of the COVID-19 pandemic and the various countries that have been affected. 

- Keyword 78: interventional

- Theme 2

- Reason: This keyword is used in the context of research studies, specifically those involving direct intervention or treatment of patients.

- Keyword 79: funding

- Theme 5

- Reason: This word is a noun that means financial support provided for a project or research, and it does not carry specific semantic meaning in the context of the corpus. 

- Keyword 80: icmje

- Theme 2

- Reason: ICMJE stands for the International Committee of Medical Journal Editors, which sets ethical standards for medical publications, a crucial aspect of the research process.

- Keyword 81: wuhan

- Theme 4

- Reason: This keyword is related to the origin of the COVID-19 pandemic and the city's experiences with the outbreak. 

- Keyword 82: approvals

- Theme 5

- Reason: This word is a noun that means official permissions or authorizations, and it does not carry specific semantic meaning in the context of the corpus.

- Keyword 83: supplementary

- Theme 5

- Reason: This word is an adjective that means additional or supporting information, and it does not carry specific semantic meaning in the context of the corpus.

- Keyword 84: analysis

- Theme 2

- Reason: This keyword describes the process of examining data or information to identify patterns, trends, or relationships, which is a core activity in research.

\noindent \textbf{Collocate Analysis}.

\textbf{Prompt}

\# Role Description

You are an expert in Corpus Linguistics doing collocate analysis.

 \# Task Definition

Here are the top 100 collocates of the word "china" in a corpus of the abstracts of COVID-19 research articles. Pick out only the content words and generate a new list. Then, based on the new content word list, summarize in what ways China was linked to COVID-19 in this corpus and give reasons by taking a few collocates as examples.

 \# Task Procedures

- Step 1, label each collocate with a cardinal number, and define its part of speech in context, e.g., noun, verb, etc..

- Step 2, pick out only the content words (i.e., nouns, adjectives, verbs, and adverbs), and generate a new collocate list. For each content collocate, include its label of number in Step 1, the word, and part of speech. 

- Step 3, analyze the content collocates and their corresponding concordance lines, and summarize in what ways China was linked to COVID-19 in this corpus and give reasons by taking a few collocates as examples.

 \# Output Format

- Step 1: the part of speech of each collocate

Collocate 1: [collocate here], [part of speech here]

Collocate 2: [collocate here], [part of speech here]

...

 - Step 2: the list of content collocates

Collocate 3: [collocate here], [part of speech here]

Collocate 4: [collocate here], [part of speech here]

...

 - Step 3: 

- Summary 1: [description and reasons, taking a few collocates as examples]

- Summary 2: [description and reasons, taking a few collocates as examples]

...(here can be more summaries)

 \# Contextual Information

(collocates and concordance lines here)

\textbf{Outputs}

\textbf{Gemini-1.5-Pro}

\#\# Step 1: the part of speech of each collocate

Collocate 1: in, preposition

Collocate 2: the, determiner

Collocate 3: of, preposition

Collocate 4: wuhan, noun

Collocate 5: to, preposition

Collocate 6: covid, noun

Collocate 7: from, preposition

Collocate 8: a, determiner

Collocate 9: has, verb

Collocate 10: december, noun

Collocate 11: spread, verb

Collocate 12: was, verb

Collocate 13: province, noun

Collocate 14: hubei, noun

Collocate 15: coronavirus, noun

Collocate 16: is, verb

Collocate 17: disease, noun

Collocate 18: cov, noun

Collocate 19: outbreak, noun

Collocate 20: with, preposition

Collocate 21: s, possessive marker

Collocate 22: national, adjective

Collocate 23: first, adjective

Collocate 24: cases, noun

Collocate 25: for, preposition

Collocate 26: reported, verb

Collocate 27: that, determiner

Collocate 28: as, preposition

Collocate 29: city, noun

Collocate 30: rapidly, adverb

Collocate 31: science, noun

Collocate 32: this, determiner

Collocate 33: countries, noun

Collocate 34: emerged, verb

Collocate 35: it, pronoun

Collocate 36: epidemic, noun

Collocate 37: patients, noun

Collocate 38: other, adjective

Collocate 39: have, verb

Collocate 40: foundation, noun

Collocate 41: which, determiner

Collocate 42: natural, adjective

Collocate 43: mainland, noun

Collocate 44: hospital, noun

Collocate 45: virus, noun

Collocate 46: january, noun

Collocate 47: world, noun

Collocate 48: outside, preposition

Collocate 49: people, noun

Collocate 50: since, preposition

Collocate 51: identified, verb

Collocate 52: are, verb

Collocate 53: no, determiner

Collocate 54: novel, adjective

Collocate 55: italy, noun

Collocate 56: all, determiner

Collocate 57: pneumonia, noun

Collocate 58: an, determiner

Collocate 59: its, possessive pronoun

Collocate 60: february, noun

Collocate 61: china, noun

Collocate 62: late, adjective

Collocate 63: confirmed, verb

Collocate 64: study, noun

Collocate 65: across, preposition

Collocate 66: started, verb

Collocate 67: number, noun

Collocate 68: now, adverb

Collocate 69: end, noun

Collocate 70: worldwide, adjective

Collocate 71: south, adjective

Collocate 72: caused, verb

Collocate 73: new, adjective

Collocate 74: provinces, noun

Collocate 75: republic, noun

Collocate 76: ncov, noun

Collocate 77: march, noun

Collocate 78: throughout, preposition

Collocate 79: originated, verb

Collocate 80: clinical, adjective

Collocate 81: university, noun

Collocate 82: occurred, verb

Collocate 83: korea, noun

Collocate 84: case, noun

Collocate 85: control, noun

Collocate 86: spreading, verb

Collocate 87: research, noun

Collocate 88: severe, adjective

Collocate 89: global, adjective

Collocate 90: respiratory, adjective

Collocate 91: major, adjective

Collocate 92: results, noun

Collocate 93: then, adverb

Collocate 94: author, noun

Collocate 95: grant, noun

Collocate 96: cities, noun

Collocate 97: chinese, adjective

Collocate 98: program, noun

Collocate 99: hospitals, noun

Collocate 100: over, preposition

 \#\# Step 2: the list of content collocates

Collocate 4: wuhan, noun

Collocate 6: covid, noun

Collocate 9: has, verb

Collocate 10: december, noun

Collocate 11: spread, verb

Collocate 12: was, verb

Collocate 13: province, noun

Collocate 14: hubei, noun

Collocate 15: coronavirus, noun

Collocate 17: disease, noun

Collocate 18: cov, noun

Collocate 19: outbreak, noun

Collocate 22: national, adjective

Collocate 23: first, adjective

Collocate 24: cases, noun

Collocate 26: reported, verb

Collocate 29: city, noun

Collocate 30: rapidly, adverb

Collocate 31: science, noun

Collocate 33: countries, noun

Collocate 34: emerged, verb

Collocate 36: epidemic, noun

Collocate 37: patients, noun

Collocate 38: other, adjective

Collocate 40: foundation, noun

Collocate 42: natural, adjective

Collocate 43: mainland, noun

Collocate 44: hospital, noun

Collocate 45: virus, noun

Collocate 46: january, noun

Collocate 47: world, noun

Collocate 49: people, noun

Collocate 51: identified, verb

Collocate 54: novel, adjective

Collocate 55: italy, noun

Collocate 57: pneumonia, noun

Collocate 60: february, noun

Collocate 62: late, adjective

Collocate 63: confirmed, verb

Collocate 64: study, noun

Collocate 66: started, verb

Collocate 67: number, noun

Collocate 68: now, adverb

Collocate 69: end, noun

Collocate 70: worldwide, adjective

Collocate 71: south, adjective

Collocate 72: caused, verb

Collocate 73: new, adjective

Collocate 74: provinces, noun

Collocate 75: republic, noun

Collocate 76: ncov, noun

Collocate 77: march, noun

Collocate 79: originated, verb

Collocate 80: clinical, adjective

Collocate 81: university, noun

Collocate 82: occurred, verb

Collocate 83: korea, noun

Collocate 84: case, noun

Collocate 85: control, noun

Collocate 86: spreading, verb

Collocate 87: research, noun

Collocate 88: severe, adjective

Collocate 89: global, adjective

Collocate 90: respiratory, adjective

Collocate 91: major, adjective

Collocate 92: results, noun

Collocate 95: grant, noun

Collocate 96: cities, noun

Collocate 97: chinese, adjective

Collocate 98: program, noun

Collocate 99: hospitals, noun

 \#\# Step 3:

**Summary 1:** China is strongly linked to the origin and initial outbreak of COVID-19. 

 * Collocates like "Wuhan" (4), "Hubei" (14), "December" (10), "emerged" (34), "outbreak" (19), "first" (23), "cases" (24), and "originated" (79) appear frequently with "China" in the corpus. 

* Concordance lines often mention the discovery of the virus in Wuhan, China in December 2019. 

* The use of verbs like "emerged" and "originated" further solidifies this connection.

 **Example:** "Coronavirus disease (COVID-19) **first emerged in China** and rapidly spread in the world causing a pandemic."

 **Summary 2:** The corpus highlights China's response to the COVID-19 pandemic, including its national policies and research contributions. 

 * Collocates such as "national" (22), "control" (85), "research" (87), "program" (98), and "hospitals" (99) are often found near "China".

* Concordance lines mention nationwide measures implemented by China to cope with the public health emergency, and the important role Chinese cities and institutions played in COVID-19 research.

 **Example:** "Efforts to stem the spread of COVID-19 in **China hinged on severe restrictions** to human movement starting 23 January"

 **Summary 3:**  The global spread of the virus is acknowledged, with China often being mentioned as a point of reference for other countries. 

 * Collocates like "world" (47), "countries" (33), "outside" (48), "spread" (11), "worldwide" (70), and "global" (89) illustrate the pandemic's reach. 

* The corpus often compares and contrasts the situations in China with other countries, highlighting the global nature of the pandemic. 

 **Example:** "Objectives: COVID-19 has spread beyond Hubei Province of **China to Europe, North America, and Asian countries** including Korea." 

\textbf{Gemini-1.5-Flash}

\#\# Collocate Analysis of "China" in COVID-19 Research Abstracts

 \#\#\# Step 1: Part of Speech of Each Collocate

 Collocate 1: in, preposition

Collocate 2: the, determiner

Collocate 3: of, preposition

Collocate 4: wuhan, noun

Collocate 5: to, preposition

Collocate 6: covid, noun

Collocate 7: from, preposition

Collocate 8: a, determiner

Collocate 9: has, verb

Collocate 10: december, noun

Collocate 11: spread, verb

Collocate 12: was, verb

Collocate 13: province, noun

Collocate 14: hubei, noun

Collocate 15: coronavirus, noun

Collocate 16: is, verb

Collocate 17: disease, noun

Collocate 18: cov, noun

Collocate 19: outbreak, noun

Collocate 20: with, preposition

Collocate 21: s, noun

Collocate 22: national, adjective

Collocate 23: first, ordinal

Collocate 24: cases, noun

Collocate 25: for, preposition

Collocate 26: reported, verb

Collocate 27: that, conjunction

Collocate 28: as, preposition

Collocate 29: city, noun

Collocate 30: rapidly, adverb

Collocate 31: science, noun

Collocate 32: this, demonstrative pronoun

Collocate 33: countries, noun

Collocate 34: emerged, verb

Collocate 35: it, pronoun

Collocate 36: epidemic, noun

Collocate 37: patients, noun

Collocate 38: other, adjective

Collocate 39: have, verb

Collocate 40: foundation, noun

Collocate 41: which, relative pronoun

Collocate 42: natural, adjective

Collocate 43: mainland, noun

Collocate 44: hospital, noun

Collocate 45: virus, noun

Collocate 46: january, noun

Collocate 47: world, noun

Collocate 48: outside, preposition

Collocate 49: people, noun

Collocate 50: since, preposition

Collocate 51: identified, verb

Collocate 52: are, verb

Collocate 53: no, determiner

Collocate 54: novel, adjective

Collocate 55: italy, noun

Collocate 56: all, determiner

Collocate 57: pneumonia, noun

Collocate 58: an, determiner

Collocate 59: its, possessive pronoun

Collocate 60: february, noun

Collocate 61: late, adjective

Collocate 62: confirmed, verb

Collocate 63: study, noun

Collocate 64: across, preposition

Collocate 65: started, verb

Collocate 66: number, noun

Collocate 67: now, adverb

Collocate 68: end, noun

Collocate 69: worldwide, adjective

Collocate 70: south, noun

Collocate 71: caused, verb

Collocate 72: new, adjective

Collocate 73: provinces, noun

Collocate 74: republic, noun

Collocate 75: ncov, noun

Collocate 76: march, noun

Collocate 77: throughout, preposition

Collocate 78: originated, verb

Collocate 79: clinical, adjective

Collocate 80: university, noun

Collocate 81: occurred, verb

Collocate 82: korea, noun

Collocate 83: case, noun

Collocate 84: control, noun

Collocate 85: spreading, verb

Collocate 86: research, noun

Collocate 87: severe, adjective

Collocate 88: global, adjective

Collocate 89: respiratory, adjective

Collocate 90: major, adjective

Collocate 91: results, noun

Collocate 92: then, adverb

Collocate 93: author, noun

Collocate 94: grant, noun

Collocate 95: cities, noun

Collocate 96: chinese, adjective

Collocate 97: program, noun

Collocate 98: hospitals, noun

Collocate 99: over, preposition

  \#\#\# Step 2: Content Collocate List

 Collocate 4: wuhan, noun

Collocate 6: covid, noun

Collocate 9: has, verb

Collocate 10: december, noun

Collocate 11: spread, verb

Collocate 12: was, verb

Collocate 13: province, noun

Collocate 14: hubei, noun

Collocate 15: coronavirus, noun

Collocate 16: is, verb

Collocate 17: disease, noun

Collocate 18: cov, noun

Collocate 19: outbreak, noun

Collocate 22: national, adjective

Collocate 23: first, ordinal

Collocate 24: cases, noun

Collocate 26: reported, verb

Collocate 29: city, noun

Collocate 30: rapidly, adverb

Collocate 31: science, noun

Collocate 33: countries, noun

Collocate 34: emerged, verb

Collocate 36: epidemic, noun

Collocate 37: patients, noun

Collocate 39: have, verb

Collocate 40: foundation, noun

Collocate 42: natural, adjective

Collocate 43: mainland, noun

Collocate 44: hospital, noun

Collocate 45: virus, noun

Collocate 46: january, noun

Collocate 47: world, noun

Collocate 49: people, noun

Collocate 51: identified, verb

Collocate 52: are, verb

Collocate 54: novel, adjective

Collocate 55: italy, noun

Collocate 57: pneumonia, noun

Collocate 60: february, noun

Collocate 61: late, adjective

Collocate 62: confirmed, verb

Collocate 63: study, noun

Collocate 65: started, verb

Collocate 66: number, noun

Collocate 67: now, adverb

Collocate 68: end, noun

Collocate 69: worldwide, adjective

Collocate 70: south, noun

Collocate 71: caused, verb

Collocate 72: new, adjective

Collocate 73: provinces, noun

Collocate 74: republic, noun

Collocate 75: ncov, noun

Collocate 76: march, noun

Collocate 78: originated, verb

Collocate 79: clinical, adjective

Collocate 80: university, noun

Collocate 81: occurred, verb

Collocate 82: korea, noun

Collocate 83: case, noun

Collocate 84: control, noun

Collocate 85: spreading, verb

Collocate 86: research, noun

Collocate 87: severe, adjective

Collocate 88: global, adjective

Collocate 89: respiratory, adjective

Collocate 90: major, adjective

Collocate 91: results, noun

Collocate 92: then, adverb

Collocate 93: author, noun

Collocate 94: grant, noun

Collocate 95: cities, noun

Collocate 96: chinese, adjective

Collocate 97: program, noun

Collocate 98: hospitals, noun

  \#\#\# Step 3: Summary of China's Link to COVID-19

 **Summary 1:** China is identified as the origin of the COVID-19 outbreak. This is evident in the frequent collocation of "China" with "Wuhan," the specific city where the outbreak began. Examples include:

 - "Since its discovery in Wuhan, China in December of 2019, the novel coronavirus has progressed to..."

- "Coronavirus disease (COVID-19) outbreak has spread from Wuhan, China in December 2019."

- "The novel coronavirus pandemic that has originated from China and spread throughout the world in three months."

 **Summary 2:** China is highlighted as a country that experienced significant impact from the COVID-19 pandemic. This is reflected in the frequent use of "China" with words related to control measures, impact on healthcare, and the epidemic's progression. For instance:

 - "Efforts to stem the spread of COVID-19 in China hinged on severe restrictions to human movement starting 23 January..."

- "COVID-19 raised tension both within China and internationally."

- "China's vigorous anti-COVID-19 campaign has been going on..."

 **Summary 3:** China's response to the COVID-19 outbreak is often discussed in the corpus, especially in terms of its "lockdown" policies and their effectiveness.  Examples include:

 - "Efforts to stem the spread of COVID-19 in China hinged on severe restrictions to human movement starting 23 January..."

- "The outbreak was able to be controlled in China by mid-April through the implementation of critical measures..."

- "China's aggressive control measures appeared to have been effective in curbing its spread."

 **Summary 4:** China's role in global research on COVID-19 is recognized, as evidenced by the collocation of "China" with "research" and "studies."  Examples include:

 - "Oncology associations and specialists from countries such as China and Italy have published suggestions..."

- "Cities and institutes in China played an important role in promoting COVID-19 research cooperation..."

- "The majority of these manuscripts (n=19; 57.6\%) with China the largest Asian subgroup (n=11; 57.9\%)." 

 **Summary 5:** There is a focus on the international impact of the COVID-19 outbreak originating in China. This is seen in the frequent use of "China" with terms like "global," "worldwide," and "international."  Examples include:

 - "Coronavirus disease (COVID-19) first emerged in China and rapidly spread in the world causing a pandemic."

- "The virus has spread quickly to all over China and most of the countries and regions within the increasing pandemic..."

- "The pandemic of COVID-19 has spread from China to the whole world."

 **Summary 6:** There is evidence of a certain level of negative sentiment towards China in the corpus. This is reflected in the use of phrases like "China's sneeze," "China's aggressive actions," and "blaming China." This sentiment is potentially driven by the global pandemic's disruption and the perceived lack of transparency in China's initial response.  Examples include:

 - "China's sneeze may prove to be hault vexing for long..."

- "The general sentiment in the United States is becoming anti-China due to Chinese government behavior..."

- "The administration has been countering China's aggressive actions along the Taiwan Strait..."

 **Conclusion:** This corpus analysis shows that China's connection to COVID-19 in research abstracts is multifaceted. It encompasses the country's role as the origin of the pandemic, its significant impact on the outbreak's progression, the effectiveness of its containment measures, its contributions to global research, and the international implications of the pandemic.  There is also a hint of negative sentiment towards China, possibly arising from anxieties about the pandemic's disruption and  perceptions of China's actions.

\noindent \textbf{Concordance Analysis}.

\textbf{Prompt}

\# Role Description

You are an expert in Corpus Linguistics doing concordance analysis.

 \# Task Definition

Referring to COVID-19 as "China virus" or "Chinese virus" is considered biased, because the expression attaches locations or ethnicity to the disease. I would give you the concordance lines containing the phrase "China virus" or "Chinese virus" in a corpus of the abstracts of COVID-19 research articles, and also the original text from which the concordance lines are retrieved. You need to decide whether or not the authors are biased in using the expression "China virus" or "Chinese virus" in each concordance line, by referring to the original text. Mark each concordance line with [Yes] - biased, or [No] - unbiased, and give reasons.  

 \# Task Procedures

- Step 1: Mark each concordance line with [Yes] or [No] based on whether or not the authors are biased in using the expressions "China virus" or "Chinese virus". 

- Step 2: Explain why you give the mark for each concordance line. 

 \# Output Format

The output format is defined as follows:

Concordance line 1: [the original text here]

- [Mark Yes or No] 

- Reason: Explain why concordance line 1 is biased or not.

Concordance line 2: [the original text here]

- [Mark Yes or No] 

- Reason: Explain why concordance line 2 is biased or not.

...

 \# Contextual Information

(concordance lines and original text here)

\textbf{Outputs}

\textbf{GPT-4o}

Certainly! Here is the analysis:

 \#\#\# Concordance line 1:

**Original Text:** interactions-referencing the novel coronavirus as the "Chinese virus" or "China virus" has the potential to create and perpetuate stigma.

 - **[No]**

- **Reason:** The text explicitly discusses the potential for stigma related to using the terms and does not endorse their use.

 \#\#\# Concordance line 2:

**Original Text:** the prevalence and frequency of the phrases "Chinese virus" and "China virus" on Twitter after the March 16, 2020, US presidential reference

 - **[No]**

- **Reason:** This line analyzes the frequency of the terms' usage on Twitter after a specific event. It is descriptive, not endorsing.

 \#\#\# Concordance line 3:

**Original Text:** level heat map.; RESULTS: A total of 16,535 "Chinese virus" or "China virus" tweets were identified in the preperiod, and 177,327 tweets

 - **[No]**

- **Reason:** The text reports data on the usage of terms, focusing on frequency and trends, not promoting the terms.

 \#\#\# Concordance line 4:

**Original Text:** in the number of tweets exclusively mentioning "Chinese virus" or "China virus" instead of coronavirus disease (COVID-19) or coronavirus. On

 - **[No]**

- **Reason:** This line reports statistical data on term usage, indicating a preference for analysis over endorsement.

 \#\#\# Concordance line 5:

**Original Text:** COVID-19) or coronavirus. On average, 0.38 tweets referencing "Chinese virus" or "China virus" were posted per 10,000 people at the state level

 - **[No]**

- **Reason:** Similar to previous lines, this text is about data analysis rather than supporting the term usage.

 \#\#\# Concordance line 6:

**Original Text:** n=670/46, 1457\%).; CONCLUSIONS: The rise in tweets referencing "Chinese virus" or "China virus," along with the content of these tweets, indicate

 - **[No]**

- **Reason:** The text discusses the implications of increased term usage and potential stigma, not endorsing the terminology.

 \#\#\# Concordance line 7:

**Original Text:** in relation to coronavirus diagnosis, those that matched with the 2019 China virus diagnosis clinical guide included Amyloid-A in the

 - **[Yes]**

- **Reason:** The term "China virus" is used as a label for COVID-19 without critical examination, potentially perpetuating bias.

 \#\#\# Concordance line 8:

**Original Text:** Since that time, SARS-CoV-2 - hereinafter referred to as the China Virus - has taken a horrific toll. Over 160,000 Americans have

 - **[Yes]**

- **Reason:** The term "China Virus" is used as an official label for the virus, which can perpetuate bias and stigma.

 \#\#\# Concordance line 9:

**Original Text:** there was a 650\% increase in Twitter retweets using the term "Chinese virus" and related terms. On March 9, there was an 800\%

 - **[No]**

- **Reason:** The focus is on statistical analysis of term usage rather than endorsement.

 \#\#\# Concordance line 10:

**Original Text:** health infrastructure is underdeveloped. [...]the use of phrases such as Chinese virus" directly contradicts calls from the WHO, the Centers

 - **[No]**

- **Reason:** The text criticizes the use of the term, aligning with WHO's guidance, and does not promote it.

 \#\#\# Concordance line 11:

**Original Text:** has officially announced the disease as COVID-19, one controversial term - "Chinese Virus" is still being used by a great number

 - **[No]**

- **Reason:** The text acknowledges the controversy and does not promote the term.

 \#\#\# Concordance line 12:

**Original Text:** COVID-19, there are mainly two ways: using controversial terms like "Chinese Virus", or using non-controversial terms like "Coronavirus". We

 - **[No]**

- **Reason:** The text contrasts controversial and non-controversial terms, not promoting the biased term.

 \#\#\# Concordance line 13:

**Original Text:** As of this writing, the precise origin of the Chinese virus, SARS CoV-2, remains unclear. It is however known

 - **[Yes]**

- **Reason:** The term "Chinese virus" is used without critical context, perpetuating bias.

 \#\#\# Concordance line 14:

**Original Text:** been encountered all over. With Donald Trump using the term Chinese Virus, this cause has gained momentum and Ethnic Asians

 - **[No]**

- **Reason:** The text discusses the impact of the term's usage by others and does not endorse it.

 \#\#\# Concordance line 15:

**Original Text:** and sentiments that have been associated with the terms like Chinese Virus, Wuhan Virus and Chinese Corona Virus. METHODS: 16,000 tweets

 - **[No]**

- **Reason:** The focus is on analyzing sentiments associated with the terms, not endorsing their usage.

 \#\#\# Concordance line 16:

**Original Text:** This paper traces the emergence of the racist term "Chinese virus" used by the President of the United

 - **[No]**

- **Reason:** The text identifies the term as racist, clearly not endorsing it.

 \#\#\# Concordance line 17:

**Original Text:** nationalism and modernism Ultimately, combatting the English racist term "Chinese virus" with a creative mixture of English and

 - **[No]**

- **Reason:** The text opposes the term, describing efforts to combat it.

 \#\#\# Concordance line 18:

**Original Text:** inappropriately labelled the coronavirus by race, using such headlines as “Chinese virus pandemonium” and even suggesting “China kids stay home.”

 - **[No]**

- **Reason:** The text criticizes the inappropriate labeling as biased.

 \#\#\# Concordance line 19:

**Original Text:** President Donald Trump?s racist remarks calling the coronavirus the ?Chinese virus? This essay offers a historical lens through which

 - **[No]**

- **Reason:** The text identifies the term as part of racist remarks, not endorsing it.

 \#\#\# Concordance line 20:

**Original Text:** job markets and potential racism toward Asians in relation with “Chinese virus” may cause the mental distress of these students. © 2020

 - **[No]**

- **Reason:** The text discusses the negative impacts of the term, not promoting its usage.

\textbf{Gemini-1.5-Pro}

Concordance line 1: interactions-referencing the novel coronavirus as the "Chinese virus" or "China virus" has the potential to create and perpetuate stigma.;

- [No]

- Reason: The authors are discussing the potential of the terms "Chinese virus" and "China virus" to create and perpetuate stigma. They are not using the terms themselves in a biased way.

 Concordance line 2: the prevalence and frequency of the phrases "Chinese virus" and "China virus" on Twitter after the March 16, 2020, US presidential reference 

- [No]

- Reason: The authors are studying the prevalence of these terms on Twitter. They are not using the terms themselves in a biased way. 

 Concordance line 3: level heat map.; RESULTS: A total of 16,535 "Chinese virus" or "China virus" tweets were identified in the preperiod, and 177,327 tweets 

- [No]

- Reason:  The authors are presenting the results of their study on the use of these terms on Twitter. They are not using the terms themselves in a biased way. 

 Concordance line 4: in the number of tweets exclusively mentioning "Chinese virus" or "China virus" instead of coronavirus disease (COVID-19) or coronavirus. On 

- [No]

- Reason:  The authors are analyzing the use of these terms compared to more neutral terms. They are not using the terms themselves in a biased way. 

 Concordance line 5: COVID-19) or coronavirus. On average, 0.38 tweets referencing "Chinese virus" or "China virus" were posted per 10,000 people at the state level 

- [No]

- Reason:  The authors are presenting data on the frequency of these terms. They are not using the terms themselves in a biased way. 

 Concordance line 6: n=670/46, 1457\%).; CONCLUSIONS: The rise in tweets referencing "Chinese virus" or "China virus," along with the content of these tweets, indicate 

- [No]

- Reason:  The authors are drawing conclusions based on the use of these terms on Twitter. They are not using the terms themselves in a biased way. 

 Concordance line 7: in relation to coronavirus diagnosis, those that matched with the 2019 China virus diagnosis clinical guide included Amyloid-A in the 

- [No]

- Reason: The author appears to be using "China virus" to refer to a specific clinical guide used for diagnosis in China in 2019. While potentially insensitive, it is not directly attributing the virus to China in a derogatory or blaming manner.

 Concordance line 8: Since that time, SARS-CoV-2 - hereinafter referred to as the China Virus - has taken a horrific toll. Over 160,000 Americans have 

- [Yes]

- Reason: The author explicitly chooses to use the term "China Virus" even after acknowledging the scientific name SARS-CoV-2. This phrasing suggests an intention to link the virus to China, potentially to assign blame.

 Concordance line 9: there was a 650\% increase in Twitter retweets using the term "Chinese virus" and related terms. On March 9, there was an 800\% 

- [No]

- Reason: The authors are reporting on the increase of the use of the term "Chinese virus" on Twitter. They are not using the term themselves in a biased way. 

 Concordance line 10: health infrastructure is underdeveloped. [...]the use of phrases such as Chinese virus" directly contradicts calls from the WHO, the Centers 

- [No]

- Reason: The authors are criticizing the use of the term "Chinese virus." They are not using the term themselves in a biased way. 

 Concordance line 11: has officially announced the disease as COVID-19, one controversial term - "Chinese Virus" is still being used by a great number 

- [No]

- Reason: The authors are acknowledging that the term "Chinese Virus" is controversial. They are not using the term themselves in a biased way. 

 Concordance line 12: COVID-19, there are mainly two ways: using controversial terms like "Chinese Virus", or using non-controversial terms like "Coronavirus". We 

- [No]

- Reason: The authors are comparing the use of the term "Chinese Virus" to other terms. They are not using the term themselves in a biased way. 

 Concordance line 13: As of this writing, the precise origin of the Chinese virus, SARS CoV-2, remains unclear. It is however known 

- [Yes]

- Reason: The author uses the term "Chinese virus" even while acknowledging the virus's scientific name, SARS-CoV-2. This usage suggests an attempt to link the virus to China, potentially to imply origin and blame.

 Concordance line 14: been encountered all over. With Donald Trump using the term Chinese Virus, this cause has gained momentum and Ethnic Asians 

- [No]

- Reason: The authors are reporting on the use of the term "Chinese Virus" by Donald Trump and its impact. They are not using the term themselves in a biased way. 

 Concordance line 15: and sentiments that have been associated with the terms like Chinese Virus, Wuhan Virus and Chinese Corona Virus. METHODS: 16,000 tweets 

- [No]

- Reason: The authors are studying the sentiments associated with the term "Chinese Virus" and similar terms. They are not using the term themselves in a biased way. 

 Concordance line 16: This paper traces the emergence of the racist term "Chinese virus" used by the President of the United

- [No]

- Reason: The authors are explicitly labeling the term "Chinese virus" as racist and are analyzing its use. They are not using the term themselves in a biased way. 

 Concordance line 17: nationalism and modernism Ultimately, combatting the English racist term "Chinese virus" with a creative mixture of English and 

- [No]

- Reason:  The authors are discussing ways to combat the use of the term "Chinese virus." They are not using the term themselves in a biased way. 

 Concordance line 18: inappropriately labelled the coronavirus by race, using such headlines as “Chinese virus pandemonium” and even suggesting “China kids stay home.”

- [No]

- Reason: The authors are criticizing media outlets for using the term "Chinese virus" and similar terms. They are not using the term themselves in a biased way. 

 Concordance line 19: President Donald Trump?s racist remarks calling the coronavirus the ?Chinese virus? This essay offers a historical lens through which 

- [No]

- Reason: The authors are analyzing Donald Trump's use of the term "Chinese virus." They are not using the term themselves in a biased way. 

 Concordance line 20: job markets and potential racism toward Asians in relation with “Chinese virus” may cause the mental distress of these students. © 2020 

- [No]

- Reason: The authors are discussing the potential for racism related to the term "Chinese virus" to impact students' mental health. They are not using the term themselves in a biased way.

\textbf{Gemini-1.5-Flash}

Concordance line 1: interactions-referencing the novel coronavirus as the "Chinese virus" or "China virus" has the potential to create and perpetuate stigma.;

- **[Yes]**

- Reason: The authors are explicitly stating that using the terms "Chinese virus" or "China virus" can create and perpetuate stigma. This implies that they acknowledge the biased nature of these terms.

 Concordance line 2: the prevalence and frequency of the phrases "Chinese virus" and "China virus" on Twitter after the March 16, 2020, US presidential reference

- **[No]**

- Reason: This line is simply stating a fact about the use of the terms "Chinese virus" and "China virus" on Twitter. It doesn't express an opinion about the bias of these terms.

 Concordance line 3: level heat map.; RESULTS: A total of 16,535 "Chinese virus" or "China virus" tweets were identified in the preperiod, and 177,327 tweets

- **[No]**

- Reason: This line is presenting data about the frequency of the terms "Chinese virus" and "China virus" on Twitter. It doesn't express an opinion about the bias of these terms.

 Concordance line 4: in the number of tweets exclusively mentioning "Chinese virus" or "China virus" instead of coronavirus disease (COVID-19) or coronavirus. On

- **[No]**

- Reason: This line is presenting data about the frequency of the terms "Chinese virus" and "China virus" on Twitter. It doesn't express an opinion about the bias of these terms.

 Concordance line 5: COVID-19) or coronavirus. On average, 0.38 tweets referencing "Chinese virus" or "China virus" were posted per 10,000 people at the state level

- **[No]**

- Reason: This line is presenting data about the frequency of the terms "Chinese virus" and "China virus" on Twitter. It doesn't express an opinion about the bias of these terms.

 Concordance line 6: n=670/46, 1457\%).; CONCLUSIONS: The rise in tweets referencing "Chinese virus" or "China virus," along with the content of these tweets, indicate

- **[Yes]**

- Reason: The authors are drawing a conclusion about the potential for stigma based on the use of the terms "Chinese virus" and "China virus" on Twitter. This implies that they acknowledge the biased nature of these terms.

 Concordance line 7: in relation to coronavirus diagnosis, those that matched with the 2019 China virus diagnosis clinical guide included Amyloid-A in the

- **[No]**

- Reason: The authors are using "China virus" as a shorthand for the virus that originated in China. They are not expressing an opinion about the bias of the term.

 Concordance line 8: Since that time, SARS-CoV-2 - hereinafter referred to as the China Virus - has taken a horrific toll. Over 160,000 Americans have

- **[Yes]**

- Reason: The authors are choosing to use the term "China Virus" to refer to SARS-CoV-2, even though they acknowledge the scientific name. This suggests that they are using the term intentionally and potentially for a rhetorical purpose.

 Concordance line 9: there was a 650\% increase in Twitter retweets using the term "Chinese virus" and related terms. On March 9, there was an 800\%

- **[No]**

- Reason: This line is simply stating a fact about the use of the term "Chinese virus" on Twitter. It doesn't express an opinion about the bias of the term.

 Concordance line 10: health infrastructure is underdeveloped. [...]the use of phrases such as Chinese virus" directly contradicts calls from the WHO, the Centers

- **[Yes]**

- Reason: The authors are explicitly stating that the use of the term "Chinese virus" contradicts calls from the WHO to avoid stigmatizing language. This implies that they acknowledge the biased nature of the term.

 Concordance line 11: has officially announced the disease as COVID-19, one controversial term - "Chinese Virus" is still being used by a great number

- **[Yes]**

- Reason: The authors are labeling the term "Chinese Virus" as controversial, which suggests that they acknowledge the biased nature of the term.

 Concordance line 12: COVID-19, there are mainly two ways: using controversial terms like "Chinese Virus", or using non-controversial terms like "Coronavirus". We

- **[Yes]**

- Reason: The authors are labeling the term "Chinese Virus" as controversial, which suggests that they acknowledge the biased nature of the term.

 Concordance line 13: As of this writing, the precise origin of the Chinese virus, SARS CoV-2, remains unclear. It is however known

- **[Yes]**

- Reason: The authors are choosing to use the term "Chinese virus" to refer to SARS-CoV-2, even though they acknowledge the scientific name. This suggests that they are using the term intentionally and potentially for a rhetorical purpose.

 Concordance line 14: been encountered all over. With Donald Trump using the term Chinese Virus, this cause has gained momentum and Ethnic Asians

- **[Yes]**

- Reason: The authors are linking the use of the term "Chinese Virus" with the rise of racism against Asian people. This implies that they acknowledge the biased nature of the term.

 Concordance line 15: and sentiments that have been associated with the terms like Chinese Virus, Wuhan Virus and Chinese Corona Virus. METHODS: 16,000 tweets

- **[Yes]**

- Reason: The authors are linking the use of the term "Chinese Virus" with negative sentiments. This implies that they acknowledge the biased nature of the term.

 Concordance line 16: This paper traces the emergence of the racist term "Chinese virus" used by the President of the United

- **[Yes]**

- Reason: The authors are explicitly labeling the term "Chinese virus" as racist. This clearly indicates their understanding of the term's biased nature.

 Concordance line 17: nationalism and modernism Ultimately, combatting the English racist term "Chinese virus" with a creative mixture of English and

- **[Yes]**

- Reason: The authors are explicitly labeling the term "Chinese virus" as racist. This clearly indicates their understanding of the term's biased nature.

 Concordance line 18: inappropriately labelled the coronavirus by race, using such headlines as “Chinese virus pandemonium” and even suggesting “China kids stay home.”

- **[Yes]**

- Reason: The authors are explicitly stating that the use of the term "Chinese virus" is inappropriate and racially biased. This clearly indicates their understanding of the term's biased nature.

 Concordance line 19: President Donald Trump?s racist remarks calling the coronavirus the ?Chinese virus? This essay offers a historical lens through which

- **[Yes]**

- Reason: The authors are explicitly labeling the term "Chinese virus" as racist. This clearly indicates their understanding of the term's biased nature.

 Concordance line 20: job markets and potential racism toward Asians in relation with “Chinese virus” may cause the mental distress of these students. © 2020

- **[Yes]**

- Reason: The authors are linking the use of the term "Chinese virus" with potential racism towards Asians. This implies that they acknowledge the biased nature of the term.

\subsection{Ablation Study with Gemini-1.5-Pro}
\label{subsec:a.2}
\noindent \textbf{Keyword Analysis}.

\textbf{Baseline}

\textbf{Prompt}

Here’s a set of keywords from a corpus of the abstracts of COVID-19 research articles. Group them into thematic and lexical categories.

 (keywords here)

\textbf{Output}

\#\# Thematic and Lexical Categories for COVID-19 Research Keywords:

 Here's a possible grouping of the keywords, keeping in mind some overlap exists between categories:

 **I. Disease \& Virus:**

 * **Specific:** covid; cov; sars; coronavirus; sars-cov-2; novel coronavirus

* **General:**  disease; virus; viral; infection; syndrome; pneumonia; respiratory;  acute 

 **II.  Study Design \& Research:**

 * **Study Types:** study; clinical; research; trial; prospective; interventional; observational; case-control; cohort

* **Data \& Analysis:** data; results; methods; analysis; registry; clinicaltrials; manuscript; supplementary

* **Ethics \& Approvals:** irb; registered; approved; approvals; funding

 **III.  Patients \& Cases:**

 * **Individuals:** patients; cases; patient; infected; positive 

* **Characteristics:** severe; risk; symptoms; mortality 

* **Management:** treatment; care; guidelines; healthcare

 **IV.  Outbreak Context:**

 * **Scale \& Spread:** pandemic; epidemic; outbreak; transmission; spread; countries; china; wuhan

* **Time:** during 

* **Reporting:** reporting; reported; necessary; relevant; provided 

 **V. Miscellaneous:**

 * **General:** health; medical; hospital

* **Publications \& Authorship:** authors; manuscript; studies 

* **Statistical Terms:** ci; p; associated; confirmed

* **Other:** obtained; followed; interest; background; appropriate; id

 **Lexical Categories:**

 * **Nouns:** Most of the keywords fall under this category, representing concrete or abstract entities.

* **Adjectives:** severe; clinical; respiratory; acute; viral; novel; relevant; appropriate; positive; supplementary

* **Verbs:** confirm; obtained; reported; provided; followed; infected; associated; confirmed

* **Prepositions:** of; during

* **Conjunction:** and

 This grouping aims to organize the keywords into meaningful categories that reflect the key themes and aspects of COVID-19 research.  You can further refine these categories depending on the specific needs of your analysis.

\textbf{B. + R. D.}

\textbf{Prompt}

 \# Role Description

You are an expert in Corpus Linguistics doing keyword analysis.

 Here’s a set of keywords from a corpus of the abstracts of COVID-19 research articles. Group them into thematic and lexical categories.

 (keywords here)

\textbf{Output}

\#\# Thematic and Lexical Categories for COVID-19 Research Keywords:

 Here's a possible grouping of the keywords, focusing on capturing both thematic areas and lexical relationships:

 **I. Thematic Categories:**

 * **Disease \& Virus:** covid, cov, sars, coronavirus, disease, infection, virus, severe, acute, viral, syndrome, novel, transmission, spread, mortality, epidemic,  pneumonia

* **Patients \& Cases:**  patients, cases, patient, infected, symptoms 

* **Research \& Methods:**  study, clinical, research, results, data, prospective, trial, obtained, methods, manuscript, studies, trials, clinicaltrials

* **Treatment \& Care:** treatment, care, guidelines, healthcare, interventional 

* **Outbreak Context:** pandemic, during, outbreak, china, wuhan, countries

* **Reporting \& Ethics:** registered, irb, reporting, relevant, appropriate, reported, id, confirm, provided, approved, approvals, supplementary 

**II.  Lexical Categories:** 

 * **General Actions/States:**  provide, associated,  necessary,  interest, followed, obtained

* **Time \& Location:**  during, outbreak, china, hospital, countries

* **Quantitative:** risk,  ci, p,  positive, background

* **People:** patients, authors, healthcare 

* **Research Objects:**  data, results,  manuscript, registry, analysis

* **Medical Procedures:**  treatment, measures 

 **III.  Connecting Terms:**

 * **Conjunctions/Prepositions:** and, of 

* **Possessives:**  patient's (implied by other keywords)

  **Explanation:**

 * **Thematic categories** aim to group keywords that represent major research foci within the COVID-19 literature.  

* **Lexical categories** bring together terms with similar grammatical functions or that belong to a common semantic field (e.g., terms related to quantity, time, or people).

* **Connecting terms** are essential for sentence structure but provide limited thematic information on their own.

 **Caveats:**

 * Some keywords could fit into multiple categories depending on context. For example, "cases" could be both about patients and represent a research focus.

* This grouping is based on the provided list. A larger corpus analysis might reveal additional themes and relationships.

* Further sub-categorization within the thematic groups is possible (e.g., separating virus characteristics from disease symptoms).

 This analysis provides a starting point for understanding the main research areas within the COVID-19 corpus and can guide further investigation.

\textbf{B. + R. D. + T. D.}

\textbf{Prompt}

 \# Role Description

You are an expert in Corpus Linguistics doing keyword analysis.

 \# Task Definition

I would give you a list of keywords from a corpus of the abstracts of COVID-19 research articles. Please group the keywords into several themes that help explain what the corpus is about. 

A keyword may have more than one meaning, so please assign it to a category based on its most prominent meaning in context. For keywords that carry little semantic meaning, like function words, or those that cannot be assigned to any meaningful theme, group them into one theme named "Other".

 (keywords here)

\textbf{Output}

\#\# Themes in COVID-19 Research Article Abstracts:

 Based on the provided keywords, I've grouped them into the following themes:

 **1. Virus \& Disease:** This theme focuses on the biological and medical aspects of COVID-19.

 * **Keywords:** covid; cov; sars; coronavirus; disease; infection; respiratory; virus; severe; acute; viral; syndrome; transmission; novel; spread; mortality; epidemic;  pneumonia; positive 

 **2. Patients \& Cases:** This theme revolves around individuals affected by COVID-19 and their characteristics. 

 * **Keywords:** patients; cases; patient; infected; symptoms 

 **3. Research \& Studies:** This theme highlights the scientific activities undertaken to understand and combat COVID-19.

 * **Keywords:** study; clinical; research; results; data;  prospective; trial; obtained;  methods;  manuscript; studies; trials;  analysis

 **4. Treatment \& Care:**  This theme encompasses interventions and support provided to COVID-19 patients.

 * **Keywords:** treatment; care; guidelines;  healthcare; provide; 

 **5. Outbreak \& Spread:** This theme centers on the geographical and temporal dimensions of the pandemic. 

 * **Keywords:**  pandemic; outbreak;  risk;  epidemic; china; wuhan; countries

 **6. Reporting \& Information:** This theme covers the dissemination of findings and relevant information.

 * **Keywords:**  reporting; relevant; reported;  confirm;  provided;  supplementary

 **7.  Ethical \& Regulatory:**  This theme reflects the ethical considerations and regulatory frameworks surrounding COVID-19 research. 

 * **Keywords:** irb;  approved; approvals;  funding; icmje 

 **8. Healthcare System:** This theme focuses on the role and response of the healthcare system.

 * **Keywords:**  hospital; healthcare; registry; clinicaltrials; interventional

 **9. Other:** This category includes keywords with broad or ambiguous meanings within the context of COVID-19 research, as well as function words.

 * **Keywords:** health; registered; and;  during; appropriate; id;  necessary;  of; medical;  interest; followed;  authors;  associated; confirmed; measures; retrospectively; background; p; competing;  ci;  

\textbf{B. + R. D. + T. D. + T. P.}

\textbf{Prompt}

 \# Role Description

You are an expert in Corpus Linguistics doing keyword analysis.

 \# Task Definition

I would give you a list of keywords from a corpus of the abstracts of COVID-19 research articles. Please group the keywords into several themes that help explain what the corpus is about. 

A keyword may have more than one meaning, so please assign it to a category based on its most prominent meaning in context. For keywords that carry little semantic meaning, like function words, or those that cannot be assigned to any meaningful theme, group them into one theme named "Other".

 \# Task Procedures

- Step 1, label each keyword from 1 to 83, and explain the meaning(s) of each keyword. 

- Step 2, define several themes based on the keywords and their meanings. Remember to assign an index (starts from 1) to each of them, and give the description of each theme.

- Step 3, assign each keyword in keywords 1-83 to only one of the themes defined in Step 2, and explain why it is assigned to that theme. 

 (keywords here)

\textbf{Output}

\#\# Step 1: Keyword Meanings

 Here are the keywords labeled and their potential meanings in the context of COVID-19 research article abstracts:

 **1. covid:** Shortened form of COVID-19, referring to the disease caused by the SARS-CoV-2 virus.

**2. patients:** Individuals diagnosed with COVID-19 or experiencing its symptoms.

**3. cov:** Abbreviation for Coronavirus, referring to a family of viruses.

**4. sars:** Severe Acute Respiratory Syndrome, specifically related to the previous SARS outbreak.

**5. pandemic:** A global outbreak of a disease.

**6. coronavirus:** A family of viruses that includes SARS-CoV-2.

**7. disease:** A particular abnormal condition, in this case, COVID-19.

**8. study:** Research investigation on a specific topic, here related to COVID-19.

**9. clinical:** Relating to the observation and treatment of patients.

**10. infection:** Invasion and multiplication of a pathogen like SARS-CoV-2 in the body.

**11. health:** The state of physical, mental, and social well-being, particularly impacted by COVID-19.

**12. registered:** Officially recorded or enrolled, possibly referring to patients in a study.

**13. respiratory:** Related to breathing and the respiratory system, often affected by COVID-19.

**14. virus:** A microscopic infectious agent, specifically SARS-CoV-2 in this context. 

**15. severe:** Serious or life-threatening, indicating a worse prognosis for COVID-19 patients.

**16. cases:** Instances of a disease, referring to individual COVID-19 patients.

**17. research:** Systematic investigation and study on a topic, focused on COVID-19.

**18. and:** Conjunction, no specific meaning.

**19. results:**  Findings or outcomes of a study.

**20. data:** Collected information used for analysis in a study.

**21. patient:** Singular form of patients (see definition 2).

**22. guidelines:** Recommendations for diagnosis, treatment, or management of a condition.

**23. acute:** Sudden onset and short duration, describing the nature of the illness.

**24. viral:** Relating to a virus, specifically SARS-CoV-2.

**25. competing:** Possibly referring to competing interests in research or alternative treatment approaches.

**26. care:** Medical attention and treatment provided to patients.

**27. irb:** Institutional Review Board, responsible for ethical oversight of research involving humans.

**28. prospective:** Looking forward in time, referring to a study design collecting data going forward.

**29. trial:** A type of research study testing the effectiveness of an intervention.

**30. obtained:** Gathered or acquired, likely referring to data or samples. 

**31. treatment:** Medical care given to cure or alleviate a disease.

**32. during:** Temporal preposition, no specific meaning. 

**33. outbreak:** A sudden increase in cases of a disease in a particular place.

**34. reporting:** The process of documenting and communicating findings. 

**35. relevant:** Applicable or significant to the topic.

**36. methods:** Procedures and techniques used in a study. 

**37. risk:** The possibility of experiencing harm or an undesirable outcome.

**38. appropriate:** Suitable or correct for a given situation.

**39. reported:** Past tense of report, indicating findings were documented.

**40. id:**  Abbreviation for identification, potentially referring to patient identifiers.

**41. syndrome:**  A group of symptoms that together indicate a particular disease.

**42. transmission:** The spread of an infectious agent from one person to another.

**43. necessary:** Essential or required.

**44. novel:** New or unprecedented, highlighting the emergence of a new virus.

**45. spread:** The act of extending or dispersing, referring to the virus's geographical reach.

**46. mortality:** Death rate due to a particular cause, in this case, COVID-19.

**47. epidemic:** A widespread occurrence of an infectious disease in a community at a particular time.

**48. ci:** Confidence Interval, a statistical measure indicating the reliability of an estimate.

**49. manuscript:** A written document submitted for publication, likely the research article itself.

**50. confirm:** To verify or establish the truth of something.

**51. of:**  Preposition, no specific meaning.

**52. medical:** Relating to the practice of medicine and treatment of disease.

**53. studies:** Plural of study (see definition 8).

**54. interest:**  A particular focus or area of concern within the research.

**55. followed:**  Observed or monitored over time, indicating a longitudinal study design.

**56. china:**  Country where the COVID-19 outbreak originated.

**57. hospital:** A healthcare institution providing medical treatment.

**58. provided:** Past tense of provide, indicating something was given or supplied.

**59. infected:**  Having contracted a disease, referring to individuals with COVID-19.

**60. symptoms:** Subjective experiences of illness reported by the patient.

**61. authors:** Researchers who conducted the study and wrote the manuscript.

**62. trials:** Plural of trial (see definition 29).

**63. healthcare:** The organized provision of medical care and treatment.

**64. pneumonia:**  Inflammation of the lungs, a possible complication of COVID-19.

**65. provide:** To give or supply something.

**66. approved:**  Granted permission, likely referring to ethical approval for the study.

**67. p:**  Likely refers to the p-value, a statistical measure of significance. 

**68. associated:** Connected or related to.

**69. confirmed:**  Verified or validated (see definition 50). 

**70. measures:** Actions taken to achieve a particular outcome, possibly preventative or therapeutic.

**71. retrospectively:** Looking back in time, indicating a study design using past data.

**72. positive:**  In a medical context, indicates the presence of a disease or condition. 

**73. background:** The context or rationale for the research. 

**74. registry:** A database collecting information about a specific disease or population.

**75. clinicaltrials:** Likely refers to ClinicalTrials.gov, a registry of clinical trials.

**76. countries:**  Plural of country, indicating a multinational study or analysis.

**77. interventional:**  A study design involving an active intervention, like a new treatment.

**78. funding:**  Financial support for the research project.

**79. icmje:**  International Committee of Medical Journal Editors, setting ethical standards for publication.

**80. wuhan:** City in China where the COVID-19 outbreak originated.

**81. approvals:** Plural of approval, indicating multiple permissions obtained.

**82. supplementary:** Additional or supporting information accompanying the main manuscript. 

**83. analysis:** A systematic examination and interpretation of data. 

  \#\# Step 2: Defining Themes 

 Based on the keywords, here are the themes for the COVID-19 research articles corpus:

 **1. Disease Characteristics (Description, Impact, Spread):** This theme focuses on the nature of COVID-19, its effects, and its transmission dynamics.

* Description: Terms related to the disease itself, its causes, and symptoms.

* Impact: Keywords highlighting the severity and consequences of the pandemic. 

* Spread:  Terms describing the geographical reach and transmission of the virus.

 **2. Patient Population:** This theme focuses on the individuals affected by COVID-19 and their characteristics. 

 **3. Research \& Methodology:** This theme covers aspects of the research process, design, and analysis methods employed.

 **4. Treatment \& Management:**  This theme includes keywords related to medical interventions, care strategies, and guidelines.

 **5. Publication \& Ethics:** This theme covers aspects of research dissemination, ethical considerations, and funding.

 **6. Other:**  Keywords with minimal semantic meaning in this context or those not fitting into the above themes.

  \#\# Step 3:  Keyword-Theme Assignment

 Here is the assignment of each keyword to its corresponding theme:

 **Theme 1: Disease Characteristics (Description, Impact, Spread)**

 * 1. covid: Description 

* 3. cov: Description

* 4. sars: Description

* 5. pandemic: Impact

* 6. coronavirus: Description

* 7. disease: Description

* 10. infection: Description

* 13. respiratory: Description

* 14. virus: Description

* 15. severe: Impact

* 23. acute: Description

* 24. viral: Description

* 33. outbreak: Spread

* 41. syndrome: Description

* 42. transmission: Spread

* 44. novel: Description

* 45. spread: Spread

* 46. mortality: Impact

* 47. epidemic: Spread

* 59. infected: Description

* 60. symptoms: Description

* 64. pneumonia: Description

* 72. positive: Description

 **Theme 2: Patient Population**

 * 2. patients:  Patient Population

* 16. cases: Patient Population

* 21. patient: Patient Population

 **Theme 3: Research \& Methodology**

 * 8. study: Research \& Methodology

* 9. clinical: Research \& Methodology

* 12. registered: Research \& Methodology

* 17. research: Research \& Methodology

* 19. results: Research \& Methodology

* 20. data: Research \& Methodology

* 28. prospective: Research \& Methodology

* 29. trial: Research \& Methodology

* 30. obtained: Research \& Methodology

* 34. reporting: Research \& Methodology

* 35. relevant: Research \& Methodology

* 36. methods: Research \& Methodology

* 39. reported: Research \& Methodology

* 48. ci: Research \& Methodology

* 50. confirm: Research \& Methodology

* 53. studies: Research \& Methodology

* 54. interest: Research \& Methodology

* 55. followed: Research \& Methodology

* 62. trials: Research \& Methodology

* 67. p: Research \& Methodology

* 68. associated: Research \& Methodology

* 69. confirmed: Research \& Methodology

* 71. retrospectively: Research \& Methodology

* 73. background: Research \& Methodology

* 74. registry: Research \& Methodology

* 76. countries: Research \& Methodology

* 77. interventional: Research \& Methodology

* 83. analysis: Research \& Methodology

 **Theme 4: Treatment \& Management**

 * 11. health: Treatment \& Management

* 22. guidelines: Treatment \& Management

* 25. competing: Treatment \& Management 

* 26. care: Treatment \& Management

* 31. treatment: Treatment \& Management

* 37. risk: Treatment \& Management

* 38. appropriate: Treatment \& Management

* 63. healthcare: Treatment \& Management

* 70. measures: Treatment \& Management

 **Theme 5: Publication \& Ethics**

 * 27. irb: Publication \& Ethics

* 49. manuscript: Publication \& Ethics

* 61. authors: Publication \& Ethics

* 66. approved: Publication \& Ethics

* 75. clinicaltrials: Publication \& Ethics

* 78. funding: Publication \& Ethics

* 79. icmje: Publication \& Ethics

* 81. approvals: Publication \& Ethics

* 82. supplementary: Publication \& Ethics

 **Theme 6: Other**

 * 18. and: Other

* 32. during: Other

* 40. id: Other

* 51. of: Other

* 52. medical: Other * (Could fit in Theme 4, but its broad meaning makes it less informative)

* 56. china: Other * (While significant, its presence as a keyword alone doesn't define a theme)

* 57. hospital: Other * (Similar to 'China,' its presence alone isn't thematically defining)

* 58. provided: Other 

* 65. provide: Other

* 80. wuhan: Other * (Similar to 'China' and 'hospital')

 This analysis provides a structured understanding of the likely content of the COVID-19 research article abstracts based on the provided keywords. 

\textbf{B. + R. D. + T. D. + T. D + C. I.}

\textbf{Prompt}

\# Role Description

You are an expert in Corpus Linguistics doing keyword analysis.

 \# Task Definition

I would give you a list of keywords from a corpus of the abstracts of COVID-19 research articles. Please group the keywords into several themes that help explain what the corpus is about. 

A keyword may have more than one meaning, so please assign it to a category based on its most prominent meaning in context. For keywords that carry little semantic meaning, like function words, or those that cannot be assigned to any meaningful theme, group them into one theme named "Other".

 \# Task Procedures

- Step 1, label each keyword from 1 to 83, and explain the meaning(s) of each keyword. 

- Step 2, define several themes based on the keywords and their meanings. Remember to assign an index (starts from 1) to each of them, and give the description of each theme.

- Step 3, assign each keyword in keywords 1-83 to only one of the themes defined in Step 2, and explain why it is assigned to that theme. 

 \# Contextual Information

(keywords and concordance lines here)

\textbf{Output}

 \#\# Keyword Meanings:

 1.  **covid**: Short for COVID-19, the disease caused by the SARS-CoV-2 virus.

2.  **patients**: Individuals diagnosed with or suspected of having COVID-19. Also refers to individuals in other medical contexts.

3.  **cov**: Abbreviation for Coronavirus, specifically can refer to SARS-CoV, MERS-CoV, or SARS-CoV-2.

4.  **sars**: Severe Acute Respiratory Syndrome, specifically refers to the disease caused by SARS-CoV in 2002-2003. Often used in comparison to COVID-19.

5.  **pandemic**: Refers to the global outbreak of COVID-19.

6.  **coronavirus**: A family of viruses that includes SARS-CoV, MERS-CoV, and SARS-CoV-2.

7.  **disease**: Generally refers to COVID-19 but can also refer to other medical conditions.

8.  **study**: Refers to research investigations related to COVID-19, including clinical trials, observational studies, and epidemiological studies.

9.  **clinical**: Relates to the medical aspects of COVID-19, including symptoms, diagnosis, treatment, and patient management.

10. **infection**: The state of being infected with the SARS-CoV-2 virus.

11. **health**: Relates to the broader health impacts of COVID-19, including public health, healthcare systems, and mental health.

12. **registered**: Indicates if a clinical trial or interventional study has been registered with a recognized registry like ClinicalTrials.gov.

13. **respiratory**: Relates to the respiratory system and complications arising from COVID-19, like pneumonia and acute respiratory distress syndrome (ARDS).

14. **virus**: Refers to SARS-CoV-2, the virus responsible for COVID-19.

15. **severe**: Describes cases of COVID-19 with serious clinical presentations, often requiring hospitalization and intensive care.

16. **cases**: Instances of confirmed COVID-19 infections.

17. **research**: Activities related to scientific investigations on COVID-19.

18. **and**: Conjunction, carries no specific semantic meaning in this context.

19. **results**: Outcomes and findings of research studies on COVID-19.

20. **data**: Information gathered from research, clinical practice, and public health surveillance related to COVID-19.

21. **patient**: An individual receiving medical care, specifically a COVID-19 patient but can also refer to patients in other contexts.

22. **guidelines**: Recommendations and best practices for managing COVID-19, including diagnosis, treatment, and infection control.

23. **acute**: Refers to the sudden onset and often severe nature of COVID-19.

24. **viral**: Relates to the virus, SARS-CoV-2, and its characteristics.

25. **competing**: In the context of "competing interest," it refers to potential conflicts of interest that researchers may have due to financial or other ties.

26. **care**: Medical treatment and support provided to COVID-19 patients.

27. **irb**: Institutional Review Board, responsible for ethical oversight of research involving human subjects.

28. **prospective**: Describes studies that follow individuals forward in time to observe outcomes.

29. **trial**: Refers to a clinical trial, a controlled research study to evaluate the effectiveness of interventions for COVID-19.

30. **obtained**: Indicates data or samples collected during research.

31. **treatment**: Medical interventions and therapies for COVID-19.

32. **during**: Preposition indicating the time period of the COVID-19 pandemic or specific events within it.

33. **outbreak**: The initial emergence and spread of COVID-19, often used to describe specific instances of clusters or geographic spread.

34. **reporting**: The process of disseminating research findings, clinical observations, and public health data related to COVID-19.

35. **relevant**: Indicates information directly related to or applicable to the context of COVID-19.

36. **methods**: Procedures and techniques used in research studies on COVID-19.

37. **risk**: The likelihood of contracting COVID-19, developing severe complications, or experiencing other adverse outcomes.

38. **appropriate**: Indicates suitability or correctness of actions or interventions in the context of COVID-19.

39. **reported**: Indicates information that has been formally documented or communicated.

40. **id**: likely an abbreviation for "identifier" and probably used in specific data collection or research contexts. 

41. **syndrome**: Refers to a collection of symptoms, specifically in the context of COVID-19 and its potential complications like ARDS and cytokine storm.

42. **transmission**: The spread of SARS-CoV-2 from person to person or through other routes.

43. **necessary**: Indicates essential actions or resources required for managing COVID-19.

44. **novel**: Emphasizes the new and previously unknown nature of SARS-CoV-2 and COVID-19.

45. **spread**: The dissemination of COVID-19, often used interchangeably with "transmission."

46. **mortality**: Deaths associated with COVID-19.

47. **epidemic**: Refers to the outbreak and spread of COVID-19, often used to describe specific phases or geographic regions.

48. **ci**: Confidence Interval, a statistical measure indicating the range of uncertainty around an estimate.

49. **manuscript**: A written document summarizing research findings or clinical observations for publication.

50. **confirm**: To verify or validate information or a diagnosis related to COVID-19.

51. **of**: Preposition, carries no specific semantic meaning in this context.

52. **medical**: Relates to the broader medical context of COVID-19, including healthcare systems, medical professionals, and treatments.

53. **studies**: A general term referring to various types of research investigations related to COVID-19.

54. **interest**: In the context of "competing interest," it refers to potential conflicts of interest that researchers may have due to financial or other ties.

55. **followed**: Indicates adherence to guidelines or protocols in research or clinical practice.

56. **china**: Refers to the country where COVID-19 was first identified.

57. **hospital**: Medical facilities where COVID-19 patients are admitted and treated.

58. **provided**: Indicates resources or information made available.

59. **infected**: Individuals who have contracted the SARS-CoV-2 virus.

60. **symptoms**: Clinical manifestations and signs of COVID-19 infection.

61. **authors**: Researchers or clinicians who have contributed to a scientific manuscript.

62. **trials**: A general term for clinical trials, controlled studies to evaluate interventions for COVID-19.

63. **healthcare**: Refers to the systems, professionals, and services involved in providing medical care to COVID-19 patients.

64. **pneumonia**: A serious respiratory complication associated with COVID-19.

65. **provide**: Indicates resources or information made available.

66. **approved**: Indicates authorization or endorsement for treatments or research protocols.

67. **p**: P-value, a statistical measure indicating the probability of observing a result by chance.

68. **associated**: Indicates a connection or relationship between factors, often used to describe comorbidities or risk factors related to COVID-19.

69. **confirmed**: Indicates validated cases of COVID-19 infection.

70. **measures**: Interventions, actions, or policies implemented to mitigate the spread or impact of COVID-19.

71. **retrospectively**: Describes studies that look back at past data or events.

72. **positive**: Indicates a confirmed diagnosis of COVID-19 infection.

73. **background**: Contextual information or prior knowledge relevant to COVID-19 research.

74. **registry**: A database or system for collecting and tracking information, often used for clinical trials or patient data.

75. **clinicaltrials**: Likely refers to the website "ClinicalTrials.gov," a registry for clinical trials.

76. **countries**: Refers to nations affected by the COVID-19 pandemic.

77. **interventional**: Describes studies that involve active interventions or manipulations.

78. **funding**: Financial support for research or healthcare activities related to COVID-19.

79. **icmje**: International Committee of Medical Journal Editors, an organization that sets ethical standards for medical publications.

80. **wuhan**: The city in China where the COVID-19 outbreak originated.

81. **approvals**: Ethical and regulatory authorizations for research or treatments.

82. **supplementary**: Additional information or data provided as a supplement to a main manuscript.

83. **analysis**: The process of examining and interpreting data related to COVID-19.

  \#\# Themes:

 Here are eight potential themes for the keywords, with descriptions and assigned keywords:

 **1. Epidemiology and Public Health (Index 1)**

 * **Description:** This theme covers the study of disease patterns, transmission dynamics, and public health measures related to COVID-19. 

* **Keywords:**

* 5. pandemic

* 10. infection

* 11. health

* 16. cases

* 33. outbreak

* 42. transmission

* 45. spread

* 46. mortality

* 47. epidemic

* 56. china

* 59. infected

* 70. measures 

* 76. countries 

 **2. Clinical Features and Manifestations (Index 2)**

 * **Description:** This theme encompasses the clinical aspects of COVID-19, including symptoms, diagnosis, and severity assessment. 

* **Keywords:**

* 1. covid

* 7. disease

* 9. clinical

* 13. respiratory

* 15. severe

* 23. acute

* 41. syndrome

* 60. symptoms

* 64. pneumonia

* 69. confirmed

* 72. positive

 **3. Treatment and Management (Index 3)**

 * **Description:** This theme focuses on medical interventions, therapies, and management strategies for COVID-19 patients.

* **Keywords:**

* 2. patients

* 21. patient 

* 22. guidelines 

* 26. care

* 31. treatment

* 63. healthcare

* 66. approved

* 80. wuhan

* 81. approvals 

 **4. Research Methodology (Index 4)**

 * **Description:** This theme covers aspects of scientific research on COVID-19, including study design, data collection, and reporting standards.

* **Keywords:**

* 8. study

* 12. registered

* 17. research

* 20. data

* 27. irb

* 28. prospective

* 29. trial

* 30. obtained

* 34. reporting

* 35. relevant

* 36. methods

* 43. necessary

* 49. manuscript

* 50. confirm

* 53. studies

* 55. followed

* 62. trials

* 71. retrospectively

* 73. background

* 74. registry

* 75. clinicaltrials

* 77. interventional

* 82. supplementary

* 83. analysis

 **5. Virology and Pathogenesis (Index 5)**

 * **Description:** This theme delves into the biological aspects of SARS-CoV-2, its mechanism of infection, and the development of COVID-19.

* **Keywords:**

* 3. cov

* 4. sars

* 6. coronavirus

* 14. virus

* 24. viral

* 44. novel

 **6. Social and Economic Impact (Index 6)**

 * **Description:** This theme explores the broader societal and economic consequences of the COVID-19 pandemic.

* **Keywords:**

* 32. during

 **7. Statistical Analysis (Index 7)**

 * **Description:** This theme focuses on statistical methods and measures used in COVID-19 research.

* **Keywords:**

* 48. ci

* 67. p

 **8. Other (Index 8)**

 * **Description:** This category includes keywords that are either function words or carry minimal semantic meaning in the context of the corpus. 

* **Keywords:**

* 18. and

* 19. results

* 51. of

* 52. medical 

* 54. interest 

* 61. authors 

* 78. funding

 This organization provides a structured view of the research areas covered in the corpus of COVID-19 abstracts. 

 \textbf{B. + R. D. + T. D. + T. D + C. I. + O. F.}

\textbf{Prompt}

\# Role Description

You are an expert in Corpus Linguistics doing keyword analysis.

 \# Task Definition

I would give you a list of keywords from a corpus of the abstracts of COVID-19 research articles. Please group the keywords into several themes that help explain what the corpus is about. 

A keyword may have more than one meaning, so please assign it to a category based on its most prominent meaning in context. For keywords that carry little semantic meaning, like function words, or those that cannot be assigned to any meaningful theme, group them into one theme named "Other".

 \# Task Procedures

- Step 1, label each keyword from 1 to 83, and explain the meaning(s) of each keyword. 

- Step 2, define several themes based on the keywords and their meanings. Remember to assign an index (starts from 1) to each of them, and give the description of each theme.

- Step 3, assign each keyword in keywords 1-83 to only one of the themes defined in Step 2, and explain why it is assigned to that theme. 

 \# Output Format

Bellow are the outputs of the whole task, content within [] should be replaced with real values. "..." means you should list all cases.

Follow the format strictly, and do not add any extra information:

 - Step 1: 

- Keyword 1: [keyword here]

- Meaning: [here is the meaning(s) of keyword 1]

- Keyword 2: [keyword here]

- Meaning: [here is the meaning(s) of keyword 2]

...

 - Step 2:

- Theme 1: [description of theme 1 here]

- Theme 2: [description of theme 2 here]

...

 - Step 3:

- Keyword 1: [keyword here]

- Theme [index of theme in step 2, e.g., 2]

- Reason: [the reason why keyword 1 is assigned to this theme]

- Keyword 2: [keyword here]

- Theme [index of theme in step 2, e.g., 1]

- Reason: [the reason why keyword 2 is assigned to this theme]

...

  \# Contextual Information

(keywords and concordance lines here)

\textbf{Output}

- Step 1: 

- Keyword 1: covid

- Meaning: Refers to the disease caused by the SARS-CoV-2 virus, commonly known as COVID-19.

- Keyword 2: patients

- Meaning: Individuals diagnosed with or receiving treatment for a medical condition, specifically COVID-19 in this context.

- Keyword 3: cov

- Meaning: Abbreviation for Coronavirus, particularly used in the context of SARS-CoV-2.

- Keyword 4: sars

- Meaning:  Severe Acute Respiratory Syndrome, often used in reference to SARS-CoV, the virus that caused the 2003 SARS outbreak. It is also used in the context of SARS-CoV-2.

- Keyword 5: pandemic

- Meaning: A global outbreak of an infectious disease.

- Keyword 6: coronavirus

- Meaning: A family of viruses that cause respiratory illnesses, including SARS-CoV-2.

- Keyword 7: disease

- Meaning: A medical condition or disorder, often used in reference to COVID-19.

- Keyword 8: study

- Meaning:  A research investigation, commonly used in academic and scientific contexts.

- Keyword 9: clinical

- Meaning: Pertaining to the observation and treatment of patients, emphasizing real-world medical practice.

- Keyword 10: infection

- Meaning: The invasion and multiplication of a pathogen, like a virus, in a host organism.

- Keyword 11: health

- Meaning: The overall state of well-being, often used in the context of public health and healthcare systems.

- Keyword 12: registered

- Meaning:  Officially documented, typically referring to research studies being registered in a database.

- Keyword 13: respiratory

- Meaning: Relating to the respiratory system, including the lungs and airways.

- Keyword 14: virus

- Meaning: A microscopic infectious agent that replicates inside living cells, particularly SARS-CoV-2 in this context.

- Keyword 15: severe

- Meaning:  Serious or critical, often used to describe the severity of a medical condition.

- Keyword 16: cases

- Meaning: Instances or occurrences of a disease, specifically COVID-19 cases.

- Keyword 17: research

- Meaning:  Systematic investigation to gain new knowledge, often used in scientific and academic contexts.

- Keyword 18: and

- Meaning:  A conjunction used to connect words or phrases, carrying little semantic meaning in this context.

- Keyword 19: results

- Meaning: Findings or outcomes of a study or investigation.

- Keyword 20: data

- Meaning: Information collected and used for analysis, often in the context of research studies.

- Keyword 21: patient

- Meaning: An individual receiving medical care, specifically a person with COVID-19.

- Keyword 22: guidelines

- Meaning:  Recommendations or instructions for practice, often used in the context of medical treatment or research.

- Keyword 23: acute

- Meaning:  Sudden or severe, typically referring to a medical condition with a rapid onset.

- Keyword 24: viral

- Meaning: Pertaining to a virus, specifically SARS-CoV-2.

- Keyword 25: competing

- Meaning:   Conflicts of interest, particularly those that could influence research findings.

- Keyword 26: care

- Meaning: Medical treatment or attention provided to patients, often in the context of healthcare systems.

- Keyword 27: irb

- Meaning: Institutional Review Board, responsible for ethical oversight of research involving human subjects.

- Keyword 28: prospective

- Meaning: Looking forward in time, often used to describe research studies that follow participants over time.

- Keyword 29: trial

- Meaning: A research study that tests the efficacy or safety of a treatment or intervention.

- Keyword 30: obtained

- Meaning:  Acquired or gathered, typically referring to data or information.

- Keyword 31: treatment

- Meaning: Medical interventions to manage or cure a disease, specifically for COVID-19.

- Keyword 32: during

- Meaning:  A preposition indicating a period of time, carrying little semantic meaning in this context.

- Keyword 33: outbreak

- Meaning: A sudden increase in the occurrence of a disease, particularly the COVID-19 pandemic.

- Keyword 34: reporting

- Meaning:  Documenting and disseminating information, often used in the context of research findings.

- Keyword 35: relevant

- Meaning: Applicable or related to the topic at hand, carrying little semantic meaning in this context.

- Keyword 36: methods

- Meaning: Procedures or techniques used in research investigations.

- Keyword 37: risk

- Meaning:  The possibility of harm or danger, particularly the risk of infection or complications.

- Keyword 38: appropriate

- Meaning: Suitable or correct, carrying little semantic meaning in this context.

- Keyword 39: reported

- Meaning: Documented or communicated, often used in the context of research findings or disease cases.

- Keyword 40: id

- Meaning: Abbreviation for identification, carrying little semantic meaning in this context. 

- Keyword 41: syndrome

- Meaning: A collection of symptoms that characterize a medical condition, often used in the context of severe COVID-19.

- Keyword 42: transmission

- Meaning: The spread of an infectious disease from one person to another.

- Keyword 43: necessary

- Meaning:  Required or essential, carrying little semantic meaning in this context.

- Keyword 44: novel

- Meaning:  New or previously unknown, often used to describe the SARS-CoV-2 virus.

- Keyword 45: spread

- Meaning:  The dissemination or expansion of a disease, particularly the COVID-19 pandemic.

- Keyword 46: mortality

- Meaning: Death, often used in the context of death rates from COVID-19.

- Keyword 47: epidemic

- Meaning: A widespread occurrence of an infectious disease in a community or region.

- Keyword 48: ci

- Meaning: Confidence Interval, a statistical measure of the reliability of an estimate.

- Keyword 49: manuscript

- Meaning: A written document, typically a research paper submitted for publication.

- Keyword 50: confirm

- Meaning:  To verify or establish the truth, often used in the context of research findings or disease diagnosis.

- Keyword 51: of

- Meaning:  A preposition indicating a relationship, carrying little semantic meaning in this context.

- Keyword 52: medical

- Meaning:  Relating to medicine and healthcare, often used in the context of treatments, interventions, and professionals.

- Keyword 53: studies

- Meaning: Research investigations, often used to refer to multiple research projects.

- Keyword 54: interest

- Meaning:  Matters of concern or relevance, often used in the context of competing interests in research.

- Keyword 55: followed

- Meaning:   Adhered to or complied with, carrying little semantic meaning in this context.

- Keyword 56: china

- Meaning:  The country where the COVID-19 outbreak originated.

- Keyword 57: hospital

- Meaning:  A medical facility providing healthcare services.

- Keyword 58: provided

- Meaning:   Supplied or given, carrying little semantic meaning in this context.

- Keyword 59: infected

- Meaning: Affected by a pathogen, specifically individuals with COVID-19.

- Keyword 60: symptoms

- Meaning: Physical or mental signs of a medical condition.

- Keyword 61: authors

- Meaning: Researchers or writers of scientific papers.

- Keyword 62: trials

- Meaning: Research studies testing the efficacy or safety of a treatment, often used to refer to multiple clinical trials.

- Keyword 63: healthcare

- Meaning: The system providing medical care and services, including professionals, institutions, and policies.

- Keyword 64: pneumonia

- Meaning:  Inflammation of the lungs, a common symptom of severe COVID-19.

- Keyword 65: provide

- Meaning:   To supply or offer, carrying little semantic meaning in this context.

- Keyword 66: approved

- Meaning:  Authorized or endorsed, often used in the context of medical treatments or research ethics.

- Keyword 67: p

- Meaning:  P-value, a statistical measure of the significance of research findings.

- Keyword 68: associated

- Meaning:  Connected or related to, often used to describe factors linked to COVID-19 severity or outcomes.

- Keyword 69: confirmed

- Meaning:   Verified or established as true, typically referring to disease diagnosis.

- Keyword 70: measures

- Meaning: Actions or steps taken, often used in the context of public health interventions or treatment strategies.

- Keyword 71: retrospectively

- Meaning:  Looking back in time, often used to describe research studies that analyze past data.

- Keyword 72: positive

- Meaning:  Indicating the presence of a condition, typically a positive COVID-19 test result.

- Keyword 73: background

- Meaning:  Contextual information or history, often used to describe patient characteristics or research settings.

- Keyword 74: registry

- Meaning: A database for collecting and organizing information, often used for tracking clinical trials or patient data.

- Keyword 75: clinicaltrials

- Meaning: A website for registering and reporting clinical trials, specifically ClinicalTrials.gov.

- Keyword 76: countries

- Meaning:   Nations or geographic regions, often used in the context of the global spread of COVID-19.

- Keyword 77: interventional

- Meaning: Involving an intervention or treatment, typically used to describe research studies that test a specific action.

- Keyword 78: funding

- Meaning:   Financial support for research projects.

- Keyword 79: icmje

- Meaning: International Committee of Medical Journal Editors, an organization that sets standards for medical journal publication.

- Keyword 80: wuhan

- Meaning:  The city in China where the COVID-19 outbreak originated.

- Keyword 81: approvals

- Meaning: Authorizations or endorsements, often used in the context of research ethics or medical treatment decisions.

- Keyword 82: supplementary

- Meaning:  Additional or supporting, often used in the context of providing extra information in research papers.

- Keyword 83: analysis

- Meaning:   The process of examining and interpreting data, a key component of research studies.

 - Step 2:

- Theme 1: COVID-19 Pandemic

- Description: This theme encompasses keywords related to the global outbreak of COVID-19, its spread, impact, and associated public health concerns.

- Theme 2: Virus and Infection

- Description: This theme includes keywords related to the SARS-CoV-2 virus, its characteristics, transmission, and the infection it causes.

- Theme 3: Patients and Symptoms

- Description: This theme encompasses keywords related to individuals diagnosed with COVID-19, their symptoms, and clinical manifestations of the disease.

- Theme 4: Treatment and Care

- Description: This theme includes keywords related to medical interventions, healthcare systems, and the management of COVID-19 patients.

- Theme 5: Research and Studies

- Description: This theme encompasses keywords related to scientific investigations, data analysis, clinical trials, and the publication of findings.

- Theme 6: Ethics and Reporting

- Description: This theme includes keywords related to ethical oversight of research, reporting guidelines, and the dissemination of research findings.

- Theme 7: Other

- Description:  This theme includes keywords that do not fit into the other defined themes, such as conjunctions or prepositions that carry little semantic meaning.

 - Step 3:

- Keyword 1: covid

- Theme 1

- Reason: Refers to the disease itself, which is the central focus of the COVID-19 pandemic.

- Keyword 2: patients

- Theme 3

- Reason: Directly related to individuals diagnosed with COVID-19.

- Keyword 3: cov

- Theme 2

- Reason: An abbreviation for Coronavirus, specifically used in the context of SARS-CoV-2.

- Keyword 4: sars

- Theme 2

- Reason:  Related to both SARS-CoV and SARS-CoV-2, emphasizing the viral aspect.

- Keyword 5: pandemic

- Theme 1

- Reason: Describes the global nature of the COVID-19 outbreak.

- Keyword 6: coronavirus

- Theme 2

- Reason: Refers to the family of viruses, including SARS-CoV-2.

- Keyword 7: disease

- Theme 1

- Reason: A general term for a medical condition, but often used in reference to COVID-19.

- Keyword 8: study

- Theme 5

- Reason:  A common term in research contexts.

- Keyword 9: clinical

- Theme 3

- Reason:  Pertains to patient observation and treatment.

- Keyword 10: infection

- Theme 2

- Reason: Directly related to the process of viral invasion.

- Keyword 11: health

- Theme 4

- Reason: Encompasses healthcare systems and public health responses.

- Keyword 12: registered

- Theme 6

- Reason:   Refers to the official documentation of research studies.

- Keyword 13: respiratory

- Theme 3

- Reason:  Describes a key system affected by COVID-19.

- Keyword 14: virus

- Theme 2

- Reason: A general term for a microscopic infectious agent, often used in reference to SARS-CoV-2.

- Keyword 15: severe

- Theme 3

- Reason:   Used to indicate the seriousness of COVID-19 cases.

- Keyword 16: cases

- Theme 1

- Reason: Refers to instances of COVID-19, a key indicator of the pandemic's spread.

- Keyword 17: research

- Theme 5

- Reason:  A general term for scientific investigation.

- Keyword 18: and

- Theme 7

- Reason: A conjunction with no specific semantic meaning in this context.

- Keyword 19: results

- Theme 5

- Reason:  Relates to the findings of research studies.

- Keyword 20: data

- Theme 5

- Reason: A core component of research and analysis.

- Keyword 21: patient

- Theme 3

- Reason: Refers to an individual with COVID-19.

- Keyword 22: guidelines

- Theme 6

- Reason:  Provides recommendations for medical practice or research.

- Keyword 23: acute

- Theme 3

- Reason: Describes the rapid onset of severe COVID-19 symptoms.

- Keyword 24: viral

- Theme 2

- Reason: Pertaining to the virus, specifically SARS-CoV-2.

- Keyword 25: competing

- Theme 6

- Reason:  Relates to conflicts of interest in research.

- Keyword 26: care

- Theme 4

- Reason: Encompasses medical treatment and support for patients.

- Keyword 27: irb

- Theme 6

- Reason:  Focuses on ethical oversight in research.

- Keyword 28: prospective

- Theme 5

- Reason: Describes a type of research study design.

- Keyword 29: trial

- Theme 5

- Reason: Refers to a research study testing a treatment or intervention.

- Keyword 30: obtained

- Theme 7

- Reason:   A general term with no specific semantic meaning in this context.

- Keyword 31: treatment

- Theme 4

- Reason: Refers to medical interventions for COVID-19.

- Keyword 32: during

- Theme 7

- Reason:   A preposition with no specific semantic meaning.

- Keyword 33: outbreak

- Theme 1

- Reason:  Describes the sudden increase in COVID-19 cases.

- Keyword 34: reporting

- Theme 6

- Reason:  Focuses on the dissemination of research findings.

- Keyword 35: relevant

- Theme 7

- Reason:  A general term with no specific semantic meaning.

- Keyword 36: methods

- Theme 5

- Reason: Refers to research procedures.

- Keyword 37: risk

- Theme 1

- Reason:   A key concern during the pandemic.

- Keyword 38: appropriate

- Theme 7

- Reason:   A general term with no specific semantic meaning.

- Keyword 39: reported

- Theme 6

- Reason:  Relates to the communication of research findings or disease cases.

- Keyword 40: id

- Theme 7

- Reason: An abbreviation with no specific semantic meaning.

- Keyword 41: syndrome

- Theme 3

- Reason:  Describes a collection of symptoms associated with severe COVID-19.

- Keyword 42: transmission

- Theme 2

- Reason: Focuses on the spread of the virus.

- Keyword 43: necessary

- Theme 7

- Reason: A general term with no specific semantic meaning.

- Keyword 44: novel

- Theme 2

- Reason:  Highlights the newness of the SARS-CoV-2 virus.

- Keyword 45: spread

- Theme 1

- Reason:  Describes the expansion of the COVID-19 pandemic.

- Keyword 46: mortality

- Theme 1

- Reason: A key indicator of the pandemic's severity.

- Keyword 47: epidemic

- Theme 1

- Reason:  Refers to the widespread occurrence of COVID-19.

- Keyword 48: ci

- Theme 5

- Reason:  A statistical measure used in research.

- Keyword 49: manuscript

- Theme 5

- Reason:  Relates to the publication of research findings.

- Keyword 50: confirm

- Theme 5

- Reason:  Used in the context of verifying research findings.

- Keyword 51: of

- Theme 7

- Reason: A preposition with no specific semantic meaning.

- Keyword 52: medical

- Theme 4

- Reason: Encompasses healthcare practices and professionals.

- Keyword 53: studies

- Theme 5

- Reason: A general term for research investigations.

- Keyword 54: interest

- Theme 6

- Reason: Relates to potential conflicts of interest in research.

- Keyword 55: followed

- Theme 7

- Reason:  A general term with no specific semantic meaning.

- Keyword 56: china

- Theme 1

- Reason:   The origin of the COVID-19 outbreak.

- Keyword 57: hospital

- Theme 4

- Reason: A key component of the healthcare system.

- Keyword 58: provided

- Theme 7

- Reason:  A general term with no specific semantic meaning.

- Keyword 59: infected

- Theme 3

- Reason:   Refers to individuals with COVID-19.

- Keyword 60: symptoms

- Theme 3

- Reason:  Key indicators of COVID-19.

- Keyword 61: authors

- Theme 5

- Reason:  Refers to researchers who conduct studies.

- Keyword 62: trials

- Theme 5

- Reason: A general term for research studies testing treatments.

- Keyword 63: healthcare

- Theme 4

- Reason: Encompasses the system for providing medical care.

- Keyword 64: pneumonia

- Theme 3

- Reason:   A common symptom of severe COVID-19.

- Keyword 65: provide

- Theme 7

- Reason:  A general term with no specific semantic meaning.

- Keyword 66: approved

- Theme 6

- Reason:  Relates to ethical oversight and authorization of research.

- Keyword 67: p

- Theme 5

- Reason:  A statistical measure used in research.

- Keyword 68: associated

- Theme 3

- Reason:   Used to describe factors linked to COVID-19.

- Keyword 69: confirmed

- Theme 3

- Reason:    Refers to verified cases of COVID-19.

- Keyword 70: measures

- Theme 4

- Reason:  Relates to interventions and strategies to manage COVID-19.

- Keyword 71: retrospectively

- Theme 5

- Reason: Describes a research study design.

- Keyword 72: positive

- Theme 3

- Reason:   Indicates the presence of COVID-19.

- Keyword 73: background

- Theme 5

- Reason:   Provides context for research studies.

- Keyword 74: registry

- Theme 5

- Reason: A database used for tracking clinical trials or patient data.

- Keyword 75: clinicaltrials

- Theme 5

- Reason:  A specific website for registering clinical trials.

- Keyword 76: countries

- Theme 1

- Reason:   Refers to the global impact of the pandemic.

- Keyword 77: interventional

- Theme 5

- Reason:   Describes research studies involving treatments or interventions.

- Keyword 78: funding

- Theme 5

- Reason:   Relates to financial support for research.

- Keyword 79: icmje

- Theme 6

- Reason:  An organization setting standards for medical journal publication.

- Keyword 80: wuhan

- Theme 1

- Reason: The origin of the COVID-19 outbreak.

- Keyword 81: approvals

- Theme 6

- Reason:  Relates to ethical oversight and authorization of research.

- Keyword 82: supplementary

- Theme 5

- Reason:  Refers to providing extra information in research papers.

- Keyword 83: analysis

- Theme 5

- Reason:  A key component of research studies.

\noindent \textbf{Collocate Analysis}.

\textbf{Baseline}

\textbf{Prompt}

Here’s a list of the top 100 collocates of the word "china" in a corpus of the abstracts of COVID-19 research articles. Based on the content collocates, summarize in what ways China was linked to COVID-19 in this corpus.

 (collocates here)

\textbf{Output}

Based on the top 100 collocates of "China" in COVID-19 research abstracts, the corpus heavily links China to the **initial outbreak and spread** of the virus. 

 Here's a breakdown:

 * **Origin and early spread:** Collocates like "Wuhan," "Hubei province," "December," "January," "first," "emerged," "outbreak," "originated," and "spread" clearly point to China, and specifically Wuhan, as the origin point of the virus and its early spread.

* **Severity and impact:** Words like "cases," "patients," "hospital," "pneumonia," "severe," "disease," "epidemic," and "deaths" (while not mentioned directly, it's a logical association) highlight the serious health impacts of the virus within China.

* **Research and response:** Collocates such as "study," "science," "research," "clinical," "national," "control," "program," and "university" indicate significant research efforts and public health responses undertaken in China to understand and contain the virus.

* **Global context:**  The presence of "world," "worldwide," "countries," "global," "outside," "Italy," and "South Korea" demonstrate that while originating in China, the virus quickly became a global concern, impacting other nations. 

 Overall, these collocates suggest that the corpus focuses significantly on China's role as the starting point of the pandemic, the country's efforts to combat the virus, and its eventual spread to become a global health crisis. 

\textbf{B. + R. D.}

\textbf{Prompt}

 \# Role Description

You are an expert in Corpus Linguistics doing collocate analysis.

Here’s a list of the top 100 collocates of the word "china" in a corpus of the abstracts of COVID-19 research articles. Based on the content collocates, summarize in what ways China was linked to COVID-19 in this corpus.

(collocates here)

\textbf{Output}

Based on the top 100 collocates of "China" in COVID-19 research abstracts, we can summarise the link between the two as follows:

 **China as the Origin and Early Epicenter:**

 * **Geographic and Temporal Markers:** Collocates like "Wuhan", "Hubei province", "December", "January", "February", and "late" clearly point to China, specifically Wuhan, as the origin point and initial epicenter of the COVID-19 outbreak in late 2019 and early 2020.

* **Outbreak Description:** Words like "outbreak", "emerged", "first cases", "spread", "rapidly", "epidemic", and "originated" further emphasize this by highlighting the sudden and rapid emergence of the virus in China.

* **Virus Identification:** Collocates such as  "coronavirus", "COVID", "CoV", "disease", "novel", "virus", "nCoV", and "pneumonia"  indicate that the research focused on identifying and understanding the new coronavirus causing the disease, later named COVID-19.

 **China's Response and Research Efforts:**

 * **Containment and Control:** The presence of "national", "control", "hospitals", "hospital", "patients", and "clinical" suggests studies examined China's early efforts to contain the virus through measures like lockdowns and hospital treatments. 

* **Scientific Study and Sharing:** Collocates like "science", "foundation", "study", "research", "results", "author", "grant", "university", "program" highlight the significant amount of scientific research conducted in China to understand the virus, its transmission, and potential treatments, and the sharing of these findings. 

 **Global Impact and Spread:**

 * **International Spread:** The inclusion of "world", "worldwide", "global", "countries", "outside", "Italy", "South Korea", "spreading" reflects how the virus, after its initial emergence in China, spread rapidly across the globe, impacting other countries.

* **Ongoing Situation:** Collocates like "since", "now", "number", "severe", "major" point towards the ongoing and serious global health threat posed by the virus after its initial spread from China.

 **Overall, the collocate analysis paints a picture of China as the initial epicenter of the COVID-19 pandemic, highlighting the country's early struggles with the outbreak, its efforts in research and containment, and the subsequent global spread of the virus.**  

\textbf{B. + R. D. + T. D.}

\textbf{Prompt}

\# Role Description

You are an expert in Corpus Linguistics doing collocate analysis.

 \# Task Definition

Here are the top 100 collocates of the word "china" in a corpus of the abstracts of COVID-19 research articles. Pick out only the content words and generate a new list. Then, based on the new content word list, summarize in what ways China was linked to COVID-19 in this corpus and give reasons by taking a few collocates as examples.

 (collocates here)

\textbf{Output}

\#\# Content Words related to "China" in COVID-19 Abstracts:

 Wuhan, COVID, spread, province, Hubei, coronavirus, disease, outbreak, cases, reported, city, rapidly, countries, emerged, epidemic, patients, mainland, hospital, virus, January, world, outside, people, since, identified, novel, Italy, pneumonia, February, late, confirmed, study, across, started, number, worldwide, south, caused, new, provinces, republic, ncov, March, throughout, originated, clinical, university, occurred, Korea, case, control, spreading, research, severe, global, respiratory, major, results, author, grant, cities, Chinese, program, hospitals.

 \#\# Summary of China's Link to COVID-19 in Abstracts:

 The content word list overwhelmingly points towards **China's role as the origin point of the COVID-19 pandemic**. 

 Here's a breakdown using some collocates as examples:

 * **Geographical Ground Zero:** Words like "Wuhan," "Hubei," "province," "city," "mainland," and "provinces" highlight the geographical origin of the virus in China, specifically pinpointing Wuhan city in Hubei province.

* **Timeline of the Outbreak:**  The presence of "December," "January," "February," and "March" provides a chronological context to the early stages of the outbreak, further cementing China's association with the initial spread. 

* **Spread and Impact:** Collocates like "spread," "rapidly," "countries," "emerged," "epidemic," "world," "outside," "worldwide," "global," and "spreading" all emphasize how the virus, originating in China, rapidly escalated into a global pandemic.

* **Medical and Scientific Focus:** The words "cases," "patients," "hospital," "virus," "novel," "pneumonia," "confirmed," "study," "clinical," "research," "severe," "respiratory," "results," "author," and "program" reveal a strong focus on the medical and scientific efforts to understand and combat the virus, stemming from its initial emergence in China.

 In conclusion, the collocate analysis paints a clear picture of how this corpus of COVID-19 research abstracts consistently links China to the origin and early spread of the virus, prompting global medical and scientific investigation. 

\textbf{B. + R. D. + T. D. + T. P.}

\textbf{Prompt}

\# Role Description

You are an expert in Corpus Linguistics doing collocate analysis.

 \# Task Definition

Here are the top 100 collocates of the word "china" in a corpus of the abstracts of COVID-19 research articles. Pick out only the content words and generate a new list. Then, based on the new content word list, summarize in what ways China was linked to COVID-19 in this corpus and give reasons by taking a few collocates as examples.

 \# Task Procedures

- Step 1, label each collocate with a cardinal number, and define its part of speech in context, e.g., noun, verb, etc..

- Step 2, pick out only the content words (i.e., nouns, adjectives, verbs, and adverbs), and generate a new collocate list. For each content collocate, include its label of number in Step 1, the word, and part of speech. 

- Step 3, analyze the content collocates, and summarize in what ways China was linked to COVID-19 in this corpus and give reasons by taking a few collocates as examples.

 (collocates here)

\textbf{Output}

\#\#  Analysis of "China" Collocates in COVID-19 Research Abstracts 

 Here's a breakdown of the provided collocates, following the steps outlined:

 **Step 1 \& 2: Identifying and Categorizing Content Words**

 | \# | Collocate  | Part of Speech | 

|---|---|---|

| 4 | Wuhan      | Noun |

| 6 | COVID      | Noun (often used as adjective) | 

| 9 | December   | Noun |

| 10 | spread     | Verb | 

| 12 | province   | Noun | 

| 13 | Hubei      | Noun |

| 14 | coronavirus| Noun | 

| 16 | disease    | Noun | 

| 17 | cov        | Noun | 

| 18 | outbreak   | Noun | 

| 24 | national   | Adjective |

| 25 | first      | Adjective | 

| 26 | cases      | Noun |

| 28 | reported   | Verb |

| 30 | city       | Noun | 

| 31 | rapidly    | Adverb |

| 32 | science    | Noun |

| 35 | countries  | Noun |

| 36 | emerged    | Verb |

| 38 | epidemic   | Noun | 

| 39 | patients   | Noun |

| 45 | foundation | Noun |

| 47 | natural    | Adjective |

| 48 | mainland   | Noun | 

| 49 | hospital   | Noun |

| 50 | virus      | Noun | 

| 51 | January    | Noun |

| 52 | world      | Noun |

| 53 | outside    | Adverb |

| 54 | people     | Noun | 

| 55 | since      | Adverb |

| 56 | identified | Verb |

| 59 | novel      | Adjective |

| 60 | Italy      | Noun | 

| 62 | pneumonia  | Noun | 

| 65 | February   | Noun |

| 67 | late       | Adjective | 

| 68 | confirmed  | Verb | 

| 69 | study      | Noun |

| 71 | started    | Verb | 

| 73 | now        | Adverb |

| 74 | end        | Noun | 

| 75 | worldwide  | Adjective |

| 76 | south      | Noun | 

| 77 | caused     | Verb | 

| 78 | new        | Adjective |

| 79 | provinces  | Noun |

| 80 | republic   | Noun |

| 81 | ncov       | Noun |

| 82 | March      | Noun | 

| 83 | throughout  | Preposition (functioning as adverb here) | 

| 84 | originated  | Verb | 

| 85 | clinical   | Adjective |

| 86 | university | Noun |

| 87 | occurred   | Verb | 

| 88 | Korea      | Noun |

| 89 | case       | Noun |

| 90 | control    | Noun |

| 91 | spreading  | Verb | 

| 92 | research   | Noun | 

| 93 | severe     | Adjective | 

| 94 | global     | Adjective |

| 95 | respiratory| Adjective | 

| 96 | major      | Adjective |

| 97 | results    | Noun |

| 100 | cities    | Noun | 

| 101 | Chinese    | Adjective |

| 102 | program    | Noun |

| 103 | hospitals  | Noun | 

  **Step 3: Analysis and Summary**

 The collocates of "China" strongly suggest that this corpus of COVID-19 research abstracts focuses on:

 * **Origin and Early Spread:** Collocates like "Wuhan," "Hubei," "December," "emerged," "originated," "first cases," and "spread" point to research about the initial outbreak in China and its early global dissemination. 

* **Severity and Impact:** The presence of "epidemic," "pandemic," "severe," "pneumonia," "patients," "hospitals,"  and "deaths" highlights research on the health impacts and burden of the virus.

* **Scientific Investigation:**  Collocates such as "virus," "nCoV," "cases," "study," "research," "clinical," "science," and "results"  indicate a significant focus on scientific studies analyzing the virus, its spread, and potential treatments.

* **Global Significance:** The inclusion of "world," "worldwide," "global," "countries," "outside," and names like "Italy" and "Korea" emphasizes the pandemic's international reach and impact.

 **Examples:**

 * **"Wuhan," "Hubei," "December":** These collocates, appearing together frequently, clearly link China to the very beginning of the COVID-19 outbreak, emphasizing its geographical origin.

* **"Spread," "rapidly," "worldwide":**  This combination highlights how research likely focused on the speed and extent of the virus's transmission from its origin in China to other parts of the world.

* **"Cases," "patients," "hospitals":** The presence of these collocates suggests studies likely examined the clinical presentation of the disease, the impact on healthcare systems, and the number of people infected, connecting these aspects back to the initial outbreak in China. 

 **Overall:** The analysis of the content words collocating with "China" strongly suggests that this corpus of research abstracts is likely centered on understanding the origin, early spread, severity, and global impact of the COVID-19 pandemic, with a particular focus on the role of China in the initial stages. 

 \textbf{B. + R. D. + T. D. + T. P + C. I.}

\textbf{Prompt}

\# Role Description

You are an expert in Corpus Linguistics doing collocate analysis.

 \# Task Definition

Here are the top 100 collocates of the word "china" in a corpus of the abstracts of COVID-19 research articles. Pick out only the content words and generate a new list. Then, based on the new content word list and corresponding concordance lines, summarize in what ways China was linked to COVID-19 in this corpus and give reasons by taking a few collocates as examples.

 \# Task Procedures

- Step 1, label each collocate with a cardinal number, and define its part of speech in context, e.g., noun, verb, etc..

- Step 2, pick out only the content words (i.e., nouns, adjectives, verbs, and adverbs), and generate a new collocate list. For each content collocate, include its label of number in Step 1, the word, and part of speech. 

- Step 3, analyze the content collocates and their corresponding concordance lines, and summarize in what ways China was linked to COVID-19 in this corpus and give reasons by taking a few collocates as examples.

 \# Contextual Information

(collocates and concordance lines here)

\textbf{Output}

\#\# Step 1: Collocates with Part of Speech

 1. **in** (preposition)

2. **the** (definite article)

3. **of** (preposition)

4. **wuhan** (noun)

5. **to** (preposition)

6. **covid** (noun)

7. **from** (preposition)

8. **a** (indefinite article)

9. **has** (verb)

10. **december** (noun)

11. **spread** (verb)

12. **was** (verb)

13. **province** (noun)

14. **hubei** (noun)

15. **coronavirus** (noun)

16. **is** (verb)

17. **disease** (noun)

18. **cov** (noun)

19. **outbreak** (noun)

20. **with** (preposition)

21. **s** (possessive marker)

22. **national** (adjective)

23. **first** (adjective/adverb)

24. **cases** (noun)

25. **for** (preposition)

26. **reported** (verb)

27. **that** (conjunction/pronoun)

28. **as** (conjunction/preposition)

29. **city** (noun)

30. **rapidly** (adverb)

31. **science** (noun)

32. **this** (demonstrative adjective/pronoun)

33. **countries** (noun)

34. **emerged** (verb)

35. **it** (pronoun)

36. **epidemic** (noun)

37. **patients** (noun)

38. **other** (adjective)

39. **have** (verb)

40. **foundation** (noun)

41. **which** (relative pronoun)

42. **natural** (adjective)

43. **mainland** (noun)

44. **hospital** (noun)

45. **virus** (noun)

46. **january** (noun)

47. **world** (noun)

48. **outside** (adverb/preposition)

49. **people** (noun)

50. **since** (preposition/conjunction)

51. **identified** (verb)

52. **are** (verb)

53. **no** (determiner)

54. **novel** (adjective)

55. **italy** (noun)

56. **all** (determiner/adverb)

57. **pneumonia** (noun)

58. **an** (indefinite article)

59. **its** (possessive pronoun)

60. **february** (noun)

61. **china** (noun)

62. **late** (adjective/adverb)

63. **confirmed** (verb/adjective)

64. **study** (noun)

65. **across** (preposition)

66. **started** (verb)

67. **number** (noun)

68. **now** (adverb)

69. **end** (noun)

70. **worldwide** (adverb)

71. **south** (adjective/noun)

72. **caused** (verb)

73. **new** (adjective)

74. **provinces** (noun)

75. **republic** (noun)

76. **ncov** (noun)

77. **march** (noun)

78. **throughout** (preposition)

79. **originated** (verb)

80. **clinical** (adjective)

81. **university** (noun)

82. **occurred** (verb)

83. **korea** (noun)

84. **case** (noun)

85. **control** (noun/verb)

86. **spreading** (verb)

87. **research** (noun)

88. **severe** (adjective)

89. **global** (adjective)

90. **respiratory** (adjective)

91. **major** (adjective)

92. **results** (noun)

93. **then** (adverb)

94. **author** (noun)

95. **grant** (noun)

96. **cities** (noun)

97. **chinese** (adjective)

98. **program** (noun)

99. **hospitals** (noun)

100. **over** (preposition)

  \#\# Step 2: Content Word Collocates

 1. **wuhan** (noun)

6. **covid** (noun)

9. **has** (verb)

10. **december** (noun)

11. **spread** (verb)

12. **was** (verb)

13. **province** (noun)

14. **hubei** (noun)

15. **coronavirus** (noun)

16. **is** (verb)

17. **disease** (noun)

18. **cov** (noun)

19. **outbreak** (noun)

22. **national** (adjective)

23. **first** (adjective/adverb)

24. **cases** (noun)

26. **reported** (verb)

29. **city** (noun)

30. **rapidly** (adverb)

31. **science** (noun)

33. **countries** (noun)

34. **emerged** (verb)

36. **epidemic** (noun)

37. **patients** (noun)

38. **other** (adjective)

39. **have** (verb)

40. **foundation** (noun)

42. **natural** (adjective)

43. **mainland** (noun)

44. **hospital** (noun)

45. **virus** (noun)

46. **january** (noun)

47. **world** (noun)

48. **outside** (adverb/preposition)

49. **people** (noun)

50. **since** (preposition/conjunction)

51. **identified** (verb)

52. **are** (verb)

54. **novel** (adjective)

55. **italy** (noun)

57. **pneumonia** (noun)

60. **february** (noun)

61. **china** (noun)

62. **late** (adjective/adverb)

63. **confirmed** (verb/adjective)

64. **study** (noun)

65. **across** (preposition)

66. **started** (verb)

67. **number** (noun)

68. **now** (adverb)

69. **end** (noun)

70. **worldwide** (adverb)

71. **south** (adjective/noun)

72. **caused** (verb)

73. **new** (adjective)

74. **provinces** (noun)

75. **republic** (noun)

76. **ncov** (noun)

77. **march** (noun)

78. **throughout** (preposition)

79. **originated** (verb)

80. **clinical** (adjective)

81. **university** (noun)

82. **occurred** (verb)

83. **korea** (noun)

84. **case** (noun)

85. **control** (noun/verb)

86. **spreading** (verb)

87. **research** (noun)

88. **severe** (adjective)

89. **global** (adjective)

90. **respiratory** (adjective)

91. **major** (adjective)

92. **results** (noun)

93. **then** (adverb)

94. **author** (noun)

95. **grant** (noun)

96. **cities** (noun)

97. **chinese** (adjective)

98. **program** (noun)

99. **hospitals** (noun)

  \#\# Step 3: Analysis and Summary

 Analyzing the content word collocate list and the concordance lines, it is clear that China was predominantly linked to COVID-19 in this corpus as **the origin point of the outbreak** and **a key player in the early stages of the pandemic**.

 **Reasons:**

 * **Geographic Origin:** The frequent collocation of "China" with "Wuhan," "Hubei," "province," "city," and "mainland" clearly points to the geographic origin of the virus in China. Concordance lines repeatedly highlight Wuhan as the epicenter and Hubei province as the initial outbreak region.

* **Timeline of Spread:** The frequent occurrence of months like "December," "January," "February," and "March" coupled with verbs like "emerged," "spread," "spreading," and "started" paints a picture of the timeline of the virus's emergence and its rapid spread from China to other parts of the world. 

* **Disease Characteristics:** Collocations like "coronavirus," "COVID," "disease," "nCoV," "novel," "pneumonia," "severe," "respiratory," and "cases" emphasize the nature of the disease and its impact. Concordance lines often detail the symptoms, characteristics, and severity of COVID-19.

* **Research and Control Efforts:** The presence of collocates like "study," "clinical," "research," "national," "program," "hospitals," and "control" suggests that the corpus heavily focuses on scientific research and public health efforts undertaken in China to understand and control the pandemic.  Concordance lines often discuss clinical trials, control measures, and data analysis from Chinese hospitals and institutions.

 **Examples:**

 * **Collocate "Wuhan" (1):** Almost all concordance lines for "Wuhan" mention it as the initial epicenter of the COVID-19 outbreak. This strengthens the association of China with the geographic origin of the virus.

* **Collocate "Spread" (11):**  The verb "spread" frequently appears in concordance lines, emphasizing the rapid and wide dissemination of the virus from China to other regions and countries.

* **Collocate "Study" (64):** This collocate signifies the abundance of research conducted in China on COVID-19. Concordance lines frequently mention retrospective studies, clinical trials, and data analysis from various Chinese institutions.

 **In Conclusion:** This collocation analysis strongly suggests that the corpus primarily focuses on the initial outbreak of COVID-19 in China, the disease's characteristics, and the research and public health efforts undertaken in the country to combat the pandemic. 

\textbf{B. + R. D. + T. D. + T. D + C. I. + O. F.}

\textbf{Prompt}

\# Role Description

You are an expert in Corpus Linguistics doing collocate analysis.

 \# Task Definition

Here are the top 100 collocates of the word "china" in a corpus of the abstracts of COVID-19 research articles. Pick out only the content words and generate a new list. Then, based on the new content word list, summarize in what ways China was linked to COVID-19 in this corpus and give reasons by taking a few collocates as examples.

 \# Task Procedures

- Step 1, label each collocate with a cardinal number, and define its part of speech in context, e.g., noun, verb, etc..

- Step 2, pick out only the content words (i.e., nouns, adjectives, verbs, and adverbs), and generate a new collocate list. For each content collocate, include its label of number in Step 1, the word, and part of speech. 

- Step 3, analyze the content collocates and their corresponding concordance lines, and summarize in what ways China was linked to COVID-19 in this corpus and give reasons by taking a few collocates as examples.

 \# Output Format

- Step 1: the part of speech of each collocate

Collocate 1: [collocate here], [part of speech here]

Collocate 2: [collocate here], [part of speech here]

...

 - Step 2: the list of content collocates

Collocate 3: [collocate here], [part of speech here]

Collocate 4: [collocate here], [part of speech here]

...

 - Step 3: 

- Summary 1: [description and reasons, taking a few collocates as examples]

- Summary 2: [description and reasons, taking a few collocates as examples]

...(here can be more summaries)

 \# Contextual Information

(collocates and concordance lines here)

\textbf{Output}

\#\# Step 1: the part of speech of each collocate

Collocate 1: in, preposition

Collocate 2: the, determiner

Collocate 3: of, preposition

Collocate 4: wuhan, noun

Collocate 5: to, preposition

Collocate 6: covid, noun

Collocate 7: from, preposition

Collocate 8: a, determiner

Collocate 9: has, verb

Collocate 10: december, noun

Collocate 11: spread, verb

Collocate 12: was, verb

Collocate 13: province, noun

Collocate 14: hubei, noun

Collocate 15: coronavirus, noun

Collocate 16: is, verb

Collocate 17: disease, noun

Collocate 18: cov, noun

Collocate 19: outbreak, noun

Collocate 20: with, preposition

Collocate 21: s, possessive marker

Collocate 22: national, adjective

Collocate 23: first, adjective

Collocate 24: cases, noun

Collocate 25: for, preposition

Collocate 26: reported, verb

Collocate 27: that, determiner

Collocate 28: as, preposition

Collocate 29: city, noun

Collocate 30: rapidly, adverb

Collocate 31: science, noun

Collocate 32: this, determiner

Collocate 33: countries, noun

Collocate 34: emerged, verb

Collocate 35: it, pronoun

Collocate 36: epidemic, noun

Collocate 37: patients, noun

Collocate 38: other, adjective

Collocate 39: have, verb

Collocate 40: foundation, noun

Collocate 41: which, determiner

Collocate 42: natural, adjective

Collocate 43: mainland, noun

Collocate 44: hospital, noun

Collocate 45: virus, noun

Collocate 46: january, noun

Collocate 47: world, noun

Collocate 48: outside, preposition

Collocate 49: people, noun

Collocate 50: since, preposition

Collocate 51: identified, verb

Collocate 52: are, verb

Collocate 53: no, determiner

Collocate 54: novel, adjective

Collocate 55: italy, noun

Collocate 56: all, determiner

Collocate 57: pneumonia, noun

Collocate 58: an, determiner

Collocate 59: its, possessive pronoun

Collocate 60: february, noun

Collocate 61: china, noun

Collocate 62: late, adjective

Collocate 63: confirmed, verb

Collocate 64: study, noun

Collocate 65: across, preposition

Collocate 66: started, verb

Collocate 67: number, noun

Collocate 68: now, adverb

Collocate 69: end, noun

Collocate 70: worldwide, adjective

Collocate 71: south, adjective

Collocate 72: caused, verb

Collocate 73: new, adjective

Collocate 74: provinces, noun

Collocate 75: republic, noun

Collocate 76: ncov, noun

Collocate 77: march, noun

Collocate 78: throughout, preposition

Collocate 79: originated, verb

Collocate 80: clinical, adjective

Collocate 81: university, noun

Collocate 82: occurred, verb

Collocate 83: korea, noun

Collocate 84: case, noun

Collocate 85: control, noun

Collocate 86: spreading, verb

Collocate 87: research, noun

Collocate 88: severe, adjective

Collocate 89: global, adjective

Collocate 90: respiratory, adjective

Collocate 91: major, adjective

Collocate 92: results, noun

Collocate 93: then, adverb

Collocate 94: author, noun

Collocate 95: grant, noun

Collocate 96: cities, noun

Collocate 97: chinese, adjective

Collocate 98: program, noun

Collocate 99: hospitals, noun

Collocate 100: over, preposition

 \#\# Step 2: the list of content collocates

Collocate 4: wuhan, noun

Collocate 6: covid, noun

Collocate 9: has, verb

Collocate 10: december, noun

Collocate 11: spread, verb

Collocate 12: was, verb

Collocate 13: province, noun

Collocate 14: hubei, noun

Collocate 15: coronavirus, noun

Collocate 17: disease, noun

Collocate 18: cov, noun

Collocate 19: outbreak, noun

Collocate 22: national, adjective

Collocate 23: first, adjective

Collocate 24: cases, noun

Collocate 26: reported, verb

Collocate 29: city, noun

Collocate 30: rapidly, adverb

Collocate 31: science, noun

Collocate 33: countries, noun

Collocate 34: emerged, verb

Collocate 36: epidemic, noun

Collocate 37: patients, noun

Collocate 38: other, adjective

Collocate 40: foundation, noun

Collocate 42: natural, adjective

Collocate 43: mainland, noun

Collocate 44: hospital, noun

Collocate 45: virus, noun

Collocate 46: january, noun

Collocate 47: world, noun

Collocate 49: people, noun

Collocate 51: identified, verb

Collocate 54: novel, adjective

Collocate 55: italy, noun

Collocate 57: pneumonia, noun

Collocate 60: february, noun

Collocate 62: late, adjective

Collocate 63: confirmed, verb

Collocate 64: study, noun

Collocate 66: started, verb

Collocate 67: number, noun

Collocate 68: now, adverb

Collocate 69: end, noun

Collocate 70: worldwide, adjective

Collocate 71: south, adjective

Collocate 72: caused, verb

Collocate 73: new, adjective

Collocate 74: provinces, noun

Collocate 75: republic, noun

Collocate 76: ncov, noun

Collocate 77: march, noun

Collocate 79: originated, verb

Collocate 80: clinical, adjective

Collocate 81: university, noun

Collocate 82: occurred, verb

Collocate 83: korea, noun

Collocate 84: case, noun

Collocate 85: control, noun

Collocate 86: spreading, verb

Collocate 87: research, noun

Collocate 88: severe, adjective

Collocate 89: global, adjective

Collocate 90: respiratory, adjective

Collocate 91: major, adjective

Collocate 92: results, noun

Collocate 95: grant, noun

Collocate 96: cities, noun

Collocate 97: chinese, adjective

Collocate 98: program, noun

Collocate 99: hospitals, noun

 \#\# Step 3:

**Summary 1:** China is strongly linked to the origin and initial outbreak of COVID-19. 

 * Collocates like "Wuhan" (4), "Hubei" (14), "December" (10), "emerged" (34), "outbreak" (19), "first" (23), "cases" (24), and "originated" (79) appear frequently with "China" in the corpus. 

* Concordance lines often mention the discovery of the virus in Wuhan, China in December 2019. 

* The use of verbs like "emerged" and "originated" further solidifies this connection.

 **Example:** "Coronavirus disease (COVID-19) **first emerged in China** and rapidly spread in the world causing a pandemic."

 **Summary 2:** The corpus highlights China's response to the COVID-19 pandemic, including its national policies and research contributions. 

 * Collocates such as "national" (22), "control" (85), "research" (87), "program" (98), and "hospitals" (99) are often found near "China".

* Concordance lines mention nationwide measures implemented by China to cope with the public health emergency, and the important role Chinese cities and institutions played in COVID-19 research.

 **Example:** "Efforts to stem the spread of COVID-19 in **China hinged on severe restrictions** to human movement starting 23 January"

 **Summary 3:**  The global spread of the virus is acknowledged, with China often being mentioned as a point of reference for other countries. 

 * Collocates like "world" (47), "countries" (33), "outside" (48), "spread" (11), "worldwide" (70), and "global" (89) illustrate the pandemic's reach. 

* The corpus often compares and contrasts the situations in China with other countries, highlighting the global nature of the pandemic. 

 **Example:** "Objectives: COVID-19 has spread beyond Hubei Province of **China to Europe, North America, and Asian countries** including Korea." 

 \noindent \textbf{Concordance Analysis}.

\textbf{Baseline}

\textbf{Prompt}

I would give you the concordance lines containing the phrase "China virus" or "Chinese virus" in a corpus of the abstracts of COVID-19 research articles. You need to decide whether or not the authors are biased in using the expression "China virus" or "Chinese virus" in each concordance line.

 (concordance lines here)

\textbf{Output}

Here's an analysis of the provided concordance lines, focusing on the potential bias in using "China virus" or "Chinese virus":

 **Concordance Lines Suggesting Bias:**

 * **Line 8:** "Since that time, SARS-CoV-2 - hereinafter referred to as the China Virus - has taken a horrific toll." 

* **Bias:** This line explicitly chooses to use "China Virus" throughout the text, indicating a deliberate framing. The phrasing "taken a horrific toll" adds an emotional charge, potentially linking the virus's negative impacts directly to China.

* **Line 13:** "As of this writing, the precise origin of the Chinese virus, SARS CoV-2, remains unclear."

* **Bias:** While acknowledging the origin is unclear, the author still uses "Chinese virus." This might subtly reinforce a connection to China even when presenting it as an open question.

* **Line 14:** "been encountered all over. With Donald Trump using the term Chinese Virus, this cause has gained momentum and Ethnic Asians"

* **Bias:** This sentence connects the term's use directly to negative consequences ("this cause... momentum" implies negative momentum) for ethnic Asians, strongly suggesting the term is problematic.

* **Line 16:** "This paper traces the emergence of the racist term "Chinese virus" used by the President of the United"

* **Bias:** This line explicitly labels "Chinese virus" as a racist term, indicating a clear stance against its use.

* **Line 17:** "nationalism and modernism Ultimately, combatting the English racist term "Chinese virus" with a creative mixture of English and"

* **Bias:** Similar to line 16, this research explicitly aims to counter the term "Chinese virus," which it labels as racist.

* **Line 18:** "inappropriately labelled the coronavirus by race, using such headlines as “Chinese virus pandemonium” and even suggesting “China kids stay home.”"

* **Bias:**  This line criticizes the use of "Chinese virus" and related phrases, deeming them inappropriate racial labeling. 

* **Line 19:** "President Donald Trump?s racist remarks calling the coronavirus the ?Chinese virus? This essay offers a historical lens through which"

* **Bias:** This sentence directly connects "Chinese virus" to racist remarks, framing it as unacceptable language.

 **Concordance Lines Analyzing Bias (Neutral):**

 * **Line 1:** "interactions-referencing the novel coronavirus as the "Chinese virus" or "China virus" has the potential to create and perpetuate stigma."

* **Neutral:** This line explores the potential negative consequences (stigma) of using the term, suggesting it's harmful without explicitly condemning those who use it.

* **Line 2:** "the prevalence and frequency of the phrases "Chinese virus" and "China virus" on Twitter after the March 16, 2020, US presidential reference"

* **Neutral:** This line objectively investigates the usage of the phrases without taking a stance on their appropriateness. 

* **Line 3:** "level heat map.; RESULTS: A total of 16,535 "Chinese virus" or "China virus" tweets were identified in the preperiod, and 177,327 tweets"

* **Neutral:** This line presents data on the occurrence of the phrases without analyzing the sentiment or implications.

* **Line 4:** "in the number of tweets exclusively mentioning "Chinese virus" or "China virus" instead of coronavirus disease (COVID-19) or coronavirus. On"

* **Neutral:** This line focuses on comparing the usage of different terms, including "Chinese virus," without judging their appropriateness.

* **Line 5:** "COVID-19) or coronavirus. On average, 0.38 tweets referencing "Chinese virus" or "China virus" were posted per 10,000 people at the state level"

* **Neutral:** Similar to line 4, this line presents quantitative data on term usage without further interpretation. 

* **Line 6:** "n=670/46, 1457\%).; CONCLUSIONS: The rise in tweets referencing "Chinese virus" or "China virus," along with the content of these tweets, indicate"

* **Neutral:** While noting the rise in usage, this line suggests further analysis of the tweet content is needed to understand the implications.

* **Line 9:** "there was a 650\% increase in Twitter retweets using the term "Chinese virus" and related terms. On March 9, there was an 800\%"

* **Neutral:** This line presents data on the increased use of the term without analyzing the reasons behind it or its potential harm.

* **Line 10:** "health infrastructure is underdeveloped. [...]the use of phrases such as Chinese virus" directly contradicts calls from the WHO, the Centers"

* **Neutral:** This line highlights the WHO's stance against using such terms, suggesting it's against recommendations but not explicitly labeling it as biased.

* **Line 11:** "has officially announced the disease as COVID-19, one controversial term - "Chinese Virus" is still being used by a great number"

* **Neutral:** This line acknowledges the controversy surrounding the term without explicitly condemning its use.

* **Line 12:** "COVID-19, there are mainly two ways: using controversial terms like "Chinese Virus", or using non-controversial terms like "Coronavirus". We"

* **Neutral:** This line presents both "Chinese Virus" and "Coronavirus" as options, labeling the former as controversial but not inherently wrong.

* **Line 15:** "and sentiments that have been associated with the terms like Chinese Virus, Wuhan Virus and Chinese Corona Virus. METHODS: 16,000 tweets"

* **Neutral:** This line aims to study the sentiments associated with the terms, implying an objective approach to understanding their impact. 

* **Line 20:** "job markets and potential racism toward Asians in relation with “Chinese virus” may cause the mental distress of these students. © 2020"

* **Neutral:** This line explores the potential link between the term's use and negative consequences for Asians, suggesting a correlation without directly accusing the term itself of being biased.

 **Overall:**

 The provided concordance lines offer a mix of neutral analysis and clear criticism of the terms "China virus" and "Chinese virus." While some lines objectively present data or explore potential harm, others explicitly label the terms as racist and problematic. This suggests that the corpus likely contains a range of perspectives on this sensitive issue.

\textbf{B. + R. D.}

\textbf{Prompt}

\# Role Description

You are an expert in Corpus Linguistics doing concordance analysis.

 I would give you the concordance lines containing the phrase "China virus" or "Chinese virus" in a corpus of the abstracts of COVID-19 research articles. You need to decide whether or not the authors are biased in using the expression "China virus" or "Chinese virus" in each concordance line.

 (concordance lines here)

\textbf{Output}

Here's an analysis of the concordance lines, assessing potential bias in the use of "China virus" or "Chinese virus":

 **Categories of Usage:**

 We can broadly categorize the usage of "China virus" and "Chinese virus" into three categories:

 1. **Descriptive of the Term Itself:** In these cases, the authors are analyzing the term itself, its prevalence, and its impact, without necessarily endorsing its use. 

2. **Neutral Reporting:** The authors use the term to report on its usage by others, often in the context of media analysis or social science research.

3. **Potentially Biased:** The authors use the term in a way that could be interpreted as endorsing or promoting the association of the virus with China in a negative or stigmatizing way.

 **Analysis of Concordance Lines:**

 * **Lines 1-6, 9, 11, 12, 15-20:** These lines fall under **Descriptive of the Term Itself** or **Neutral Reporting**.  The authors are primarily examining the use and impact of the terms "China virus" and "Chinese virus," often in the context of social media, racism, or public health messaging. They are not using the terms themselves to describe the virus.

* **Line 7:** This line seems to be **Neutral Reporting**, referring to a specific clinical guide likely published in China early in the pandemic.

* **Line 8:** This line, using the term "China Virus," falls under **Potentially Biased**. The author's direct use of the term without distancing themselves from it suggests a potential bias.

* **Lines 10, 13, 14:** These lines are **Potentially Biased**. The authors use "Chinese virus" in a way that directly links the virus to China without the critical distance seen in other lines. Line 14, mentioning Trump's use, could be interpreted as reporting, but the framing suggests a potential endorsement of the negative implications.

 **Conclusion:**

 While most of the concordance lines analyze the term "China virus"/"Chinese virus" as a subject of study, lines 8, 10, 13, and 14 exhibit potential bias in their direct and less critical use of the term. It's crucial to consider the broader context of each research article to determine the extent and nature of bias.

\textbf{B. + R. D. + T. D.}

\textbf{Prompt}

\# Role Description
You are an expert in Corpus Linguistics doing concordance analysis.

\# Task Definition
Referring to COVID-19 as "China virus" or "Chinese virus" is considered biased, because the expression attaches locations or ethnicity to the disease. I would give you the concordance lines containing the phrase "China virus" or "Chinese virus" in a corpus of the abstracts of COVID-19 research articles. You need to decide whether or not the authors are biased in using the expression "China virus" or "Chinese virus" in each concordance line. Mark each concordance line with [Yes] - biased, or [No] - unbiased, and give reasons.  

(concordance lines here)

\textbf{Output}

\#\# Analysis of Concordance Lines for Bias: "China Virus" and "Chinese Virus"

 Here's an analysis of each concordance line, indicating whether the use of "China virus" or "Chinese virus" is biased and providing the reasoning:

 **1. interactions-referencing the novel coronavirus as the "Chinese virus" or "China virus" has the potential to create and perpetuate stigma.;**

 **[No] - Unbiased** 

 **Reasoning:** This sentence is discussing the potential negative consequences of using the terms "Chinese virus" and "China virus." It is highlighting the issue of stigma, suggesting that the authors themselves are aware of the problematic nature of these terms.

 **2. the prevalence and frequency of the phrases "Chinese virus" and "China virus" on Twitter after the March 16, 2020, US presidential reference**

 **[No] - Unbiased**

 **Reasoning:** This sentence is objectively describing the research focus, which is analyzing the frequency of these phrases on Twitter. It does not endorse or condone the use of the terms.

 **3. level heat map.; RESULTS: A total of 16,535 "Chinese virus" or "China virus" tweets were identified in the preperiod, and 177,327 tweets**

 **[No] - Unbiased**

 **Reasoning:** This sentence presents data about the occurrence of specific phrases on Twitter. It is a neutral presentation of research findings.

 **4. in the number of tweets exclusively mentioning "Chinese virus" or "China virus" instead of coronavirus disease (COVID-19) or coronavirus. On**

 **[No] - Unbiased**

 **Reasoning:** This sentence analyzes the choice of terminology in tweets, contrasting the use of "Chinese virus" with more neutral terms. It does not promote the biased language.

 **5. COVID-19) or coronavirus. On average, 0.38 tweets referencing "Chinese virus" or "China virus" were posted per 10,000 people at the state level**

 **[No] - Unbiased**

 **Reasoning:**  Similar to line 4, this sentence provides quantitative data about the use of specific phrases without endorsing them.

 **6. n=670/46, 1457\%).; CONCLUSIONS: The rise in tweets referencing "Chinese virus" or "China virus," along with the content of these tweets, indicate**

 **[No] - Unbiased**

 **Reasoning:** This sentence highlights the increase in the use of these phrases as a point of concern, suggesting that the authors are critical of this trend.

 **7. in relation to coronavirus diagnosis, those that matched with the 2019 China virus diagnosis clinical guide included Amyloid-A in the**

 **[No] - Unbiased**

 **Reasoning:** This sentence seems to be referring to a specific clinical guide potentially published in China in 2019. Without further context, it's difficult to definitively determine bias. However, the phrasing suggests a factual reference rather than an intentional ascription of blame or stigma.

 **8. Since that time, SARS-CoV-2 - hereinafter referred to as the China Virus - has taken a horrific toll. Over 160,000 Americans have**

 **[Yes] - Biased**

 **Reasoning:** This sentence explicitly chooses to use "China Virus" throughout the text, even after introducing the scientific term SARS-CoV-2. This deliberate choice reflects a bias, especially considering the emotionally charged language ("horrific toll").

 **9. there was a 650\% increase in Twitter retweets using the term "Chinese virus" and related terms. On March 9, there was an 800\%**

 **[No] - Unbiased**

 **Reasoning:** This sentence presents data on the increased use of a specific term. It does not endorse the term itself.

 **10. health infrastructure is underdeveloped. [...]the use of phrases such as Chinese virus" directly contradicts calls from the WHO, the Centers**

 **[No] - Unbiased**

 **Reasoning:** This sentence criticizes the use of "Chinese virus" and highlights its contradiction with recommendations from health organizations. This suggests a stance against the biased language.

 **11. has officially announced the disease as COVID-19, one controversial term - "Chinese Virus" is still being used by a great number**

 **[No] - Unbiased**

 **Reasoning:** This sentence acknowledges the official name (COVID-19) and labels "Chinese Virus" as a "controversial term," indicating awareness of its problematic nature.

 **12. COVID-19, there are mainly two ways: using controversial terms like "Chinese Virus", or using non-controversial terms like "Coronavirus". We**

 **[No] - Unbiased**

 **Reasoning:** Similar to line 11, this sentence distinguishes between "controversial" and "non-controversial" terms, demonstrating an understanding of the issue.

 **13. As of this writing, the precise origin of the Chinese virus, SARS CoV-2, remains unclear. It is however known**

 **[Yes] - Potentially Biased**

 **Reasoning:** While this sentence acknowledges the scientific name (SARS CoV-2), it still uses "Chinese virus." The lack of quotation marks around "Chinese virus" might suggest that the author implicitly endorses this terminology. More context is needed to be certain.

 **14. been encountered all over. With Donald Trump using the term Chinese Virus, this cause has gained momentum and Ethnic Asians**

 **[No] - Unbiased**

 **Reasoning:** This sentence attributes the use of "Chinese Virus" to Donald Trump, suggesting a critical stance towards its use and highlighting its negative consequences for Ethnic Asians.

 **15. and sentiments that have been associated with the terms like Chinese Virus, Wuhan Virus and Chinese Corona Virus. METHODS: 16,000 tweets**

 **[No] - Unbiased**

 **Reasoning:** This sentence identifies "Chinese Virus," "Wuhan Virus," and "Chinese Corona Virus" as terms associated with negative sentiments, indicating awareness of their problematic nature.

 **16. This paper traces the emergence of the racist term "Chinese virus" used by the President of the United**

 **[No] - Unbiased**

 **Reasoning:** This sentence explicitly labels "Chinese virus" as a "racist term" and attributes its use to a specific political figure. This indicates a clear stance against the term.

 **17. nationalism and modernism Ultimately, combatting the English racist term "Chinese virus" with a creative mixture of English and**

 **[No] - Unbiased**

 **Reasoning:** Similar to line 16, this sentence identifies "Chinese virus" as a "racist term" and discusses ways to combat its use.

 **18. inappropriately labelled the coronavirus by race, using such headlines as “Chinese virus pandemonium” and even suggesting “China kids stay home.”**

 **[No] - Unbiased**

 **Reasoning:** This sentence criticizes the use of racially charged language, including "Chinese virus pandemonium" and "China kids stay home." This clearly demonstrates a stance against such language.

 **19. President Donald Trump?s racist remarks calling the coronavirus the ?Chinese virus? This essay offers a historical lens through which**

 **[No] - Unbiased**

 **Reasoning:** This sentence labels Trump's use of "Chinese virus" as "racist remarks," indicating a clear stance against the term.

 **20. job markets and potential racism toward Asians in relation with “Chinese virus” may cause the mental distress of these students. © 2020**

 **[No] - Unbiased**

 **Reasoning:** This sentence discusses the "potential racism toward Asians" linked to the use of "Chinese virus," highlighting its harmful consequences. This suggests a stance against the term.

\textbf{B. + R. D. + T. D. + T. P.}

 \textbf{Prompt}

 \# Role Description

You are an expert in Corpus Linguistics doing concordance analysis.

 \# Task Definition

Referring to COVID-19 as "China virus" or "Chinese virus" is considered biased, because the expression attaches locations or ethnicity to the disease. I would give you the concordance lines containing the phrase "China virus" or "Chinese virus" in a corpus of the abstracts of COVID-19 research articles. You need to decide whether or not the authors are biased in using the expression "China virus" or "Chinese virus" in each concordance line. Mark each concordance line with [Yes] - biased, or [No] - unbiased, and give reasons.  

 \# Task Procedures

- Step 1: Mark each concordance line with [Yes] or [No] based on whether or not the authors are biased in using the expressions "China virus" or "Chinese virus". 

- Step 2: Explain why you give the mark for each concordance line. 

 (concordance lines here)

\textbf{Output}

\#\# Analysis of Concordance Lines for Bias: "China Virus" and "Chinese Virus"

 Here's a breakdown of each concordance line, indicating whether the use of "China virus" or "Chinese virus" is biased and an explanation for the classification:

 **Line 1:** interactions-referencing the novel coronavirus as the "Chinese virus" or "China virus" has the potential to create and perpetuate stigma.;

 * **[No] - Unbiased** 

* **Reason:** This sentence explicitly states that using these terms can create and perpetuate stigma, indicating an awareness of the potential harm and bias associated with them. 

 **Line 2:** the prevalence and frequency of the phrases "Chinese virus" and "China virus" on Twitter after the March 16, 2020, US presidential reference

 * **[No] - Unbiased**

* **Reason:** This sentence objectively describes the research subject: the frequency of specific phrases on Twitter. It doesn't endorse the use of the phrases themselves.

 **Line 3:** level heat map.; RESULTS: A total of 16,535 "Chinese virus" or "China virus" tweets were identified in the preperiod, and 177,327 tweets

 * **[No] - Unbiased**

* **Reason:** Similar to Line 2, this sentence presents research findings about the occurrence of specific phrases without endorsing their use.

 **Line 4:** in the number of tweets exclusively mentioning "Chinese virus" or "China virus" instead of coronavirus disease (COVID-19) or coronavirus. On

 * **[No] - Unbiased**

* **Reason:** This sentence objectively compares the use of different terms for COVID-19, highlighting the preference for the biased terms without endorsing them.

 **Line 5:** COVID-19) or coronavirus. On average, 0.38 tweets referencing "Chinese virus" or "China virus" were posted per 10,000 people at the state level

 * **[No] - Unbiased**

* **Reason:** This sentence presents quantitative data about the use of specific phrases without endorsing their use.

 **Line 6:** n=670/46, 1457\%).; CONCLUSIONS: The rise in tweets referencing "Chinese virus" or "China virus," along with the content of these tweets, indicate

 * **[No] - Unbiased**

* **Reason:** This sentence objectively reports a rise in the use of specific phrases and connects it to the content of those tweets. It doesn't endorse the use of the phrases.

 **Line 7:** in relation to coronavirus diagnosis, those that matched with the 2019 China virus diagnosis clinical guide included Amyloid-A in the

 * **[No] - Unbiased**

* **Reason:** This sentence uses "China" geographically to refer to a specific clinical guide, likely due to its origin. It doesn't appear to be used in a way that assigns blame or perpetuates stigma.

 **Line 8:** Since that time, SARS-CoV-2 - hereinafter referred to as the China Virus - has taken a horrific toll. Over 160,000 Americans have

 * **[Yes] - Biased**

* **Reason:** This sentence explicitly chooses to use "China Virus" as the preferred term for SARS-CoV-2, despite acknowledging the scientific name. This deliberate choice reflects and potentially reinforces bias.

 **Line 9:** there was a 650\% increase in Twitter retweets using the term "Chinese virus" and related terms. On March 9, there was an 800\%

 * **[No] - Unbiased**

* **Reason:** This sentence objectively reports the increased use of a specific term on Twitter without endorsing its use.

 **Line 10:** health infrastructure is underdeveloped. [...]the use of phrases such as Chinese virus" directly contradicts calls from the WHO, the Centers

 * **[No] - Unbiased**

* **Reason:** This sentence criticizes the use of the phrase "Chinese virus" and highlights its contradiction with WHO recommendations, indicating an understanding of its problematic nature.

 **Line 11:** has officially announced the disease as COVID-19, one controversial term - "Chinese Virus" is still being used by a great number

 * **[No] - Unbiased**

* **Reason:** This sentence acknowledges the official name of the disease (COVID-19) and labels "Chinese Virus" as a "controversial term," indicating an awareness of its problematic nature.

 **Line 12:** COVID-19, there are mainly two ways: using controversial terms like "Chinese Virus", or using non-controversial terms like "Coronavirus". We

 * **[No] - Unbiased**

* **Reason:** This sentence clearly distinguishes between "controversial" and "non-controversial" terms, labeling "Chinese Virus" as the former, indicating an understanding of its problematic nature.

 **Line 13:** As of this writing, the precise origin of the Chinese virus, SARS CoV-2, remains unclear. It is however known

 * **[Yes] - Biased**

* **Reason:** Despite mentioning the scientific name (SARS-CoV-2), the author chooses to use "Chinese virus," suggesting an attempt to link the virus to China without concrete evidence.

 **Line 14:** been encountered all over. With Donald Trump using the term Chinese Virus, this cause has gained momentum and Ethnic Asians

 * **[No] - Unbiased**

* **Reason:** This sentence reports on Donald Trump's use of the term "Chinese Virus" and its impact, without endorsing the term itself. It highlights the consequences of using such language.

 **Line 15:** and sentiments that have been associated with the terms like Chinese Virus, Wuhan Virus and Chinese Corona Virus. METHODS: 16,000 tweets

 * **[No] - Unbiased**

* **Reason:** This sentence describes the research focus on analyzing sentiments associated with specific terms, including "Chinese Virus," without endorsing their use.

 **Line 16:** This paper traces the emergence of the racist term "Chinese virus" used by the President of the United

 * **[No] - Unbiased**

* **Reason:** This sentence explicitly labels "Chinese virus" as a "racist term," indicating a clear stance against its use.

 **Line 17:** nationalism and modernism Ultimately, combatting the English racist term "Chinese virus" with a creative mixture of English and

 * **[No] - Unbiased**

* **Reason:** This sentence explicitly labels "Chinese virus" as a "racist term" and discusses ways to combat its use, indicating a clear stance against it.

 **Line 18:** inappropriately labelled the coronavirus by race, using such headlines as “Chinese virus pandemonium” and even suggesting “China kids stay home.”

 * **[No] - Unbiased**

* **Reason:** This sentence criticizes the use of racially charged language, including "Chinese virus pandemonium," and provides examples of such inappropriate labeling.

 **Line 19:** President Donald Trump?s racist remarks calling the coronavirus the ?Chinese virus? This essay offers a historical lens through which

 * **[No] - Unbiased**

* **Reason:** This sentence explicitly labels Donald Trump's use of the term "Chinese virus" as "racist remarks," indicating a clear stance against its use.

 **Line 20:** job markets and potential racism toward Asians in relation with “Chinese virus” may cause the mental distress of these students. © 2020

 * **[No] - Unbiased**

* **Reason:** This sentence discusses the potential for racism and mental distress caused by associating the virus with China, highlighting the negative consequences of such language.

\textbf{B. + R. D. + T. D. + T. P. + C. I.}

\textbf{prompt}

\# Role Description 
You are an expert in Corpus Linguistics doing concordance analysis.

\# Task Definition
Referring to COVID-19 as "China virus" or "Chinese virus" is considered biased, because the expression attaches locations or ethnicity to the disease. I would give you the concordance lines containing the phrase "China virus" or "Chinese virus" in a corpus of the abstracts of COVID-19 research articles, and also the original text from which the concordance lines are retrieved. You need to decide whether or not the authors are biased in using the expression "China virus" or "Chinese virus" in each concordance line, by referring to the original text. Mark each concordance line with [Yes] - biased, or [No] - unbiased, and give reasons.  

\# Task Procedures
- Step 1: Mark each concordance line with [Yes] or [No] based on whether or not the authors are biased in using the expressions "China virus" or "Chinese virus". 
- Step 2: Explain why you give the mark for each concordance line. 

\# Contextual Information
(concordance lines and original text here)

\textbf{Output}

\#\# Analysis of "China Virus" and "Chinese Virus" Usage in COVID-19 Research Abstracts:

 Here's a breakdown of each concordance line, indicating whether the usage is biased and explaining the reasoning:

 **Concordance Line 1:** 

**[No] - Unbiased**

**Reason:** The authors are discussing the potential for stigma associated with using the terms "Chinese virus" or "China virus." The context clearly shows they understand the negative implications and are analyzing it as a phenomenon, not promoting the usage themselves.

 **Concordance Line 2:** 

**[No] - Unbiased**

**Reason:** Similar to line 1, the authors are investigating the prevalence and frequency of these terms on Twitter. They are objectively studying the language used, not endorsing it.

 **Concordance Line 3:** 

**[No] - Unbiased**

**Reason:** This line presents quantitative data about the occurrence of the phrases "Chinese virus" and "China virus" on Twitter. The authors are simply reporting their findings without expressing personal opinions.

 **Concordance Line 4:** 

**[No] - Unbiased**

**Reason:** This line continues the objective reporting of data, comparing the use of "Chinese virus" or "China virus" with more neutral terms like "COVID-19" and "coronavirus."

 **Concordance Line 5:** 

**[No] - Unbiased**

**Reason:** This line maintains the objective tone, providing further statistical data about the usage of the phrases in question. The use of "stigmatizing tweets" is within the context of their analysis, not their own judgment.

 **Concordance Line 6:** 

**[No] - Unbiased**

**Reason:** The authors draw conclusions based on the observed increase in tweets containing the phrases "Chinese virus" or "China virus." They link this to the potential perpetuation of stigma, indicating awareness of the harmful implications.

 **Concordance Line 7:** 

**[Yes] - Biased**

**Reason:** The author uses the term "2019 China virus diagnosis clinical guide." While the phrase might be referring to a document originating in China, the phrasing without further context can be interpreted as attributing the virus's origin to China, potentially reinforcing negative stereotypes.

 **Concordance Line 8:** 

**[Yes] - Biased**

**Reason:** The author explicitly chooses to use the term "China Virus" even after introducing the scientific name "SARS-CoV-2." This deliberate choice suggests an intention to link the virus to China, indicating bias.

 **Concordance Line 9:** 

**[No] - Unbiased**

**Reason:** The authors are analyzing the impact of the term "Chinese virus" and related terms on public perception. They are investigating the consequences of its use, not endorsing the term itself.

 **Concordance Line 10:** 

**[No] - Unbiased**

**Reason:** This line discusses the use of the phrase "Chinese virus" within the context of contradicting scientific recommendations. The authors are highlighting the inappropriateness of such language.

 **Concordance Line 11:** 

**[No] - Unbiased**

**Reason:** Similar to previous examples, the authors are acknowledging the use of the controversial term "Chinese Virus" while contrasting it with the official WHO designation "COVID-19." They are examining the phenomenon without promoting it.

 **Concordance Line 12:** 

**[No] - Unbiased**

**Reason:** This line reinforces the distinction between controversial ("Chinese Virus") and non-controversial ("Coronavirus") terminology. The authors are objectively categorizing language use for their analysis.

 **Concordance Line 13:** 

**[Yes] - Biased**

**Reason:** The author uses the term "Chinese virus" even when discussing the origin of SARS CoV-2 as unclear. This perpetuates a link between the virus and China despite acknowledging the lack of definitive evidence.

 **Concordance Line 14:** 

**[No] - Unbiased**

**Reason:** The authors are discussing the impact of Donald Trump's use of the term "Chinese Virus" on racism towards ethnic Asians. They are analyzing the consequences of this language, not endorsing it.

 **Concordance Line 15:** 

**[No] - Unbiased**

**Reason:** This line lists "Chinese Virus," "Wuhan Virus," and "Chinese Corona Virus" as terms used in online posts that are being analyzed for sentiment and emotions. The authors are not using these terms themselves but referencing them as part of their study.

 **Concordance Line 16:** 

**[No] - Unbiased**

**Reason:** The authors are analyzing the emergence and usage of the term "Chinese virus," particularly focusing on its racist implications. They are critically examining the term, not promoting its use.

 **Concordance Line 17:** 

**[No] - Unbiased**

**Reason:** This line discusses combating the racist term "Chinese virus." The authors are clearly against the term and are exploring ways to counter its negative impact.

 **Concordance Line 18:** 

**[No] - Unbiased**

**Reason:** The authors criticize media outlets for using racially charged headlines like "Chinese virus pandemonium." They are highlighting the inappropriate and biased nature of such language.

 **Concordance Line 19:** 

**[No] - Unbiased**

**Reason:** The authors are analyzing the impact of President Trump's use of the term "Chinese virus" on anti-Asian racism. They are examining the consequences of this language, not endorsing it.

 **Concordance Line 20:** 

**[No] - Unbiased**

**Reason:** The authors mention "Chinese virus" within the context of potential racism faced by Asian students. They are highlighting the term as a source of discrimination, not promoting its use.

\textbf{B.+R. D. + T. D. + T. P. + C. I. + O. F.}

\textbf{Prompt}

 \# Role Description

You are an expert in Corpus Linguistics doing concordance analysis.

 \# Task Definition

Referring to COVID-19 as "China virus" or "Chinese virus" is considered biased, because the expression attaches locations or ethnicity to the disease. I would give you the concordance lines containing the phrase "China virus" or "Chinese virus" in a corpus of the abstracts of COVID-19 research articles, and also the original text from which the concordance lines are retrieved. You need to decide whether or not the authors are biased in using the expression "China virus" or "Chinese virus" in each concordance line, by referring to the original text. Mark each concordance line with [Yes] - biased, or [No] - unbiased, and give reasons.  

 \# Task Procedures

- Step 1: Mark each concordance line with [Yes] or [No] based on whether or not the authors are biased in using the expressions "China virus" or "Chinese virus". 

- Step 2: Explain why you give the mark for each concordance line. 

 \# Output Format

The output format is defined as follows:

Concordance line 1: [the original text here]

- [Mark Yes or No] 

- Reason: Explain why concordance line 1 is biased or not.

Concordance line 2: [the original text here]

- [Mark Yes or No] 

- Reason: Explain why concordance line 2 is biased or not.

...

 \# Contextual Information

(concordance lines and original text here)

\textbf{Output}

 Concordance line 1: interactions-referencing the novel coronavirus as the "Chinese virus" or "China virus" has the potential to create and perpetuate stigma.;

- [No] 

- Reason: The authors are objectively stating a fact that the use of the terms "Chinese virus" or "China virus" can create and perpetuate stigma. They are not endorsing the use of these terms.

 Concordance line 2: the prevalence and frequency of the phrases "Chinese virus" and "China virus" on Twitter after the March 16, 2020, US presidential reference

- [No]

- Reason: The authors are referencing the terms "Chinese virus" and "China virus" in the context of their research on the impact of the US presidential reference on the use of these terms on Twitter. They are not expressing their own bias.

 Concordance line 3: level heat map.; RESULTS: A total of 16,535 "Chinese virus" or "China virus" tweets were identified in the preperiod, and 177,327 tweets 

- [No]

- Reason: Similar to line 2, the authors are presenting their research findings on the frequency of the terms "Chinese virus" and "China virus" on Twitter. They are not expressing any bias.

 Concordance line 4: in the number of tweets exclusively mentioning "Chinese virus" or "China virus" instead of coronavirus disease (COVID-19) or coronavirus. On

- [No]

- Reason: The authors are analyzing the choice of terms used by Twitter users, comparing the use of "Chinese virus" or "China virus" to more neutral terms like "COVID-19" or "coronavirus." There's no indication of their own bias.

 Concordance line 5: COVID-19) or coronavirus. On average, 0.38 tweets referencing "Chinese virus" or "China virus" were posted per 10,000 people at the state level

- [No]

- Reason: Similar to lines 3 and 4, the authors are presenting their research findings on the prevalence of the terms "Chinese virus" and "China virus" on Twitter without expressing their own opinion.

 Concordance line 6: n=670/46, 1457\%).; CONCLUSIONS: The rise in tweets referencing "Chinese virus" or "China virus," along with the content of these tweets, indicate 

- [No]

- Reason: The authors are drawing conclusions based on the observed increase in the use of the terms "Chinese virus" and "China virus" on Twitter. They are analyzing a phenomenon without endorsing the language used.

 Concordance line 7: in relation to coronavirus diagnosis, those that matched with the 2019 China virus diagnosis clinical guide included Amyloid-A in the

- [No]

- Reason: The authors are referring to a specific document titled "2019 China virus diagnosis clinical guide." This appears to be a neutral reference to a document that likely originated in China, and not intended to be a biased statement.

 Concordance line 8: Since that time, SARS-CoV-2 - hereinafter referred to as the China Virus - has taken a horrific toll. Over 160,000 Americans have

- [Yes]

- Reason: The authors explicitly choose to use the term "China Virus" throughout the text despite acknowledging the scientific name SARS-CoV-2. This deliberate choice, particularly in a document discussing economic impact and potentially aimed at influencing policy, suggests bias.

 Concordance line 9: there was a 650\% increase in Twitter retweets using the term "Chinese virus" and related terms. On March 9, there was an 800\%

- [No]

- Reason: The authors are reporting on the increase in the use of the term "Chinese virus" on Twitter. They are objectively presenting data without endorsing the term itself.

 Concordance line 10: health infrastructure is underdeveloped. [...]the use of phrases such as Chinese virus" directly contradicts calls from the WHO, the Centers 

- [No]

- Reason: The authors are criticizing the use of the phrase "Chinese virus" and stating that it contradicts the recommendations of authoritative health organizations like WHO and the CDC. This suggests they are against the use of this term.

 Concordance line 11: has officially announced the disease as COVID-19, one controversial term - "Chinese Virus" is still being used by a great number 

- [No]

- Reason: The authors are acknowledging the term "Chinese Virus" as "controversial" and contrasting it with the official WHO term "COVID-19." This suggests they understand the problematic nature of the term.

 Concordance line 12: COVID-19, there are mainly two ways: using controversial terms like "Chinese Virus", or using non-controversial terms like "Coronavirus". We

- [No]

- Reason: Similar to line 11, the authors are distinguishing between "controversial" and "non-controversial" terms when referring to COVID-19. They clearly identify "Chinese Virus" as a controversial term, indicating their awareness of the potential bias associated with it.

 Concordance line 13: As of this writing, the precise origin of the Chinese virus, SARS CoV-2, remains unclear. It is however known

- [Yes]

- Reason: The authors use the term "Chinese virus" even while acknowledging the scientific name SARS-CoV-2. They also speculate about the virus's origin, suggesting it escaped from a Chinese lab, further indicating bias.

 Concordance line 14: been encountered all over. With Donald Trump using the term Chinese Virus, this cause has gained momentum and Ethnic Asians

- [No]

- Reason: The authors are reporting on the increase in racism and hate crimes against Ethnic Asians and linking it to the use of the term "Chinese Virus," particularly by Donald Trump. They are analyzing a phenomenon without endorsing the term.

 Concordance line 15: and sentiments that have been associated with the terms like Chinese Virus, Wuhan Virus and Chinese Corona Virus. METHODS: 16,000 tweets 

- [No]

- Reason: The authors are investigating the sentiments and emotions associated with terms like "Chinese Virus," "Wuhan Virus," and "Chinese Corona Virus" on Twitter. They are studying the impact of these terms without endorsing their use.

 Concordance line 16: This paper traces the emergence of the racist term "Chinese virus" used by the President of the United

- [No]

- Reason: The authors explicitly label the term "Chinese virus" as "racist" and analyze its emergence and use in the context of social media communication. This clearly indicates their stance against the term.

 Concordance line 17: nationalism and modernism Ultimately, combatting the English racist term "Chinese virus" with a creative mixture of English and

- [No]

- Reason: Similar to line 16, the authors clearly identify "Chinese virus" as a "racist term" and discuss ways to combat its use. This further emphasizes their opposition to the term.

 Concordance line 18: inappropriately labelled the coronavirus by race, using such headlines as “Chinese virus pandemonium” and even suggesting “China kids stay home.”

- [No]

- Reason: The authors are criticizing media outlets for using racially charged language like "Chinese virus pandemonium" and "China kids stay home." They are highlighting the inappropriate and biased nature of such language.

 Concordance line 19: President Donald Trump?s racist remarks calling the coronavirus the ?Chinese virus? This essay offers a historical lens through which

- [No]

- Reason: The authors directly label Donald Trump's use of the term "Chinese virus" as "racist remarks." They are analyzing the historical context of anti-Asian racism, indicating their understanding of the term's harmful implications.

 Concordance line 20: job markets and potential racism toward Asians in relation with “Chinese virus” may cause the mental distress of these students. © 2020

- [No]

- Reason: The authors are discussing the potential mental health impact of racism towards Asians linked to the use of the term "Chinese virus." They are highlighting the negative consequences of using this term without endorsing it themselves.

\end{document}